\documentclass[mnsc,nonblindrev]{informs3_hide}
\OneAndAHalfSpacedXI 

\usepackage[english]{babel}
\usepackage[autostyle, english = american]{csquotes}
\MakeOuterQuote{"}
\usepackage[colorlinks,linkcolor=blue, citecolor=blue]{hyperref}
\usepackage[normalem]{ulem}

\usepackage{cleveref}
\crefname{subsection}{section}{subsections}
\usepackage{eqnarray}
\usepackage{algorithm,algpseudocode}
\usepackage{amsmath, bbm, enumitem, xparse, bm, lipsum}
\usepackage{amssymb}

\newcommand{\eps}{\epsilon}
\newcommand{\veps}{\varepsilon}
\newcommand{\bI}{\mathbbm{1}}

\newcommand{\Rad}{\mathrm{Rad}}

\newcommand{\VI}{V^{\mathrm{Infi}}}
\newcommand{\VF}{V^{\mathrm{Fi}}}
\newcommand{\Gap}{\mathrm{Gap}}

\newcommand{\bmu}{\bm{\mu}}

\usepackage[dvipsnames]{xcolor}

\usepackage{xparse}

\NewDocumentEnvironment{myproof}{o}
{\IfNoValueTF{#1}{\paragraph{{Proof.} }} {\paragraph{{#1.} }} }
{\hfill$\Halmos$}

\usepackage{natbib,bbm,multirow,multicol}
 \bibpunct[, ]{(}{)}{,}{a}{}{,}%
 \def\bibfont{\small}%
\TheoremsNumberedThrough     
\ECRepeatTheorems

\EquationsNumberedThrough    

\allowdisplaybreaks

\begin{document}


\RUNAUTHOR{Jiang, Ye}

\RUNTITLE{Approaching Constrained Markov Decision Process}

\TITLE{
\Large Achieving Instance-dependent Sample Complexity for Constrained Markov Decision Process
}

\ARTICLEAUTHORS{%
\AUTHOR{$\text{Jiashuo Jiang}^{\dag}$, $~\text{Yinyu Ye}^{\ddag \S}$}

\AFF{\  \\
$\dag~$Department of Industrial Engineering \& Decision Analytics, Hong Kong University of Science and Technology\\
$\ddag~$Institute for Computational and Mathematical Engineering, Stanford University\\
$\S~$Department of Management Science \& Engineering, Stanford University
}
}

\ABSTRACT{
 We consider the reinforcement learning problem for the constrained Markov decision process (CMDP), which plays a central role in satisfying safety or resource constraints in sequential learning and decision-making. In this problem, we are given finite resources and a MDP with unknown transition probabilities. At each stage, we take an action, collecting a reward and consuming some resources, all assumed to be unknown and need to be learned over time. In this work, we take the first step towards deriving optimal problem-dependent guarantees for the CMDP problems. We derive a logarithmic regret bound, which translates into a $O(\frac{1}{\Delta\cdot\veps}\cdot\log^2(1/\veps))$ sample complexity bound, with $\Delta$ being a problem-dependent parameter, yet independent of $\veps$. Our sample complexity bound is the first problem-dependent guarantee for CMDP problems. 
 To achieve this advance, we develop a new framework for analyzing CMDP problems. To be specific, our algorithm operates in the primal space and we resolve the primal LP for the CMDP problem at each period in an online manner, with \textit{adaptive} remaining resource capacities. The key elements of our algorithm are: i) a characterization of the instance hardness via LP basis, ii) an eliminating procedure that identifies one optimal basis of the primal LP, and; iii) a resolving procedure that is adaptive to the remaining resources and sticks to the characterized optimal basis.

}

\KEYWORDS{Constrained MDP, Reinforcement Learning, Online Packing/Linear Programming, Problem-dependent Bound}


\maketitle

\section{Introduction}

Reinforcement learning (RL) is pivotal in the realm of dynamic decision-making under uncertainty, where the objective is to maximize total reward through ongoing interaction with and learning from an enigmatic environment. Markov Decision Processes (MDPs) are a prevalent framework for encapsulating environmental dynamics. MDPs have been instrumental in various domains, such as inventory management \citep{alvo2023neural}, video gaming \citep{mnih2013playing}, robotics \citep{kober2013reinforcement}, recommender systems \citep{shani2005mdp}, and beyond. Yet, they fall short in accommodating additional constraints that may influence the formulation of the optimal policy and the decision-maker's engagement with the uncertain environment. Often, in MDP applications, there are stringent constraints on utilities or costs, emanating from areas like safe autonomous driving \citep{fisac2018general}, robotics \citep{ono2015chance}, and financial management \citep{tamar2012policy}. These constraints might also symbolize limitations on resources within resource allocation contexts with applications to revenue management \citep{jia2023online, jiang2023constant, li2023revenue}, inventory management and queueing systems \citep{chen2021primal}. Constrained MDPs (CMDPs), as introduced in \cite{altman1999constrained}, enhance MDPs to factor in constraints affecting long-term policy results. In CMDPs, the decision-maker aims to optimize cumulative rewards while adhering to these constraints. Our paper focuses on CMDPs, and we aim to develop efficient algorithmic solutions.

The significance of RL in CMDP contexts has garnered substantial attention in recent years. A variety of methods for tackling CMDPs have been developed, including the primal-dual technique (e.g. \cite{efroni2020exploration}) which leverages the Lagrangian dual of CMDPs and implements an online learning strategy for the iterative update of dual variables. Other methods encompass constrained optimization \citep{achiam2017constrained, sun2024constrained}, the Lyapunov technique \citep{chow2018lyapunov}, among others. Previous research has established \textit{minimax} bounds for CMDPs, representing the optimal regret that can be achieved for the most challenging problem within a specific class of problems. Nonetheless, these minimax regret bounds and worst-case scenarios can be overly conservative, leading to a disconnect between theoretical guarantees and practical performance for specific problem instances. A more tailored approach is warranted—one that ensures great performance on every single problem instance and furnishes problem-dependent guarantees. Our research takes the first step towards deriving optimal problem-dependent guarantees for CMDP problems.

\subsection{Main Results and Contributions}

In the exploration of the $\gamma$-discounted infinite horizon CMDP problem within a tabular context, we leverage a generative model that (as studied in \cite{kearns2002sparse, kakade2003sample, li2020breaking}), for each state-action pair, yields a sample from the unknown stochastic reward, costs, and state transitions. The main result of our research is the introduction of a novel algorithm that promises a $O(\kappa\cdot\log N)$ regret upper bound for both the cumulative reward and constraint violations, where $N$ is the number of queries to the generative model and $\kappa$ is a problem-specific parameter, yet independent of $N$. This translates into a $\tilde{O}\left(1/\veps\right)$ sample complexity bound, where $\tilde{O}(\cdot)$ hides a term polynomial in $\log(1/\veps)$. Note that the state-of-the-art sample complexity bound under the worst-case scenarios is $O(1/\veps^2)$, which has been established in a series of work \citep{wang2020randomized, jin2020efficiently, efroni2020exploration}. Though the $O(1/\veps)$ or better iteration complexity has been achieved in \cite{li2021faster, liu2021policy, zhou2022anchor, gladin2023algorithm}, it comes with a sample complexity bound no better than $O(1/\veps^2)$.
Our algorithm enjoys a sample complexity bound that has a better dependency in terms of $\veps$. To achieve these problem-dependent guarantees, we develop several elements.

First, we develop new characterizations of the problem instance hardness for CMDP problems. Note that a key component for achieving instance-dependent bounds is to characterize the ``hardness'' of the underlying problem instance. That is, we need to identify a positive gap to separate the optimal policy from the sub-optimal policies for a particular problem instance. The importance of identifying such a gap has been demonstrated in instance-optimal learning for multi-arm-bandit problems (e.g. \cite{lai1985asymptotically}) and reinforcement learning problems (e.g. \cite{auer2008near}), among others. The gap characterizes the hardness of the underlying problem instance (e.g. \cite{simchowitz2019non}).  For CMDP, identifying such a gap is non-trivial because the optimal policies for CMDP are randomized policies \citep{altman1999constrained}. Then, the policies can be represented by distributions over the action set and the sub-optimal policies can be arbitrarily close to the optimal policy as long as the corresponding sub-optimal distributions converge to the optimal one. To tackle this problem, we show that the feasible region for the policies can be represented as a polytope and we only need to focus on the corner points of this polytope to find an optimal policy. Therefore, the hardness can simply be characterized as the distance between the optimal corner point and the sub-optimal corner point, as illustrated in detail in \Cref{sec:characterization}. This is the first characterization of problem instance hardness for CMDP problems.

Then, we devise a new algorithmic framework to analyze CMDP problems, inspired by the online packing/linear programming (LP) literature \citep{agrawal2014dynamic, kesselheim2014primal, li2022online, li2021symmetry}. Specifically, we utilize a linear programming reformulation of the CMDP problem, where policies are delineated via \textit{occupancy measures} \citep{altman1999constrained}. The optimal policy emerges from the LP's solution; however, the indeterminate model parameters mean the LP cannot be solved directly but must be approached online as we obtain more samples from the generative model. Each generative model query leads to solving an empirical LP with accrued samples, and our final policy is derived from averaging these solutions—an approach akin to the methodology in online LP. A critical feature of our algorithm is the adaptiveness of the LP constraints' right-hand side to the input samples, a technique demonstrated to achieve logarithmic regret in online LP literature, which we now apply to CMDP problems.

Furthermore, we extend our contributions to the online LP literature. The prevalent assumption of non-degeneracy in online LP, which postulates a unique LP solution, is not a requirement in our approach. We introduce an innovative method to identify and consistently apply one optimal LP basis, even when multiple bases exist. Aligning with a single optimal basis—as opposed to requiring uniqueness—affords the same benefits and leads to logarithmic regret.
To delve into details, our algorithm enhances performance through two new components. The first is a systematic procedure to pinpoint a particular optimal LP basis from potentially several (outlined in \Cref{alg:Idenbasis}), diverging from the traditional literature that presupposes a sole optimal solution and basis. The second component is a resolving procedure that capitalizes on the structure of the identified optimal basis. Once an optimal basis is selected, we can determine the corresponding optimal solution by solving a set of linear equations, which involves only the non-zero basic variables and the active constraints. This resolving process transforms the algorithmic idea of resolving an LP, as seen in previous works \citep{agrawal2014dynamic, bumpensanti2020re, li2022online, vera2021bayesian, li2021symmetry}, into solving specific linear equations based on the chosen optimal basis (detailed in \Cref{alg:Twophase}).
Previous research relies on the non-degeneracy assumption to stabilize the optimal basis throughout the resolving process. However, our two innovative elements enable our algorithm to consistently adhere to one optimal basis, thus removing the non-degeneracy assumption. This not only simplifies the algorithm's execution (from resolving LP to resolving linear equations) but also extends its applicability, freeing it from assumptions that have traditionally limited the scope of online LP techniques.

We finally discuss the dependency of our sample complexity bound on problem parameters other than $\veps$. We restrict to the MDP context without resource constraints. Denote by $\mathcal{S}$ the state set and $\mathcal{A}$ the action set. We show a sample complexity bound of $O\left( \frac{|\mathcal{S}||\mathcal{A}|}{(1-\gamma)^4\cdot\Delta}\cdot\frac{\log^2(1/\veps)}{\veps}\right)$, where $\Delta$ is the constant that represents the hardness of the underlying problem instance.
Compared to the optimal worst-case sample complexity $O\left( \frac{|\mathcal{S}|\cdot|\mathcal{A}|}{(1-\gamma)^3\cdot\veps^2} \right)$ that is achieved in a series of work \citep{sidford2018near, wainwright2019variance, wang2020randomized, agarwal2020model, he2021nearly}, our bound has a worse dependency over $1-\gamma$. This is due to our algorithm being LP-based and the dimension of the LP (can be bounded by $1-\gamma$) will influence our final bounds. However, our bound enjoys a better dependency in terms of $\veps$. For the general CMDP problem, our bound will depend additionally on the conditional number of the constraint matrix in the LP formulation, which is a byproduct of the resolving LP heuristics \citep{vera2021bayesian, li2021symmetry}. However, our $\tilde{O}(1/\veps)$ sample complexity bound depends polynomially on other parameters including $|\mathcal{S}|$, $|\mathcal{A}|$, $1-\gamma$, and the number of constraints. 

\noindent\textbf{Extension to the finite horizon episodic setting}. We extend our approaches and results to the finite horizon setting, where there are multiple periods in a horizon and the horizon can repeated by multiple times to learn the optimal policy. Note that the optimal policy under the finite horizon setting can still be written as a LP. We directly extend the online LP approach to handle the finite horizon setting and obtain similar results. To be specific, we show that in order to obtain an $\veps$-approximate policy where we approximate the optimal reward by a $\eps$ gap and allows at most a $\veps$ constraint violation, we need at most $\tilde{O}(1/\veps)$ number of samples. The details are given in \Cref{sec:Finitehorizon}.

\noindent\textbf{Extension to the off-policy learning settings}. We consider more involved sampling procedure beyond assuming generative model. Note that in practice, the agent usually interacts with an offline simulator (offline learning setting), where there is a fixed historical data set which may be collected using some other algorithms. Different from the generative model where we can obtain information for arbitrary states and actions, the offline learning settings put further restrictions on the freedom of obtaining samples and require additional considerations when designing algorithms. However, we show that our approach can still be directly extended to handle the offline learning settings, and we are also able to derive the similar $\tilde{O}(1/\veps)$ sample complexity. The details are presented in \Cref{sec:offpolicy}.

\noindent\textbf{Extension to the on-policy learning settings}. We consider a more involved sampling procedure when the agent interacts with the true environment (online learning setting), where the agent can only obtain the information regarding the state it is currently visiting and the action it is currently taking. Different from the off-policy learning setting where we can use an additional behavior policy to collect samples, which could be different from the target policy, in the on-policy learning setting, we need to collect samples using the output generated by our algorithm. Therefore, our algorithm not only needs to approximate the optimal policy, but also needs to enjoys the property to do exploration automatically by visiting the desired states as much as possible. We further modify our algorithm and show that our approach can still handle the on-policy learning setting and achieve a similar $\tilde{O}(1/\veps)$ sample complexity. The details are presented in \Cref{sec:onpolicy}.

In summary, our work presents a new algorithm for CMDP problems to achieve the problem-dependent logarithmic regret or $\tilde{O}(1/\veps)$ sample complexity bound, for the first time in the literature. Firstly, we have pioneered the application of the online LP framework to analyze CMDP problems. Our algorithm operates in the primal space, in contrast to the primal-dual framework associated with worst-case guarantees (e.g. \cite{ding2021provably}). Secondly, our novel approach to online LP results in logarithmic regret without the non-degeneracy assumption, expanding its applicability and providing new insights into both CMDP and online LP problems. We finally explore extensions to the finite horizon setting, as well as the online and offline learning settings, and obtain similar instance-dependent guarantees. These extensions demonstrate the generality of our approach.

\subsection{Further Related Work}
Deriving the (logarithmic) problem-dependent guarantees is a long-standing topic in online decision-making \citep{lai1985asymptotically, graves1997asymptotically}. Notably, there has been a recent trend of deriving (logarithmic) problem-dependent guarantees for reinforcement learning problems, where a series of work \citep{zanette2019tighter, simchowitz2019non, foster2020instance, dann2021beyond, bhandari2021linear, wagenmaker2022beyond, wagenmaker2023instance, xu2024towards, duan2024taming, chen2024landscape} has proposed various ways to characterize the hardness of the problem instance. A prevalent way to characterize the hardness is to define a constant that represents the policy sub-optimality gap between the optimal policy and a policy that takes a sub-optimal action given a state. Then, the logarithmic guarantees depend on this constant. Compared to the previous approaches and guarantees, our guarantees adopt this idea of policy sub-optimal gap. Moreover, our approach can handle resource or safety constraints that need to be satisfied by the policy. In this way, our approach generalizes to multi-objective (safe) RL problems. Safe reinforcement learning problems have been investigated extensively under various settings in a series of work (e.g. \cite{liu2021learning}), where the $O(\sqrt{T})$ regret bounds or $O(1/\veps^2)$ sample complexity bounds are proved.


Our method is motivated by the online LP/resource allocation literature. The online LP problem \citep{molinaro2014geometry, agrawal2014dynamic, gupta2014experts} covers a wide range of applications through different ways of specifying the underlying LP, including secretary problem \citep{ferguson1989solved}, online knapsack problem \citep{arlotto2020logarithmic}, quantity-based network revenue management problem \citep{jasin2012re}, network routing problem \citep{buchbinder2009online}, matching problem \citep{mehta2007adwords}, etc. Notably, the problem has been studied under either (i) the stochastic input model where the coefficient in the objective function, together with the corresponding column in the constraint matrix is drawn from an unknown distribution $\mathcal{P}$, or (ii) the random permutation model where they arrive in a random order \citep{molinaro2014geometry, agrawal2014dynamic, kesselheim2014primal, gupta2014experts}. With an additional non-degeneracy assumption, the logarithmic regret has been derived for the quantity-based network revenue management problem \citep{jasin2012re, jasin2014reoptimization}, the online LP problem \citep{li2022online}, and the more general convex allocation problem \citep{ma2022optimal}. The non-degeneracy assumption has been relaxed in a series of work \citep{bumpensanti2020re, vera2021bayesian, jiang2022degeneracy, wei2023constant, ao2024online, banerjee2024good} under various settings.


\subsection{Roadmap}

In \Cref{sec:Infinitedis}, we describe the formal problem formulation and introduce the notations that we use in the paper. In \Cref{sec:reformulation}, we present the linear programming reformulation of our problem and we give a general overview of our approach to illustrate the main idea. Then, in \Cref{sec:Estimates}, we describe how to construct the estimates of the LP reformulation of the CMDP problems, which will be used in our algorithm. The formal algorithm is presented in \Cref{sec:algorithm} and the corresponding theoretical analysis is conducted in \Cref{sec:analysis}. Our approaches are extended to the finite-horizon episodic setting, off-policy setting, and the on-policy setting in \Cref{sec:extensions}, where we obtain the similar instance-dependent guarantees.

\section{Preliminaries}\label{sec:Infinitedis}
We consider a CMDP problem with a finite set of states $\mathcal{S}=\{1,2,\dots, |\mathcal{S}|\}$ and a finite set of actions $\mathcal{A}=\{1,2,\dots, |\mathcal{A}|\}$. We denote by $\gamma\in(0,1)$ a discount factor. We also denote by $P: \mathcal{S}\times\mathcal{A}\rightarrow \mathcal{D}(\mathcal{S})$ the probability transition kernel of the CMDP, where $\mathcal{D}(\mathcal{S})$ denotes a probability measure over the state space $\mathcal{S}$. Then, $P(s'|s, a)$ denotes the probability of transiting from state $s$ to state $s'$ when the action $a$ is executed. The initial distribution over the states of the CMDP is denoted by $\mu_1$.

There is a \textit{stochaastic} reward function $r: \mathcal{S}\times\mathcal{A}\rightarrow\mathcal{D}[0,1]$ and $K$ \textit{stochastic} cost functions $c_k: \mathcal{S}\times\mathcal{A}\rightarrow\mathcal{D}[0,1]$ for each $k\in[K]$. We also denote by $\hat{r}(s,a)=\mathbb{E}[r(s,a)]$ for each $(s,a)$ and $\hat{c}_k(s,a)=\mathbb{E}[c_k(s,a)]$ for each $(s,a)$, for each $k$. For any Markovian policy $\pi$, where the action of $\pi$ depends only on the current state and the action of $\pi$ is allowed to be randomized, we denote by $V_r(\pi, \mu_1)$ the infinite horizon discounted reward of the policy $\pi$, with the formulation of $V_r(\pi, \mu_1)$ given below:
\begin{equation}\label{eqn:Disreward}
V_r(\pi, \mu_0)=\mathbb{E}\left[ \sum_{t=0}^{\infty}\gamma^t\cdot r(s_t, a_t)\mid \mu_1 \right],
\end{equation}
where $(s_t, a_t)$ is generated according to the policy $\pi$ and the transition kernel $P$ with the initial state distribution $\mu_1$.
For each $k\in[K]$, the infinite horizon discounted cost of the policy $\pi$ is denoted by $V_k(\pi, \mu_1)$, and the following constraint needs to be satisfied by the policy $\pi$,
\begin{equation}\label{eqn:Discost}
V_k(\pi, \mu_1)=\mathbb{E}\left[ \sum_{t=0}^{\infty}\gamma^t\cdot c_k(s_t, a_t)\mid \mu_1 \right] \leq \alpha_k, ~~\forall k\in[K].
\end{equation}
To solve the CMDP problem, we aim to find an optimal Markovian policy, denoted by $\pi^*$, that maximizes the reward in \eqref{eqn:Disreward} while satisfying the cost constraint \eqref{eqn:Discost} for each $k\in[K]$, with $\alpha_k\in\left[0,\frac{1}{1-\gamma}\right]$ being a pre-specified value for each $k\in[K]$. Importantly, we assume that the reward function $r$, the cost functions $\{c_k\}_{k=1}^K$, and the transition kernel $P$, are all \textbf{unknown} to the decision maker. Our goal is to obtain a policy $\pi$ that approximates the optimal policy $\pi^*$ with as few samples as possible. We now describe the sampling procedure and present the performance measure of our policy.

We assume the existence of a stylized \textit{generative model} $\mathcal{M}$. The model $\mathcal{M}$ satisfies the following condition.
\begin{assumption}\label{assump:1}
For each state and action pair $(s,a)$, we can query the model $\mathcal{M}$ to obtain an observation of $r(s,a)$, $c_k(s,a)$ for each $k\in[K]$, and the new state $s'\in\mathcal{S}$, where the transition from $s$ to $s'$ follows the probability kernel $P(s'|s,a)$ independently.
\end{assumption}
In reinforcement learning for CMDP problems, querying the generative model $\mathcal{M}$ can be costly. Therefore, it is desirable to query the model $\mathcal{M}$ as less as possible, while guaranteeing the near optimality of the approximate policy. To this end, we study two measures of the performance of our policy. One is \textit{regret}, i.e., conditioning on we can query the model $\mathcal{M}$ at most $N$ times, we aim to derive a policy $\pi$ and bound
\begin{equation}\label{eqn:regret}
\text{Regret}_r(\pi, N):= V_r(\pi^*, \mu_1) - V_r(\pi, \mu_1)\text{~~and~~}\text{Regret}_k(\pi, N):=  V_k(\pi, \mu_1) - \alpha_k,~\forall k\in[K].
\end{equation}
The other one is \textit{sample complexity}, i.e., for any $\veps$, we aim to find an $\veps$-accurate policy $\pi$ such that
\begin{equation}\label{eqn:complexity}
V_r(\pi^*, \mu_1) - V_r(\pi, \mu_1)\leq\veps\text{~~and~~}V_k(\pi, \mu_1) - \alpha_k\leq\veps,~\forall k\in[K],
\end{equation}
with as few samples as possible. Indeed, these two measures can be transferred to each other, and in our analysis, we derive the sample complexity bound from the regret bound.

Note that there are other formulations of the constraints in CMDP, for example,
\begin{equation}\label{eqn:Discost2}
V_k(\pi, \mu_1)=\mathbb{E}\left[ \sum_{t=0}^{\infty}\gamma^t\cdot c_k(s_t, a_t)\mid \mu_1 \right] \geq \lambda_k, ~~\forall k\in[K].
\end{equation}
in a series of work that studies safe reinforcement learning. However, the formulation \eqref{eqn:Discost2} can be transferred from the formulation \eqref{eqn:Discost}. One can set $\alpha_k=\frac{1}{1-\gamma}-\lambda_k$ for each $k\in[K]$, and it is easy to see that the two inequalities are equivalent to each other,
\[
\mathbb{E}\left[ \sum_{t=0}^{\infty}\gamma^t\cdot c_k(s_t, a_t)\mid \mu_1 \right] \geq \lambda_k \Leftrightarrow \mathbb{E}\left[ \sum_{t=0}^{\infty}\gamma^t\cdot (1-c_k(s_t, a_t))\mid \mu_1 \right] \leq \alpha_k.
\]
Therefore, we can equivalently use formulation \eqref{eqn:Discost} with the cost function defined as $1-c_k$ for each $k\in[K]$.

\subsection{Notations}
Throughout the paper, we use $n\in [N]$ to denote the time step. For the infinite horizon discounted problem in \Cref{sec:Infinitedis}, we can query the generative model $\mathcal{M}$ for each $(s,a)\in\mathcal{S}\times\mathcal{A}$ for one time, for a time step $n\in[N]$. 
For a matrix $A$, we denote by $\mathcal{J}$ a row index set and $\mathcal{I}$ a column index set. Then, $A(:, \mathcal{I})$ denotes a sub-matrix of $A$ with columns drawn from the index set $\mathcal{I}$, and $A(\mathcal{J}, :)$ denotes a sub-matrix of $A$ with rows drawn from the index set $\mathcal{J}$. Similarly, $A(\mathcal{J}, \mathcal{I})$ denotes a sub-matrix of $A$ with rows drawn from $\mathcal{J}$ and columns drawn from $\mathcal{I}$. For any vector $\bm{\beta}$, we use $\bm{\beta}_{\mathcal{J}}$ to denote a sub-vector of $\bm{\beta}$ with element drawn from the index set $\mathcal{J}$.

\section{Problem Reformulation and Overview of Our Approach}\label{sec:reformulation}

The infinite horizon discounted setting described in \Cref{sec:Infinitedis} admits a linear programming reformulation. To be specific, due to the existence of the constraints, the optimal policy of a CMDP can be randomized policies (e.g. \cite{altman1999constrained}), where it is optimal to take a stochastic action given the current state. Therefore, it is convenient to represent a policy through the \textit{occupation measure}, which gives us the desired linear programming reformulations of the CMDP problems.

For the infinite horizon discounted problem, the occupancy measure is defined as $q_{\pi}(s, a)$ for any state $s$, action $a$, and policy $\pi$. Note that $q_{\pi}(s, a)$ represents the total expected discounted time spent on the state-action pair $(s,a)$, under policy $\pi$, multiplied by $1-\gamma$. Then, following \cite{altman1999constrained}, the optimal policy (and the optimal occupancy measure) can be obtained from the following linear programming.
\begin{subequations} \label{lp:Infinite}
\begin{align}
\VI=~\max\ &\sum_{s\in\mathcal{S}}\sum_{a\in\mathcal{A}} \hat{r}(s,a)\cdot q(s,a)& \label{eqn:InfiObj}
\\ \mathrm{s.t.\ }&\sum_{s\in\mathcal{S}}\sum_{a\in\mathcal{A}} \hat{c}_k(s,a)\cdot q(s,a)\leq \alpha_k &\forall k\in[K] \label{eqn:Inficonstraint1}
\\ &\sum_{s'\in\mathcal{S}}\sum_{a\in\mathcal{A}} q(s',a)\cdot (\delta_{s,s'}-\gamma\cdot P(s|s',a))=(1-\gamma)\cdot \mu_1(s) &\forall s\in\mathcal{S} \label{eqn:Inficonstraint2}
\\ & q(s,a) \geq 0 &\forall s\in \mathcal{S}, a\in \mathcal{A}, \label{eqn:Inficonstraint3}
\end{align}
\end{subequations}
where $\delta_{s,s'}=\bI_{s=s'}$, and $\mu_1(s)$ denotes the probability for the first state to be realized as $s\in\mathcal{S}$ following the initial distribution $\mu_1$. Note that for an optimal solution $\{q^*(s,a)\}_{\forall s\in\mathcal{S}, \forall a\in\mathcal{A}}$ to \eqref{lp:Infinite}, the corresponding optimal policy $\pi^*$ will be
\begin{equation}\label{eqn:010901}
P(\pi^*(s)=a) = \left\{
\begin{aligned}
&\frac{q^*(s,a)}{\sum_{a'\in\mathcal{A}}q^*(s,a')}, && \text{if~}\sum_{a'\in\mathcal{A}}q^*(s,a')>0\\
&1/|\mathcal{A}|, &&\text{if~}\sum_{a'\in\mathcal{A}}q^*(s,a')=0,
\end{aligned}\right.
\end{equation}
where $P(\pi^*(s)=a)$ denotes the probability for the policy $\pi^*$ to take the action $a\in\mathcal{A}$ given the state $s\in\mathcal{S}$. In fact, when $\sum_{a'\in\mathcal{A}}q^*(s,a')=0$, we can take an arbitrary action.
In what follows, we rely on the linear programming formulation \eqref{lp:Infinite} to derive our results.

\subsection{Characterization of Instance Hardness}\label{sec:characterization}
We illustrate our approach to solving the CMDP problems and present the formal algorithms. Our approach is based on the linear programming reformulations presented in \eqref{lp:Infinite}. We now rewrite the LP \eqref{lp:Infinite} into the following standard formulation to proceed with our illustration.
\begin{equation}\label{lp:standard3}
\begin{aligned}
V=& \max&&\hat{\bm{r}}^\top \bm{q} \\
&~~ \mbox{s.t.} &&C\bm{q}\leq\bm{\alpha}\\
& && B\bm{q}=\bm{\mu}\\
& &&\bm{q}\geq0.
\end{aligned}
\end{equation}
For the infinite horizon discounted problem described in \Cref{sec:Infinitedis}, the matrix $B\in\mathbb{R}^{|\mathcal{S}|\times |\mathcal{S}||\mathcal{A}|}$ and the element at $s$-th row and $(s',a')$th column of the matrix $B$ is $(\delta_{s,s'}-\gamma\cdot P(s|s',a'))$. The matrix $C\in\mathbb{R}^{K\times|\mathcal{S}|}$, and the element at the $k$-th row and $(s,a)$-th column of the matrix $C$ is $\hat{c}_k(s,a)$. The vector $\bmu\in\mathbb{R}^{|\mathcal{S}|}$ and the $s$-th element of $\bmu$ is $(1-\gamma)\cdot \mu_1(s)$, for $s\in\mathcal{S}$.

Note that one crucial step for achieving problem-dependent bounds is to characterize the hardness of the underlying problem instance and define a gap that separates the optimal policies from the others. For multi-arm-bandit problem, the characterization of the hardness can be the gap between the optimal arm and the best sub-optimal arm (e.g. \cite{lai1985asymptotically}). For reinforcement learning problem, the characterization of the hardness can also be the gap between the optimal policy and the best sub-optimal policy (e.g. \cite{auer2008near}). As long as this separation gap is a positive constant, denoted by $\Delta$, separating the optimal policy from the others with a probability at least $1-\epsilon$ would require samples at most $\frac{1}{\Delta}\cdot\log(1/\epsilon)$, for an arbitrary $\eps$, which finally implies the instance-optimal sample complexity bound.

\begin{figure}
  \centering
  \includegraphics[width=0.35\textwidth]{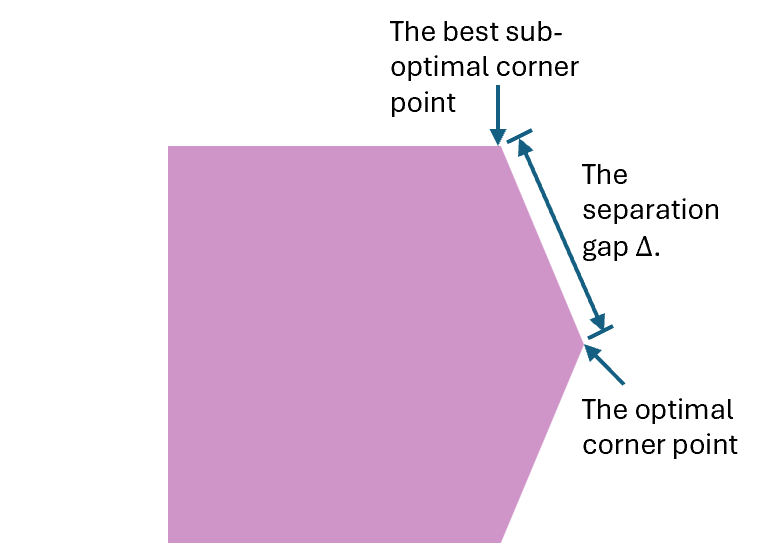}
  \caption{A graph illustration of the hardness characterization via LP basis, where the shaded area denotes the feasible region for the policies. }\label{Fig:01}
\end{figure}

For CMDP problem, characterizing the hardness of the problem instance can be hard. Based on LP \eqref{lp:standard3}, we know that a feasible policy corresponds to a feasible solution and the sub-optimal solution can be arbitrarily close to the optimal solution since the feasible set is ``continuous''. Therefore, there is no direct way to identify a positive gap between the optimal policies and the sub-optimal ones. However, from standard LP theory, we know that one \textbf{corner point} of the feasible region must be one optimal solution. Therefore, we can simply focus on the corner points when solving LP \eqref{lp:standard3} and we define the gap as \textit{the distance between the optimal corner point and the sub-optimal corner point}, as illustrated in \Cref{Fig:01}. As we will show later, the LP reformulation \eqref{lp:standard3} and such a characterization of hardness via corner points will inspire our entire approach.

Since the problem hardness is characterized via corner points, it is essential to provide further characterization of the corner points. Note that in LP theory, the corner point is called \textit{basic solution} and can be represented by \textit{LP basis}, which involves the set of basic variables that are allowed to be non-zero, and the set of active constraints that are binding under the corresponding basic solution.
For our approach, we identify one optimal basic solution and then stick to this optimal basic solution.
However, there can be multiple optimal bases of the LP \eqref{lp:standard3}. We characterize one optimal basis and we exploit its structure during the entire horizon.
We first observe the following result regarding an optimal basis to the LP \eqref{lp:standard3}, which follows from standard LP theory. The proof of Lemma \ref{lem:InfiBasis} is relegated to \Cref{pf:Lemma1} for completeness.

\begin{lemma}\label{lem:InfiBasis}
Denote by $b$ the number of rows in the matrix $B$. Then,
there exists subsets $\mathcal{J}_1^*\subset [K]$, $\mathcal{J}_2^*\subset\mathcal{S}$, and a subset $\mathcal{I}^*\subset\mathcal{S}\times\mathcal{A}$ with $m=|\mathcal{J}_1^*|+|\mathcal{J}_2^*|=|\mathcal{I}^*|$ such that there exists an optimal solution $\bm{q}^*$ to LP \eqref{lp:standard3} with the elements of $\bm{q}^*$ in the index set $\mathcal{I}^*$ are non-zero and the optimal solution $\bm{q}^*$ can be uniquely determined as the solution to the linear system
\begin{subequations}\label{eqn:Lsystem}
\begin{align}
&C(\mathcal{J}^*_1,\mathcal{I}^*)\bm{q}_{\mathcal{I}^*}=\bm{\alpha}_{\mathcal{J}^*},\\
&B(\mathcal{J}^*_2,\mathcal{I}^*)\bm{q}_{\mathcal{I}^*}=\bmu, \\
& \bm{q}_{\mathcal{I}^{*c}}=0.
\end{align}
\end{subequations}
with $\mathcal{I}^{*c}$ being the complementary set of the index set $\mathcal{I}^*$.
\end{lemma}

\begin{remark}\label{remark:1}
The set $\mathcal{I}^*$ and $\mathcal{J}^*=(\mathcal{J}^*_1, \mathcal{J}^*_2)$ described in \Cref{lem:InfiBasis} may not strictly equal to the optimal basis of the LP $V$, but are enough to characterize the optimal basis. To be more precise, the set $\mathcal{I}^*$ actually contains all non-zero basic variables and the set $\mathcal{J}^*$ gives the supporting constraints for these non-zero basic variables such that their value can be uniquely determined by solving the linear equation in \eqref{eqn:Lsystem}. Note that when the LP $V$ is degenerate, it can happen that one basic variable can be $0$ which is not contained in $\mathcal{I}^*$. However, this will not influence the optimality of $\mathcal{I}^*$ and $\mathcal{J}^*$ since we only need to compute the value of the non-zero part of basic variables. Therefore, in what follows, we refer to $\mathcal{I}^*$ and $\mathcal{J}^*$ as an optimal basis of the LP $V$.
\end{remark}

\subsection{Overview of Our Approach}

In the first part, we aim to identify the set $\mathcal{I}^*$ and $\mathcal{J}^*$, which characterizes one optimal basis of LP \eqref{lp:Infinite}. By doing so, we also obtain the optimal actions to take given each state (and each period), by noting the relationship between the optimal actions and the optimal solution (occupancy measure) in \eqref{eqn:010901}. In the second part, we aim to obtain the optimal solution (occupancy measure) by solving the linear equations \eqref{eqn:Lsystem}. However, since the parameters of the linear equations \eqref{eqn:Lsystem} are unknown, we cannot directly solve \eqref{eqn:Lsystem}. Instead, we need to use historical samples to construct estimates of the parameters of \eqref{eqn:Lsystem} and gradually approximate the optimal solution.
To this end, we seek tools from the online packing/linear programming literature (e.g. \cite{agrawal2014dynamic, li2022online, vera2021bayesian, bumpensanti2020re}), where we interpret the right-hand side (RHS) of the linear equations \eqref{eqn:Lsystem} as resource capacities. Note that a crucial element in deriving the logarithmic or better problem-dependent regret in the previous literature is to let our algorithm be \textit{adaptive} to the remaining resource capacities in real-time. We adopt this adaptive resolving idea in our algorithm. 

In the remaining sections, we first describe how our LP-based approach implies a characterization of the ``hardness'' of the underlying problem instance.
We then describe how to construct estimates of unknown parameters for the CMDP problem and we present our formal algorithms. We show how to identify one optimal basis, and how to approximate the optimal solution with an adaptive constraint design.


\section{Construct Estimates in CMDP}\label{sec:Estimates}
In this section, we describe how to construct estimates for LP \eqref{lp:standard3}. To this end, for a round $N_0$, we denote by $\mathcal{F}_{N_0}$ the filtration of all the information collected up to round $N_0$. Then, we denote by $\bar{C}_{N_0}$ (resp. $\bar{B}_{N_0}$) an estimate of the matrix $C$ (resp. $B$), constructed using the information in the set $\mathcal{F}_{N_0}$. We also denote by $\bar{\bm{r}}_{N_0}$ and estimate of $\hat{\bm{r}}$ constructed from the information in the set $\mathcal{F}_{N_0}$. Then, similar to \cite{efroni2020exploration}, we can use the following LP to obtain an estimate of $V$ \eqref{lp:standard3}.
\begin{equation}\label{lp:UB}
\begin{aligned}
\bar{V}_{N_0}=& \max&&(\bar{\bm{r}}_{N_0})^\top \bm{q} \\
&~~ \mbox{s.t.} &&\bar{C}_{N_0}\bm{q}\leq \bm{\alpha}+\lambda_{N_0}\\
& && \bar{B}_{N_0}\bm{q}\leq\bm{\mu}+\lambda_{N_0}\\
& && \bar{B}_{N_0}\bm{q}\geq\bm{\mu}-\lambda_{N_0}\\
& &&\bm{q}\geq0,
\end{aligned}
\end{equation}
with $\lambda_{N_0}$ being a parameter that we specify later. 
To bound the estimation gap, it is useful to bound the optimal dual solution to \eqref{lp:standard3}. To this end, we adopt the approach in \cite{hiriart1996convex, nedic2009subgradient} that utilizes Slater's condition, which is imposed as an assumption below.
\begin{assumption}\label{assump:Slater}
There exists a policy $\bar{\pi}$ such that all the resource constraints are satisfied strictly. In other words, there exists an occupancy measure $\bar{\bm{q}}$ such that $B\bar{\bm{q}}=\bmu$ and $C\bar{\bm{q}}<\bm{\alpha}$. In fact, for each state $s\in\mathcal{S}$, there exists a null action that consumes no resource.
\end{assumption}
The Slater point $\bar{\bm{q}}$ can be set as the policy that takes the null action given each state.
The estimation error will be related to the gap between the Slater point $\bar{\bm{q}}$ and the optimal point $\bm{q}^*$. For an arbitrary $\eps$, we then define the lower gap as
\begin{equation}\label{def:Lowergap}
    \Gap_1(N_0,\eps)\geq V-\bar{V}_{N_0}
\end{equation}
and the upper gap as
\begin{equation}\label{def:Uppergap}
    \Gap_2(N_0, \eps)\geq\bar{V}_{N_0}-V
\end{equation}
with both inequalities \eqref{def:Lowergap} and \eqref{def:Uppergap} hold with probability at least $1-\eps$.

\subsection{Estimates for the Infinite Horizon Discounted Problem}
We define the following empirical estimates over the LP parameters.
\begin{equation}\label{eqn:empirical}
\bar{r}_{N_0}(s,a)=\frac{\sum_{n=1}^{N_0}r^n(s,a)}{N_0}, \bar{c}_{k, N_0}(s,a)=\frac{\sum_{n=1}^{N_0}c^n_{k}(s,a)}{N_0}, \text{~and~} \bar{P}_{N_0}(s'|s,a)=\frac{\sum_{n=1}^{N_0}\bI_{s^n(s,a)=s'}}{N_0},
\end{equation}
where $r^n(s,a)$ denotes the $n$-th observation of the reward, and $c^n_{k}(s,a)$ denotes the $n$-th observation of the $k$-th cost, and $s^n(s,a)$ denotes the $n$-th observation of the state transition for the state-action pair $(s,a)$, for $n\in[N_0]$.
We denote by
$\Rad(N_0, \eps)=\sqrt{\frac{\log(2/\eps)}{2N_0}}$.
Following the standard Hoeffding's inequality, we know that $|\bar{r}_{N_0}(s,a)-\hat{r}(s,a)|$, $|\bar{c}_{k, N_0}(s,a)-\hat{c}_k(s,a)|$, and $|\bar{P}_{N_0}(s'|s,a)-P(s'|s,a)|$ are all upper bounded by $\Rad(N_0, \eps)$ with probability at least $1-\eps$.
Then, we use the following LP with empirical estimates to approximate the value of $V$, as given in \eqref{lp:UB}.
Here, $\bar{C}_{N_0}\in\mathbb{R}^{K\times |\mathcal{S}||\mathcal{A}|}$, and the element at the $k$-th row and $(s,a)$-th column of the matrix $\bar{C}_{N_0}$ is $\bar{c}_{k, N_0}(s,a)$. $\bar{B}_{N_0}\in\mathbb{R}^{|\mathcal{S}|\times |\mathcal{S}||\mathcal{A}|}$, and the element at $s$-th row and $(s',a')$-th column of the matrix $\bar{B}_{N_0}$ is $(\delta_{s,s'}-\gamma\cdot \bar{P}_{N_0}(s|s',a'))$. $\bar{\bm{r}}_{N_0}\in\mathbb{R}^{|\mathcal{S}||\mathcal{A}|}$ with the $(s,a)$-th element being $\bar{r}_{N_0}(s,a)$.
We have the following result, where the proof is relegated to \Cref{pf:Lemma2}.
\begin{lemma}\label{lem:LUbound}
As long as $\lambda_{N_0}=\Rad(N_0,\eps)$,
the following inequality
\begin{equation}\label{eqn:Vgap1}
\begin{aligned}
V \leq\bar{V}_{N_0}+\Rad(N_0, \eps)\leq V+\Rad(N_0, \eps)+\frac{2\Rad(N_0,\eps)}{\min_{k\in[K]}\{\alpha_k\}}\cdot\left(1+\frac{|\mathcal{S}|}{1-\gamma} \right)+\frac{\Rad^2(N_0,\eps)}{\min_{k\in[K]}\{\alpha_k\}}\cdot\left(|\mathcal{S}|+\frac{|\mathcal{S}|^2}{1-\gamma} \right).
\end{aligned}
\end{equation}
holds with probability at least $1-(K|\mathcal{S}||\mathcal{A}|-|\mathcal{S}|^2|\mathcal{A}|)\cdot\eps$.
\end{lemma}
Therefore, we can simply set
\begin{equation}\label{eqn:021707}
    \Gap_1(N_0,\eps)=\Rad(N_0,\eps)
\end{equation}
and
\begin{equation}\label{eqn:021708}
\Gap_2(N_0,\eps)=\frac{2\Rad(N_0,\eps)}{\min_{k\in[K]}\{\alpha_k\}}\cdot\left(1+\frac{|\mathcal{S}|}{1-\gamma} \right)+\frac{\Rad^2(N_0,\eps)}{\min_{k\in[K]}\{\alpha_k\}}\cdot\left(|\mathcal{S}|+\frac{|\mathcal{S}|^2}{1-\gamma} \right).
\end{equation}

\section{Algorithm Design}\label{sec:algorithm}
In this section, we present our algorithms for solving the CMDP problem. We first show how to identify one optimal basis of LP \eqref{lp:standard3} and then present the formal algorithm to achieve an instance-dependent guarantee.

\subsection{Optimal Basis Characterization (on True Problem)}
To motivate our formal algorithm that identifies the optimal basis of the LP \eqref{lp:standard3} from samples, we start with identifying one optimal basis assuming that we have the access to the true LP formulation in \eqref{lp:standard3}, which is equivalent to assuming that the number of samples $N_0\rightarrow\infty$ such that $\bar{C}_{N_0} = C$ and $\bar{B}_{N_0} = B$. The optimal basis of the LP \eqref{lp:standard3} can actually be identified by sequentially discarding the sub-optimal actions and the redundant constraints. We present the full algorithm in \Cref{alg:IdenbasisTrue}. We now explain the procedures in \Cref{alg:IdenbasisTrue} and show \Cref{alg:IdenbasisTrue} successfully identify one optimal basis $\mathcal{I}^*$ and $\mathcal{J}^*$.

One key design in our algorithm is to consider the LP $V_{\mathcal{I}}$ defined for a set $\mathcal{I}\subset\mathcal{S}\times\mathcal{A}$ as follows
\begin{equation}\label{lp:Discardtrue}
\begin{aligned}
V_{\mathcal{I}}= \max~~&\hat{\bm{r}}^\top \bm{q} \\
\mbox{s.t.} ~~&C\bm{q}\leq\bm{\alpha}\\
& B\bm{q}=\bm{\mu} \\
& \bm{q}_{\mathcal{I}^c}=0 \\
& \bm{q}\geq0.
\end{aligned}
\end{equation}
Note that if it holds $V=V_{\mathcal{I}}$ for a set $\mathcal{I}$, then it implies that the set $\mathcal{I}$ contains a set of basic variables for the original LP $V$ such that restricting the elements of the decision variable $\bm{q}$ in the set $\mathcal{I}^c$ to be $0$ will not change the LP value. Following this intuition, starting from $\mathcal{I}$ denoting the whole index set, we sequentially delete one element $i$ (denoting one state-action pair $(s,a)$ in the infinite horizon discounted problem) from the set $\mathcal{I}$. Once we detected that $V-V_{\mathcal{I}\backslash\{i\}}>0$, we know that $i$ is a non-zero basic variable and we remain $i$ in the set $\mathcal{I}$. If we detected that $V=V_{\mathcal{I}\backslash\{i\}}$, we know that we can safely drop the element $i$ from the set $\mathcal{I}$ and we will do that permanently, which implies that the dropped elements will never be added back to the set $\mathcal{I}$.
In this way, we can classify all the non-zero basic variables into the set $\mathcal{I}$. This describes the procedure of step 3-8 in \Cref{alg:IdenbasisTrue}.

\begin{algorithm}[ht!]
\caption{Algorithm for identifying one optimal basis on the true problem}
\label{alg:IdenbasisTrue}
\begin{algorithmic}[1]
\State \textbf{Input:} the true LP $V$.
\State Set $\mathcal{I}_0$ to be the whole index set of $\mathcal{S}\times\mathcal{A}$ and set $\mathcal{J}_0=[K]\cup\mathcal{S}$.

\State Initialize $\mathcal{I}=\mathcal{I}_0$ and $\mathcal{J}=\mathcal{J}_0$.

\For{$i\in\mathcal{I}_0$}
\State Let $\mathcal{I}'=\mathcal{I}\backslash\{i\}$.
\State Compute the value of $V_{\mathcal{I}'}$ as in \eqref{lp:Discard}.
\State If $V_{\mathcal{I}'}=V$, then we set $\mathcal{I}=\mathcal{I}'$.
\EndFor
\For{$j\in\mathcal{J}_0$}
\State Let $\mathcal{J}'=\mathcal{J} \backslash \{ j \}$ and let $\mathcal{J}'_1=\mathcal{J}'\cap[K]$, $\mathcal{J}'_2=\mathcal{J}'\cap\mathcal{S}$.
\State Compute the value of $\text{Dual}_{\mathcal{J}', \mathcal{I}}$ as in \eqref{dual:Discard}.
\State If $V=\text{Dual}_{\mathcal{J}', \mathcal{I}}$ and the matrix $\begin{bmatrix}
C(\mathcal{J}'_1, \mathcal{I})\\
B(\mathcal{J}'_2, \mathcal{I})
\end{bmatrix}$ is full rank, then we set $\mathcal{J}=\mathcal{J}'$.
\State \textbf{Break} if $|\mathcal{J}|=|\mathcal{I}|$.
\EndFor
\State \textbf{Output}: the set of indexes $\mathcal{I}$ and $\mathcal{J}$.
\end{algorithmic}
\end{algorithm}

We now aim to obtain the supporting constraints. Denote by $M$ the number of non-zero basic variables detected in \Cref{alg:IdenbasisTrue}. We then denote by $M'$ the number of binding constraints corresponding to the optimal solution given by the optimal basis we detected. Our goal is to identify the \textit{linearly independent} binding constraints such that the non-zero basic variables can be uniquely determined as the solution to $M$ number of linear equations, as described in Lemma \ref{lem:InfiBasis}.
To tell which constraints are redundant and thus can be deleted, we consider the dual program of $V_{\mathcal{I}}$, where $\mathcal{I}$ is determined in steps 3-8 in \Cref{alg:IdenbasisTrue}, and similarly, we test which dual variable can be set to $0$ without influencing the dual objective value. 
For a dual variable index subset $\mathcal{J}=(\mathcal{J}_1, \mathcal{J}_2)$ where $\mathcal{J}_1\subset[K]$ and $\mathcal{J}_2\subset\mathcal{S}$, we consider the dual program as follows.
\begin{subequations}\label{dual:Discard}
\begin{align}
\text{Dual}_{\mathcal{J},\mathcal{I}}= \min~~&\bm{\alpha}^\top \bm{y}+\bmu^\top \bm{z} \\
 \mbox{s.t.}~~ &(C(:,\mathcal{I}))^\top\bm{y}+(B(:,\mathcal{I}))^\top \bm{z}\geq\hat{\bm{r}}_{\mathcal{I}}\label{const:Discard}\\
 &\bm{y}_{\mathcal{J}_1^c}=0, ~\bm{z}_{\mathcal{J}_2^c}=0, \\
&\bm{y}\geq0, \bm{z}\geq-\infty,
\end{align}
\end{subequations}
where $\mathcal{J}_1^c=[K]\backslash\mathcal{J}_1$ and $\mathcal{J}_2^c=\mathcal{S}\backslash\mathcal{J}_2$. 

In step 9-13 of \Cref{alg:IdenbasisTrue}, we shrinkage the size of $\mathcal{J}$ (with initialization $\mathcal{J}=[K]\cup\mathcal{S}$) by checking whether dropping an element $j$ from $\mathcal{J}$, with $\mathcal{J}'=\mathcal{J}\backslash\{j\}$ will make the LP value $\text{Dual}_{\mathcal{J},\mathcal{I}}$ different from $V$. If $\text{Dual}_{\mathcal{J}',\mathcal{I}}> V$, then we know that $j$ is a supporting constraint and we will remain $j$ in the set $\mathcal{J}$. Otherwise, if $\text{Dual}_{\mathcal{J}',\mathcal{I}}= V$ and in addition the matrix $\begin{bmatrix}
C(\mathcal{J}'_1, \mathcal{I})\\
B(\mathcal{J}'_2, \mathcal{I})
\end{bmatrix}$ is full rank, it can be inferred that $j$ is not a supporting constraint and can be safely dropped from the set $\mathcal{J}$. Note that the requirement of the matrix $\begin{bmatrix}
C(\mathcal{J}'_1, \mathcal{I})\\
B(\mathcal{J}'_2, \mathcal{I})
\end{bmatrix}$ being full-rank is in correspondence to the requirement in \Cref{lem:InfiBasis} that the solution to the linear system \eqref{eqn:Lsystem} is uniquely determined. We can finally show that the sets $\mathcal{I}$ and $\mathcal{J}$ output by \Cref{alg:IdenbasisTrue} satisfy the conditions in \Cref{lem:InfiBasis}, as formalized in the following theorem.
\begin{theorem}\label{thm:BasisTrue}
For the true LP $V$, the sets $\mathcal{I}$ and $\mathcal{J}$ output by \Cref{alg:IdenbasisTrue} satisfy the conditions in \Cref{lem:InfiBasis}.
\end{theorem}
The proof of \Cref{thm:BasisTrue} is relegated to \Cref{pf:thmBasisTrue}. We now in the following remark address the main differences between our \Cref{alg:IdenbasisTrue} and other algorithms in the previous literature.

\begin{remark}
The \Cref{alg:IdenbasisTrue} is motivated from the variable-deletion method in \cite{li2021symmetry} but enjoys several key differences. The key difference is that when deciding whether or not to drop an element $i$ from the set $\mathcal{I}$, our \Cref{alg:IdenbasisTrue} will check whether $V=V_{\mathcal{I}\backslash\{i\}}$, while the algorithm in \cite{li2021symmetry} check whether $V=V_{\mathcal{I}_0\backslash\{i\}}$. In other words, if an element $i$ is about to be dropped and determined as being $0$ in an optimal solution, our \Cref{alg:IdenbasisTrue} will drop the element $i$ \textit{permanently} from the set $\mathcal{I}$, while the algorithm in \cite{li2021symmetry} only adds the element $i$ to a set containing non-basic variables without dropping it from $\mathcal{I}_0$. This is one key reason why our \Cref{alg:IdenbasisTrue} works without any assumption on the LP $V$, while the algorithm in \cite{li2021symmetry} requires the LP $V$ to be \textit{non-degenerate}. 

Note that there are also other algorithms existing in the literature that can identify an optimal basis of an LP (e.g. \cite{megiddo1991finding}). However, as illustrated in \Cref{remark:1}, our \Cref{alg:IdenbasisTrue} is not to find exactly an optimal basis, instead, we find the set of non-zero basic variables and the corresponding supporting constraints. When the LP $V$ is degenrate, the basic variables founded by the algorithm in \cite{megiddo1991finding} \textit{can be zero}, which would cause troubles when doing the online learning extension in  . Moreover, our \Cref{alg:IdenbasisTrue} only involves operations such as checking the LP values and the non-singularity of a matrix, which are easy to be extended to the data-driven setting where we only have sample average estimations of the constraint matrix. In general, our \Cref{alg:IdenbasisTrue} is new in the literature and brings advantages as discussed above to handling the problem in our settings.
\end{remark}

\subsection{Characterize One Optimal Basis From Samples}
We now describe how to carry out our \Cref{alg:IdenbasisTrue} when we don't have access to the true LP $V$.  We will use the estimates constructed in the previous \Cref{sec:Estimates} to carry out the procedure.
The formal algorithm is presented in \Cref{alg:Idenbasis}.

\begin{algorithm}[ht!]
\caption{Algorithm for identifying one optimal basis}
\label{alg:Idenbasis}
\begin{algorithmic}[1]
\State \textbf{Input:} set $\mathcal{F}_{N_0}$ containing $N_0$ samples for each $(s,a)\in\mathcal{S}\times\mathcal{A}$ and failure probability $\eps$.
\State Set $\mathcal{I}_0$ to be the whole index set of $\mathcal{S}\times\mathcal{A}$ and set $\mathcal{J}_0=[K]\cup\mathcal{S}$, and compute the value of $\bar{V}_{N_0}$ as in \eqref{lp:UB}.

\State Initialize $\mathcal{I}=\mathcal{I}_0$ and $\mathcal{J}=\mathcal{J}_0$.

\For{$i\in\mathcal{I}_0$}
\State Let $\mathcal{I}'=\mathcal{I}\backslash\{i\}$.
\State Compute the value of $\bar{V}_{\mathcal{I}', N_0}$ as in \eqref{lp:Discard}.
\State If $|\bar{V}_{\mathcal{I}', N_0}-\bar{V}_{N_0}|\leq2\Gap_1(N_0,\eps)+2\Gap_2(N_0,\eps)$, then we set $\mathcal{I}=\mathcal{I}'$.
\EndFor
\For{$j\in\mathcal{J}_0$}
\State Let $\mathcal{J}'=\mathcal{J} \backslash \{ j \}$ and let $\mathcal{J}'_1=\mathcal{J}'\cap[K]$, $\mathcal{J}'_2=\mathcal{J}'\cap\mathcal{S}$.
\State Compute $\text{Dual}_{\mathcal{J}', \mathcal{I}, N_0}$ as in \eqref{lp:Discard}, and the smallest singular value $\bar{\sigma}$ of $\begin{bmatrix}
\bar{C}_{N_0}(\mathcal{J}'_1, \mathcal{I})\\
\bar{B}_{N_0}(\mathcal{J}'_2, \mathcal{I})
\end{bmatrix}$.
\If{$|\bar{V}_{N_0}-\text{Dual}_{\mathcal{J}', \mathcal{I}, N_0}|\leq2\Gap_1(N_0, \eps)+2\Gap_2(N_0, \eps)$ and $\bar{\sigma}\geq |\mathcal{J}'||\mathcal{I}|\cdot\Rad(N_0,\eps)$}
\State set $\mathcal{J}=\mathcal{J}'$.
\EndIf
\State \textbf{Break} if $|\mathcal{J}|=|\mathcal{I}|$.
\EndFor
\State \textbf{Output}: the set of indexes $\mathcal{I}$ and $\mathcal{J}$.
\end{algorithmic}
\end{algorithm}


We now explain why \Cref{alg:Idenbasis} works. The \Cref{alg:Idenbasis} is developed based on \Cref{alg:IdenbasisTrue}. From \Cref{thm:BasisTrue}, we know that the \Cref{alg:IdenbasisTrue} would be able to find the optimal basis $\mathcal{I}^*$ and $\mathcal{J}^*$. Therefore, as long as the \Cref{alg:Idenbasis} follows the same procedure of \Cref{alg:IdenbasisTrue} by dropping exactly the same element $i$ from the set $\mathcal{I}$ and the same element $j$ from the set $\mathcal{J}$ as \Cref{alg:IdenbasisTrue}, we know that \Cref{alg:Idenbasis} finds the optimal basis $\mathcal{I}^*$ and $\mathcal{J}^*$. However, the true LP $V$ is unknown to the \Cref{alg:Idenbasis}. Therefore, the \Cref{alg:Idenbasis} focuses on the sample average estimation $\bar{V}_{\mathcal{I}, N_0}$ of $V_{\mathcal{I}}$, given as follows, as well as the sample average estimation $\text{Dual}_{\mathcal{J},\mathcal{I}, N_0}$ of $\text{Dual}_{\mathcal{J},\mathcal{I}}$, given as follows.
\begin{equation}\label{lp:Discard}
\begin{aligned}
\bar{V}_{\mathcal{I}, N_0}=& \max&&(\bar{\bm{r}}_{N_0})^\top \bm{q} && \text{Dual}_{\mathcal{J},\mathcal{I}, N_0}=\bar{V}_{\mathcal{J}, \mathcal{I}, N_0}= && \max&&(\bar{\bm{r}}_{N_0})^\top \bm{q} \\
&~~ \mbox{s.t.} &&\bar{C}_{N_0}\bm{q}\leq\bm{\alpha}+\lambda_{N_0} &&  &&~~\mbox{s.t.} &&\bar{C}_{N_0}(\mathcal{J}_1,:)\bm{q}\leq \bm{\alpha}_{\mathcal{J}_1}+\lambda_{N_0}\\
& && \|\bar{B}_{N_0}\bm{q}-\bm{\mu}\|_{\infty}\leq \lambda_{N_0} && &&  && \|\bar{B}_{N_0}(\mathcal{J}_2, :)\bm{q}-\bm{\mu}\|_{\infty}\leq \lambda_{N_0}\\
& &&\bm{q}_{\mathcal{I}^c}=0 && && &&\bm{q}_{\mathcal{I}^c}=0\\
& &&\bm{q}\geq0, && && &&\bm{q}\geq0.
\end{aligned}
\end{equation}
where $\mathcal{I}^c$ denotes the complementary set of $\mathcal{I}$, and $\mathcal{J}_1=\mathcal{J}\cap[K]$, $\mathcal{J}_2=\mathcal{J}\cap\mathcal{S}$.

The steps 4-8 in \Cref{alg:Idenbasis} is similar to the steps 4-8 in \Cref{alg:IdenbasisTrue} except that the \Cref{alg:Idenbasis} is applied to the sample average estimation of the LP $V$. Note that in order to characterize an optimal basis of $V$, we need to drop an element $i$ from $\mathcal{I}$ as long as $V=V_{\mathcal{I}'}$ with $\mathcal{I}'=\mathcal{I}\backslash\{i\}$. However, since the value of both $V$ and $V_{\mathcal{I}'}$ are know, we can only use the estimates $\bar{V}_{N_0}$ and $\bar{V}_{\mathcal{I}', N_0}$ to judge whether it holds $V=V_{\mathcal{I}'}$. From \Cref{lem:LUbound}, we know that with a high probability, it holds that 
\[
|V-\bar{V}_{N_0}|\leq \Gap_1(N_0, \eps)+\Gap_2(N_0, \eps)\text{~and~}|V_{\mathcal{I}'}-\bar{V}_{\mathcal{I}', N_0}|\leq \Gap_1(N_0, \eps)+\Gap_2(N_0, \eps).
\]
Therefore, we know that the condition in step 7 of \Cref{alg:Idenbasis} that $|\bar{V}_{\mathcal{I}', N_0}-\bar{V}_{N_0}|\leq2\Gap_1(N_0,\eps)+2\Gap_2(N_0,\eps)$ is a necessary condition of $V=V_{\mathcal{I}'}$ (with a high probability). In order to derive the sufficient condition of $V=V_{\mathcal{I}'}$ (with a high probability), we need to introduce an intrinsic gap between $V$ and all possible $V_{\mathcal{I}'}$, i.e., we define a constant  
\begin{equation}\label{eqn:deltaGap}
\delta_1=\min_{\mathcal{I}}\{V-V_{\mathcal{I}}: V-V_{\mathcal{I}}>0\}
\end{equation}
It is clear to see that as long as the sample size $N_0$ is large enough such that 
$4\Gap_1(N_0,\eps)+4\Gap_2(N_0,\eps)<\delta_1$, the condition $|\bar{V}_{\mathcal{I}', N_0}-\bar{V}_{N_0}|\leq2\Gap_1(N_0,\eps)+2\Gap_2(N_0,\eps)$ would imply that $|V-V_{\mathcal{I}'}|<\delta_1$. Given the definition of $\delta_1$ in \eqref{eqn:deltaGap}, we must have $V=V_{\mathcal{I}'}$. Therefore, we know that with a high probability, the step 7 of \Cref{alg:Idenbasis} is exactly the same as the step 7 of \Cref{alg:IdenbasisTrue} and each element $i\in\mathcal{I}_0$ will be dropped/remained in a same way between \Cref{alg:Idenbasis} and \Cref{alg:IdenbasisTrue}.


To tell which constraints are redundant and thus can be deleted, we consider the estimate of the dual LP of $V_{\mathcal{I}}$, which is given as $\text{Dual}_{\mathcal{J}, \mathcal{I}, N_0}$ in \eqref{lp:Discard}. Then, the step 9-15 of \Cref{alg:Idenbasis} carries out the same procedure as the step 9-14 of \Cref{alg:IdenbasisTrue}. To see this point, we define a constant
\begin{equation}\label{eqn:deltaDual}
\delta_2=\min_{\mathcal{I}, \mathcal{J}}\{\text{Dual}_{\mathcal{J}, \mathcal{I}}-V_{\mathcal{I}}: \text{Dual}_{\mathcal{J}, \mathcal{I}}-V_{\mathcal{I}}>0\}.
\end{equation}
Following the same procedure as above, it is easy to see that when the sample size $N_0$ is large enough such that $4\Gap_1(N_0,\eps)+4\Gap_2(N_0,\eps)<\delta_2$, the condition of $|\bar{V}_{N_0}-\text{Dual}_{\mathcal{J}', \mathcal{I}, N_0}|\leq2\Gap_1(N_0, \eps)+2\Gap_2(N_0, \eps)$ is a sufficient and necessary condition for $V=\text{Dual}_{\mathcal{J}', \mathcal{I}}$, with a high probability. 

Moreover, we can apply the sensitivity of matrix singular values (e.g. Theorem 1 of \citet{stewart1998perturbation}) to show that the gap between the smallest singular value of the matrix $\begin{bmatrix}
\bar{C}_{N_0}(\mathcal{J}'_1, \mathcal{I})\\
\bar{B}_{N_0}(\mathcal{J}'_2, \mathcal{I})
\end{bmatrix}$ and the matrix $\begin{bmatrix}
C(\mathcal{J}'_1, \mathcal{I})\\
B(\mathcal{J}'_2, \mathcal{I})
\end{bmatrix}$ is upper bounded by $|\mathcal{J}'||\mathcal{I}|\cdot\Rad(N_0,\eps)$, with a high probability. Therefore, denote by $\sigma_0$ the smallest possible singular value of the matrix $\begin{bmatrix}
C(\mathcal{J}'_1, \mathcal{I})\\
B(\mathcal{J}'_2, \mathcal{I})
\end{bmatrix}$ for all possible $\mathcal{J}', \mathcal{I}$ such that the matrix $\begin{bmatrix}
C(\mathcal{J}'_1, \mathcal{I})\\
B(\mathcal{J}'_2, \mathcal{I})
\end{bmatrix}$ is full rank, i.e., we define the following constant 
\begin{equation}\label{def:sigma0}
\sigma_0=\min_{\mathcal{I}, \mathcal{J}}\left\{ \sigma_{\min}\left(\begin{bmatrix}
C(\mathcal{J}_1, \mathcal{I})\\
B(\mathcal{J}_2, \mathcal{I})
\end{bmatrix}\right)\text{~with~}\mathcal{J}_1=\mathcal{J}\cap[K], \mathcal{J}_2=\mathcal{J}\cap\mathcal{S}: \sigma_{\min}\left(\begin{bmatrix}
C(\mathcal{J}_1, \mathcal{I})\\
B(\mathcal{J}_2, \mathcal{I})
\end{bmatrix}\right)>0 \right\}
\end{equation}
where $\sigma_{\min}(A)$ denotes the smallest singular value of a matrix $A$.
We can show that as long as the sample size $N_0$ is large enough such that it holds $\sigma_0\geq2|\mathcal{J}'||\mathcal{I}|\cdot\Rad(N_0,\eps)$, the condition that the smallest singular value of the matrix $\begin{bmatrix}
\bar{C}_{N_0}(\mathcal{J}'_1, \mathcal{I})\\
\bar{B}_{N_0}(\mathcal{J}'_2, \mathcal{I})
\end{bmatrix}$ to be lower bounded by $|\mathcal{J}'||\mathcal{I}|\cdot\Rad(N_0,\eps)$ is a sufficient and necessary condition for the matrix $\begin{bmatrix}
C(\mathcal{J}'_1, \mathcal{I})\\
B(\mathcal{J}'_2, \mathcal{I})
\end{bmatrix}$ to be full rank, with a high probability.  

In this way, we know that with a high probability, the step 12 of \Cref{alg:Idenbasis} is exactly the same as the step 12 of \Cref{alg:IdenbasisTrue} and each element $j\in\mathcal{J}_0$ will be dropped/remained in a same way between \Cref{alg:Idenbasis} and \Cref{alg:IdenbasisTrue}.


\subsection{Our Formal Algorithm}\label{sec:Allocation}
We now describe our formal algorithm.
From the output of \Cref{alg:Idenbasis}, we characterize one optimal solution. We denote by $\hat{\mathcal{I}}$ and $\hat{\mathcal{J}}$ the output of \Cref{alg:Idenbasis}. To be specific, we have $\bm{q}^*_{\hat{\mathcal{I}}^c}=0$ and the non-zero elements $\bm{q}^*_{\hat{\mathcal{I}}}$ can be given as the solution to
\begin{equation}\label{eqn:OptQ}
\begin{bmatrix}
C(\hat{\mathcal{J}}_{1}, \hat{\mathcal{I}})\\
B(\hat{\mathcal{J}}_{2}, \hat{\mathcal{I}})
\end{bmatrix}
\cdot\bm{q}^*_{\hat{\mathcal{I}}}=\begin{bmatrix}
\bm{\alpha}_{\hat{\mathcal{J}}_{1}}\\
\bmu_{\hat{\mathcal{J}}_{2}}
\end{bmatrix},
\end{equation}
where $\hat{\mathcal{J}}_{1}=\hat{\mathcal{J}}\cap[K]$ and $\hat{\mathcal{J}}_{2}=\hat{\mathcal{J}}\cap\mathcal{S}$.
However, in practice, both the matrices $C(\hat{\mathcal{J}}_{1}, \hat{\mathcal{I}})$ and $B(\hat{\mathcal{J}}_{2}, \hat{\mathcal{I}})$ are unknown. We aim to use the samples to learn the matrices $C(\hat{\mathcal{J}}_{1}, \hat{\mathcal{I}})$ and $B(\hat{\mathcal{J}}_{2}, \hat{\mathcal{I}})$ such that the $\bm{q}^*_{\hat{\mathcal{I}}}$ can also be well-approximated. 
Our formal algorithm is given in \Cref{alg:Twophase}. The steps 2-3 in \Cref{alg:Twophase} is to use \Cref{alg:Idenbasis} as a subroutine to identify the set $\hat{\mathcal{I}}$ and $\hat{\mathcal{J}}$ that satisfy the conditions in \Cref{lem:InfiBasis} with a high probability. We can show that as long as $N_0\geq N_0'$, where $N_0'$ is a threshold that depends on the problem parameters, \Cref{alg:Idenbasis} correctly obtains the set $\mathcal{I}$ and $\mathcal{J}$ satisfying the conditions in \Cref{lem:InfiBasis}, with a high probability. 


\begin{algorithm}[ht!]
\caption{The Adaptive-resolving Algorithm}
\label{alg:Twophase}
\begin{algorithmic}[1]
\State \textbf{Input:} the number of rounds $N_1$ carrying out \Cref{alg:Idenbasis} and the number of rounds $N_2$ for resolving.
\State Query the generative model for each $(s,a)\in\mathcal{S}\times\mathcal{A}$ by $N_1$ times and obtain the dataset $\mathcal{F}_{N_1}$.
\State Obtain the output $\hat{\mathcal{I}}$ and $\hat{\mathcal{J}}$ from \Cref{alg:Idenbasis} with input $\mathcal{F}_{N_1}$ and $\eps=O(\veps^2/(K\cdot|\mathcal{S}||\mathcal{A}|))$.
\State Let $\hat{\mathcal{J}}_{1}=\hat{\mathcal{J}}\cap[K]$ and $\hat{\mathcal{J}}_{2}=\hat{\mathcal{J}}\cap\mathcal{S}$.
\State Initialize $\mathcal{H}_1=\emptyset$, $\bm{\alpha}^1=N_2\cdot \bm{\alpha}$ and $\bmu^1=N_2\cdot\bmu$.
\For{$n=1,\dots, N_2$}
\State Construct estimates $\bar{C}^n(\hat{\mathcal{J}}_{1}, \hat{\mathcal{I}})$ and $\bar{B}^n(\hat{\mathcal{J}}_{2}, \hat{\mathcal{I}})$ using the sample set $\mathcal{H}_{n}$.
\State Construct a solution $\tilde{\bm{q}}^n$ such that $\tilde{\bm{q}}^n_{\hat{\mathcal{I}}^c}=0$ and $\tilde{\bm{q}}^*_{\hat{\mathcal{I}}}$ is the solution to
\begin{equation}\label{eqn:OptQ2}
\left[\begin{aligned}
&\bar{C}^n(\hat{\mathcal{J}}_{1}, \hat{\mathcal{I}})\\
&\bar{B}^n(\hat{\mathcal{J}}_{2}, \hat{\mathcal{I}})
\end{aligned}\right]
\cdot\tilde{\bm{q}}^n_{\hat{\mathcal{I}}}=\left[\begin{aligned}
&\frac{\bm{\alpha}^n_{\hat{\mathcal{J}}_{1}}}{N_2-n+1}\\
&\frac{\bmu^n_{\hat{\mathcal{J}}_2}}{N_2-n+1}
\end{aligned}\right].
\end{equation}
\State Project $\tilde{\bm{q}}^n$ to the set $\{\bm{q}: \|\bm{q}\|_1\leq 2, \bm{q}_{\hat{\mathcal{I}}^c}=0\}$ to obtain $\bm{q}^n$.
\State For each $(s,a)\in\hat{\mathcal{I}}$, we query the model $\mathcal{M}$ to obtain a sample of the reward $r^n(s,a)$ and the costs $c^n_k(s,a)$ for each $k\in\hat{\mathcal{J}}_1\subset[K]$, as well as the state transition $s^n(s,a)$.
\State Update $\mathcal{H}_{n+1}=\mathcal{H}_n\cup\left\{ r^n(s,a), c^n_k(s,a), s^n(s,a), \forall (s,a)\in\hat{\mathcal{I}}, \forall k\in\hat{\mathcal{J}}_1\right\}$.
\State Denote by $\bm{c}^n(s,a)=(c^n_k(s,a))_{\forall k\in\hat{\mathcal{J}}_{1}}$ and do the update:
\begin{equation}\label{eqn:UpdateAlpha2}
\bm{\alpha}^{n+1}_{\hat{\mathcal{J}}_{1}}=\bm{\alpha}^{n}_{\hat{\mathcal{J}}_{1}}-\sum_{(s,a)\in\hat{\mathcal{I}}}\bm{c}^n(s,a)\cdot q^n(s,a).
\end{equation}
\State Do the update:
\begin{equation}\label{eqn:UpdateMu2}
\bmu^{n+1}_{s'}=\bmu^n_{s'}-\sum_{(s,a)\in\hat{\mathcal{I}}}q^n(s,a)\cdot(\delta_{s',s}-\gamma\bI_{\{s'=s^n(s,a)\}}),~~~\forall s'\in\hat{\mathcal{J}}_{2}.
\end{equation}
where $\bI_{\{s'=s^n(s,a)\}}$ is an indicator function of whether the state transition $s^n(s,a)$ equals $s'$.
\EndFor
\State \textbf{Output:} $\bm{q}^1, \dots, \bm{q}^{N_2}$.
\end{algorithmic}
\end{algorithm}

A crucial element in \Cref{alg:Twophase} (step 8-13) is that we adaptively update the value of $\bm{\alpha}^{n}_{\hat{\mathcal{J}}_{1}}$ and $\bmu^{n}_{\hat{\mathcal{J}}_2}$ as in \eqref{eqn:UpdateAlpha2} and \eqref{eqn:UpdateMu2}. We then use the updated $\bm{\alpha}^{n}_{\hat{\mathcal{J}}_{1}}$ and $\bmu^{n}_{\hat{\mathcal{J}}_{2}}$ to obtain the value of $\tilde{\bm{q}}^n_{\hat{\mathcal{I}}}$ as in \eqref{eqn:OptQ2}. Such an algorithmic design has the following natural interpretation and is the key for us to achieve a logarithmic-level regret. Note that if we adopt a static way to compute $\bm{q}^n$, for example, we simply solve \eqref{eqn:OptQ} with the matrices $C(\hat{\mathcal{J}}_{1}, \hat{\mathcal{I}})$ and $B(\hat{\mathcal{J}}_{2}, \hat{\mathcal{I}})$ replaced by their estimates, then the gap between $\bm{q}^n$ and $\bm{q}^*$ will be at the same order of the estimation error of $C(\hat{\mathcal{J}}_{1}, \hat{\mathcal{I}})$ and $B(\hat{\mathcal{J}}_{2}, \hat{\mathcal{I}})$, which is $O(1/\sqrt{n})$. Thus, the gap between $\frac{\sum_{n=1}^{N_2}\bm{q}^n}{N_2}$ and $\bm{q}^*$ will be at the order of $O(1/\sqrt{N_2})$, which translates into $O(\sqrt{N_2})$ regret bound. Instead, we use $\bm{\alpha}^{n}_{\hat{\mathcal{J}}_{1}}$ and $\bmu^{n}_{\hat{\mathcal{J}}_{2}}$ that are adaptive to the historical information. To be specific, if for one $n\in[N_2]$ and one binding constraint $k\in\hat{\mathcal{J}}_{1}$, the constraint value under the action $\sum_{n'=1}^{n}\bm{q}^{n'}/n$ is below the target $\alpha_k$, we know that $\alpha^n_k/(N_2-n+1)$ is greater than $\alpha_k$, which results in a greater constraint value for $\tilde{\bm{q}}^{n+1}$ (it is in fact $\alpha^n_k/(N_2-n+1)$ beyond the target $\alpha_k$). In this way, any gap between the real-time constraint value and the target will be self-corrected in the next period as we adaptively obtain $\tilde{\bm{q}}^{n+1}$. Therefore, such an adaptive design can re-adjust the possible constraint violation by itself, which results in a regret bound of a lower order. In step 11, we further project $\tilde{\bm{q}}^{n+1}$ to the set $\{\bm{q}\geq0: \|\bm{q}\|_1\leq 2, \bm{q}_{\hat{\mathcal{I}}^c}=0\}$ to obtain $\bm{q}^{n+1}$. This prevents $\tilde{\bm{q}}^n$ from behaving ill when $n$ is not large and the estimates of $C(\hat{\mathcal{J}}_{1}, \hat{\mathcal{I}})$ and $B(\hat{\mathcal{J}}_{2}, \hat{\mathcal{I}})$ are not accurate enough. We can show that when $n$ is large enough, $\tilde{\bm{q}}^{n+1}$ automatically belongs to the set $\{\bm{q}\geq0: \|\bm{q}\|_1\leq 2, \bm{q}_{\hat{\mathcal{I}}^c}=0\}$.

\begin{remark} 
Note that the adaptive resolving step \eqref{eqn:OptQ2} is akin to the idea of resolving an LP in online LP, Bandits with Knapsacks, online allocation problems, where logarithmic or better regret is obtained in a series of work \cite{flajolet2015logarithmic, arlotto2019uniformly, li2022online, li2021symmetry, vera2021bayesian, vera2021online, bumpensanti2020re, jiang2022degeneracy}. However, since we have identified the optimal basis $\mathcal{I}$ and $\mathcal{J}$ and we stick to this optimal basis to avoid the non-degeneracy assumption, resolving the LP would become resolving the linear equations in \eqref{eqn:OptQ2}.
\end{remark}


\section{Theoretical Analysis}\label{sec:analysis}
In this section, we conduct our theoretical analysis. The analysis can be divided into two parts. In the first part, we show that \Cref{alg:Idenbasis} can successfully help us identify one optimal basis to work with. In the second part, we show how to learn the optimal distribution over the optimal basis we have identified.
\subsection{High Probability Bound for Identifying One Optimal Basis}
We now present the theorem showing that \Cref{alg:Idenbasis} indeed helps us identify one optimal basis with a high probability at least $1-\eps$. We specify the value of $\eps$ in the following parts.
\begin{theorem}\label{thm:Infibasis2}
For any $\eps>0$, as long as $N_0$ satisfies the condition
\begin{equation}\label{eqn:N0condition}
\Rad(N_0, \eps) = \sqrt{\frac{\log(2/\eps)}{2N_0}}\leq O\left(\min\left\{ \delta_1, \delta_2, \sigma_0 \right\}\right)
\end{equation}
where $\delta_1$ defined in \eqref{eqn:deltaGap}, $\delta_2$ defined in \eqref{eqn:deltaDual} and $\sigma_0$ defined in \eqref{def:sigma0},
the outputs $\mathcal{I}_{N_0}$ and $\mathcal{J}_{N_0}$ of \Cref{alg:Idenbasis} satisfy the conditions described in Lemma \ref{lem:InfiBasis} with probability at least $1-(K|\mathcal{S}||\mathcal{A}|-|\mathcal{S}|^2|\mathcal{A}|)\cdot\eps$. Moreover, denote by $\mathcal{I}^*$ and $\mathcal{J}^*$ the output of \Cref{alg:IdenbasisTrue} on the true LP $V$. Then, for any $N_0$ satisfying \eqref{eqn:N0condition}, with a probability at least $1-(K|\mathcal{S}||\mathcal{A}|-|\mathcal{S}|^2|\mathcal{A}|)\cdot\eps$, we have $\mathcal{I}_{N_0}=\mathcal{I}^*$ and $\mathcal{J}_{N_0}=\mathcal{J}^*$.
\end{theorem}
The formal proof of \Cref{thm:Infibasis2} is relegated to \Cref{pf:Thm1}.
Note that there can be multiple optimal bases to the LP \eqref{lp:standard3}, and our approach essentially identifies one particular optimal basis. To see this point, denote by $\mathcal{I}^*$ and $\mathcal{J}^*$ the output of \Cref{alg:IdenbasisTrue} on the true LP $V$. The second part of \Cref{thm:Infibasis2} says that with a high probability, we will have exactly $\mathcal{I}_{N_0}=\mathcal{I}^*$ and $\mathcal{J}_{N_0}=\mathcal{J}^*$. Therefore, though the LP \eqref{lp:standard3} can be degenerate and enjoys multiple optimal basis, our \Cref{alg:Idenbasis} identifies one particular collection of non-zero basic variables and the supporting constraints. Note that though the bound in \Cref{thm:Infibasis2} depends on the parameters $\delta_1$, $\delta_2$, and $\sigma_0$ that could be unknown, we can actually modify the step 2-3 of our \Cref{alg:Twophase} a little bit to apply a doubling trick to exponentially increase the sample size that serves as input to \Cref{alg:Idenbasis}. We can guarantee that the input $N_1$ to \Cref{alg:Twophase} is large enough such that the condition in \eqref{eqn:N0condition} will be satisfied. 
Then, \Cref{thm:Infibasis2} guarantees that with a high probability, we find $\mathcal{I}^*$ and $\mathcal{J}^*$ in \Cref{alg:Twophase}, i.e., it holds that $\hat{\mathcal{I}}=\mathcal{I}^*$ and $\hat{\mathcal{J}}=\mathcal{J}^*$ with a high probability.

An important problem parameter related to $\mathcal{I}^*$ and $\mathcal{J}^*$ can be described as follows. Define
\begin{equation}\label{def:Amatrix}
A^*=\left[\begin{aligned}
&C(\mathcal{J}_1^*, \mathcal{I}^*)\\
&B(\mathcal{J}_2^*, \mathcal{I}^*)
\end{aligned}\right]
\end{equation}
where $\mathcal{J}^*_1=\mathcal{J}^*\cap[K]$ and $\mathcal{J}^*_2=\mathcal{J}^*\cap\mathcal{S}$.
We define $\sigma^*$ as
\begin{equation}\label{def:sigma}
\sigma^*= \sigma_{\min}(A^*)
\end{equation}
where $\sigma_{\min}(A^*)$ refers to the smallest singular value of the matrix $A^*$.
From the non-singularity of the matrix $A^*$, we know that $\sigma^*>0$.

\begin{remark} 
It is clear to see that $\sigma^*$ in \eqref{def:sigma} relates to the conditional number of the matrix $A^*$. In fact, it holds that $\kappa(A^*)\leq \frac{|\mathcal{S}|+K}{\sigma^*}$, where $\kappa(A^*)$ denotes the conditional number of the matrix $A^*$. The value of $\sigma^*$ will show up in our final bounds as it characterizes how hard it is to learn the matrix $A^*$. Also, $\sigma^*$ gives a natural upper bound of the range of the optimal dual variable corresponding to the optimal basis $\mathcal{I}^*$ and $\mathcal{J}^*$. Note that in previous literature that establishes logarithmic regret for online LP/resource allocation (e.g. \cite{vera2021bayesian, li2021symmetry, jiang2022degeneracy}), the conditional number of the constraint matrix will also show up in the final bounds. Therefore, we regard the existence of $\sigma^*$ as the consequence of adopting the online LP/resource allocation framework. On the other hand, for the MDP problem without resource constraints, the matrix $A^*$ can simply be represented as the probability transition matrix and one can easily show that $\sigma^*\geq1-\gamma$. We will use this bound when deriving our guarantees for MDP problems without resource constraints.
\end{remark}

\subsection{Regret Bounds}

We now show that \Cref{alg:Twophase} achieves a logarithmic regret, as defined in \eqref{eqn:regret}.
The next lemma shows that to analyze the regret, it suffices to consider the remaining constraint capacities belonging to the binding constraints. The formal proof is relegated to \Cref{pf:Lemma3}.

\begin{lemma}\label{lem:RemainRe}
For the optimal basis $\mathcal{I}^*$ and $\mathcal{J}^*$ identified in \Cref{alg:Idenbasis} with a high probability, we denote by $\bm{q}^*$ the corresponding primal optimal basic solution. We also denote by $(\bm{y}^*, \bm{c}^*)$ the optimal dual solution corresponding to $\bm{q}^*$. Then, it holds that
\begin{equation}\label{eqn:Redompose}
N_2\cdot V_r(\pi^*, \mu_1)-\sum_{n=1}^{N_2} \hat{\bm{r}}^\top\mathbb{E}[\bm{q}^n]\leq \sum_{j\in\mathcal{J}_1^*} y^*_j\cdot \mathbb{E}[\alpha^{N_2}_j]+\sum_{s\in\mathcal{J}_2^*}z^*_s\cdot \mathbb{E}[\mu^{N_2}_s],
\end{equation}
where $\{\bm{q}^1,\dots,\bm{q}^{N_2}\}$ is the output of \Cref{alg:Twophase}.
\end{lemma}
We now condition on the event that $\hat{\mathcal{I}}=\mathcal{I}^*$ and $\hat{\mathcal{J}}=\mathcal{J}^*$. It suffices to analyze how the ``remaining resources'' $(\bm{\alpha}_{\mathcal{J}_1^*}^n, \bmu_{\mathcal{J}^*_2}^n)$ behave. We now define
\begin{equation}\label{eqn:Average}
\tilde{\alpha}_k(n)=\frac{\alpha^n_k}{N_2-n},~\forall k\in\mathcal{J}_1^*\text{~and~} \tilde{\mu}_s(n)=\frac{\mu^n_s}{N_2-n}, ~\forall s\in\mathcal{J}^*_2, \forall n\in[N_2].
\end{equation}
The key is to show that the stochastic process $\tilde{\alpha}_k(n)$ and $\tilde{\mu}_s$ possess some concentration properties such that they will stay within a small neighborhood of their initial value $\alpha^1_k$ and $\mu^1_s$ for a sufficiently long time. We denote by $\tau$ the time that one of $\tilde{\alpha}_k(n)$ for each $k\in\mathcal{J}_1^*$ and $\tilde{\mu}_s(n)$ for each $s\in\mathcal{J}^*_2$ escape this neighborhood. Then, both $\text{Regret}_r(\pi, N_2)$ and $\text{Regret}_k(\pi, N_2)$ for each $k\in[K]$ can be upper bounded by $\mathbb{E}[N_2-\tau]$. From the update rule \eqref{eqn:UpdateAlpha2} and \eqref{eqn:UpdateMu2}, we know that
\begin{equation}\label{eqn:Aveupalpha}
\tilde{\alpha}_k(n+1)=\tilde{\alpha}_k(n)-\frac{\sum_{(s,a)\in\mathcal{I}^*}\bm{c}^n(s,a)\cdot q^n(s,a)-\tilde{\alpha}_k(n)}{N_2-n-1},~\forall k\in\mathcal{J}_1^*
\end{equation}
and
\begin{equation}\label{eqn:Aveupmu}
\tilde{\mu}_s(n+1)=\tilde{\mu}_s(n)-\frac{\sum_{(s,a)\in\mathcal{I}^*}q^n(s,a)\cdot(\delta_{s',s}-\gamma\bI_{\{s'=s^n(s,a)\}})-\tilde{\mu}_s(n)}{N_2-n-1},~\forall s\in\mathcal{J}_2^*.
\end{equation}
Ideally, both $\tilde{\alpha}_k(n+1)$ and $\tilde{\mu}_s(n+1)$ will have the same expectation as $\tilde{\alpha}_k(n)$ and $\tilde{\mu}_s(n)$ such that they become a martingale. However, this is not true since we have estimation error over $C(\mathcal{J}_1^*, \mathcal{I}^*)$ and $B(\mathcal{J}^*_2,\mathcal{I}^*)$, and we only use their estimates to compute $\bm{q}^n$. Nevertheless, we can show that $\tilde{\alpha}_k(n)$ for each $k\in\mathcal{J}_1^*$ and $\tilde{\mu}_s$ for each $s\in\mathcal{J}_2^*$ behave as a sub-martingale. Then, from the concentration property of the sub-martingale, we upper bound $\mathbb{E}[\alpha^{N_2}_k]$ for each $k\in\mathcal{J}_1^*$ and $\mathbb{E}[\mu^{N_2}_s]$ for each $s\in\mathcal{J}_2^*$. The term $|\mathbb{E}[\alpha^{N_2}_k]|$ for each $k\in[K]\backslash\mathcal{J}_1^*$, and the term $\mathbb{E}[\mu^{N_2}_s]$ for each $s\in\mathcal{S}\backslash\mathcal{J}_2^*$, can be upper bounded as well. The results are presented in the following theorem. The formal proof is relegated to \Cref{pf:Thm2} and we provide a brief sketch of the proof for illustration.

\begin{theorem}\label{thm:BoundRe}
Denote by $N_2$ the number of resolving rounds in \Cref{alg:Twophase}. As long as $N_2$ satisfies that
\[
N_2\geq \frac{128m^8}{(\sigma^*)^2\cdot(1-\gamma)^2}\cdot\log(1/\eps)
\]
where $m=|\mathcal{I}^*|=|\mathcal{J}^*|$, $\eps=$, $\sigma^*$ defined in \eqref{def:sigma}, it holds that
\[
N_2\cdot V_r(\pi^*, \mu_1)-\sum_{n=1}^{N_2} \hat{\bm{r}}^\top\mathbb{E}[\bm{q}^n]\leq \frac{16m^4}{\sigma^*}+\frac{64m^7}{(\sigma^*)^3}\cdot\log(1/\eps)
\]
Also, for any $k\in[K]$, we have
\begin{equation}\label{eqn:020412}
N_2\cdot\alpha_k-\sum_{n=1}^{N_2}\hat{\bm{c}}_k^\top\mathbb{E}\left[\bm{q}^n \right] \leq \frac{16m^4}{\sigma^*}+\frac{64m^7}{(\sigma^*)^3}\cdot\log(1/\eps)
\end{equation}
and
\begin{equation}
    \left\|N_2\cdot\bmu-B\cdot\sum_{n=1}^{N_2}\mathbb{E}[\bm{q}^n] \right\|_{\infty}\leq \frac{16m^4}{\sigma^*}+\frac{64m^7}{(\sigma^*)^3}\cdot\log(1/\eps).
\end{equation}
\end{theorem}
\begin{myproof}[Sketch of proof]
We first note that following \Cref{thm:Infibasis2}, as long as $N_1$ satisfies the condition \eqref{eqn:N0condition}, we correctly identify the basis $\mathcal{I}^*$ and $\mathcal{J}^*$ with a high probability. Then, the resolving step \eqref{eqn:OptQ2} simply becomes
\begin{equation}\label{eqn:022201}
\left[\begin{aligned}
&\bar{C}^n(\mathcal{J}_1^*, \mathcal{I}^*)\\
&\bar{B}^n(\mathcal{J}_2^*, \mathcal{I}^*)
\end{aligned}\right]
\cdot\tilde{\bm{q}}^n_{\mathcal{I}^*}=\left[\begin{aligned}
&\frac{\bm{\alpha}^n_{\mathcal{J}_1^*}}{N_2-n+1}\\
&\frac{\bmu_{\mathcal{J}^*_2}^n}{N_2-n+1}
\end{aligned}\right].
\end{equation}
Our remaining proof can be classified into the following two steps.
\begin{itemize}
    \item[1.] For a fixed $\nu>0$, we define a set
\begin{equation}\label{eqn:defX2}
\mathcal{X}=\{ \bm{\alpha}'\in\mathbb{R}^{|\mathcal{J}_1^*|}: \alpha'_k\in[\alpha_k-\nu, \alpha_k+\nu], \forall k\in\mathcal{J}_1^* \},
\end{equation}
and
\begin{equation}\label{eqn:defY2}
\mathcal{Y}=\{ \bm{\mu}'\in\mathbb{R}^{|\mathcal{J}^*_2|}: \mu'_s\in[\mu_s-\nu, \mu_s+\nu], \forall s\in\mathcal{J}^*_2 \}.
\end{equation}
We can show that $\tilde{\bm{\alpha}}(n)$ and $\tilde{\bmu}(n)$ behave well as long as they stay in the region $\mathcal{X}$ and $\mathcal{Y}$. To be specific, it automatically satisfies that $\|\tilde{\bm{q}}^n_{\mathcal{I}^*}\|_1\leq2$ and thus $\bm{q}^n_{\mathcal{I}^*}$ can be expressed as the solution to the linear equations \eqref{eqn:022201}.

We can also show that $\tilde{\bm{\alpha}}(n)$ and $\tilde{\bmu}(n)$ stay in the region $\mathcal{X}$ and $\mathcal{Y}$ for a sufficiently long time.
To this end, we define a stopping time
\begin{equation}\label{eqn:Stoptime}
\tau=\min_{n\in[N]}\{ \tilde{\bm{\alpha}}(n) \notin\mathcal{X}\text{~or~}\tilde{\bmu}(n)\notin\mathcal{Y} \}.
\end{equation}
and we derive an upper bound of $\mathbb{E}[N_2-\tau]$, which scales as $O(\log(N_2))$.

\item[2.] By using our upper bound of $\mathbb{E}[N_2-\tau]$ and combining with \Cref{lem:RemainRe}, we show that the gap over the reward $N_2\cdot V_r(\pi^*, \mu_1)-\sum_{n=1}^{N_2} \hat{\bm{r}}^\top\mathbb{E}[\bm{q}^n]$ and the gap over the constraint $N_2\cdot\alpha_k-\sum_{n=1}^{N_2}\hat{\bm{c}}_k^\top\bm{q}^n$ for $k\in\mathcal{J}^{*}_1$ and $N_2\cdot\bmu_{\mathcal{J}^*_2}-\sum_{n=1}^{N_2}B(\mathcal{J}^*_2, :)\bm{q}^n$ can all be upper bounded at the order of $O(\log(N_2))$. By further exploiting the structure of the matrix $A^*$, we can show that the gap over the constraint $N_2\cdot\alpha_k-\sum_{n=1}^{N_2}\hat{\bm{c}}_k^\top\bm{q}^n$ for $k\in(\mathcal{J}_1^*)^c$, and $N_2\cdot\bmu_{(\mathcal{J}^*_2)^c}-\sum_{n=1}^{N_2}B((\mathcal{J}^*_2)^c, :)\bm{q}^n$ can be bounded at the order of $O(\log(N))$ as well.

\end{itemize}
\end{myproof}

\subsection{Sample Complexity Bounds}

Denote by $\bm{q}^1, \dots, \bm{q}^{N_2}$ the output of \Cref{alg:Twophase}. Then, We define $\bar{\bm{q}}^{N_2}$ such that
$\bar{\bm{q}}^{N_2}_{\hat{\mathcal{I}}^{c}}=0$
and
\[
\bar{\bm{q}}^{N_2}_{\hat{\mathcal{I}}}=\frac{1}{N_2}\cdot\sum_{n=1}^{N_2}\bm{q}^n_{\hat{\mathcal{I}}}.
\]
From \Cref{thm:Infibasis2}, as long as $N_1$satisfies the condition \eqref{eqn:N0condition}, we know that $\hat{\mathcal{I}}=\mathcal{I}^*$ with a high probability.
We can construct a policy $\bar{\pi}$ as follows. For each $(s,a)\in\mathcal{S}\times\mathcal{A}$, it holds
\begin{equation}\label{eqn:012501}
P(\bar{\pi}(s)=a) = \left\{
\begin{aligned}
&\frac{\bar{q}^{N_2}(s,a)}{\sum_{a'\in\mathcal{A}}\bar{q}^N(s,a')}, && \text{if~}\sum_{a'\in\mathcal{A}}\bar{q}^{N_2}(s,a')>0\\
&1/|\mathcal{A}|, &&\text{if~}\sum_{a'\in\mathcal{A}}\bar{q}^{N_2}(s,a')=0.
\end{aligned}\right.
\end{equation}
We aim to bound the gap between $V_r(\pi^*, \mu_1)$ and $V_r(\bar{\pi}, \mu_1)$, as well as $V_k(\bar{\pi}, \mu_1)$ and $\alpha_k$ for each $k\in[K]$. Note that in the definition \eqref{eqn:012501}, the value of $\bar{\pi}$ is essentially random, where the randomness comes from the filtration $\mathcal{H}_{N_2}$ that decides the value of $\bm{q}^1, \dots, \bm{q}^{N_2}$. Therefore, in the definition of $V_r(\bar{\pi}, \mu_1)$ and $V_k(\bar{\pi}^{N_2}, \mu_1)$ for each $k\in[K]$, we take expectation over the filtration $\mathcal{F}_{N_2}$. With the above definitions,
the regret bound established in \Cref{thm:BoundRe} can be transferred into the sample complexity bound of the policy $\bar{\pi}$ given below. The formal proof is relegated to \Cref{pf:Thm3}.
\begin{theorem}\label{thm:sample}
With a sample complexity bound of
\[
N=O\left( \frac{|\mathcal{S}||\mathcal{A}|\cdot\log(1/\veps)}{\min\{\delta_1^2, \delta_2^2, \sigma_0^2\}}+\frac{m^7\cdot\log(1/\xi)}{\xi(\sigma^*)^3(1-\gamma)}\cdot\frac{\log^2(1/\veps)}{\veps} \right),
\]
where $\xi=\min_{(s,a)\in\mathcal{I}^*} \left\{ q^*(s,a) \right\}>0$ and $\bm{q}^*$ denotes the optimal solution to LP \eqref{lp:standard3} corresponding to the optimal basis $\mathcal{I}^*$ and $\mathcal{J}^*$,
we obtain a policy $\bar{\pi}$ from \Cref{alg:Twophase} (defined in \eqref{eqn:012501}) such that
\[
V_r(\pi^*, \mu_1) - V_r(\bar{\pi}, \mu_1)\leq\veps\text{~~and~~}V_k(\bar{\pi}, \mu_1) - \alpha_k\leq\veps,~\forall k\in[K].
\]
\end{theorem}
Note that one caveat in transferring the bounds in \Cref{thm:BoundRe} into the sample complexity bound of the policy $\bar{\pi}$ is that the vector $\bar{\bm{q}}$ is \textit{not} an occupancy measure since there is a gap between $B\bar{\bm{q}}^{N_2}$ and $\bmu$, though bounded at the order of $O(\log N_2/N_2)$. However, one can utilize the structure \eqref{eqn:OptQ} to show a bound over the gap between $\bar{\bm{q}}^{N_2}$ and $\bm{q}^*$, pointwise. In this way,
the gap between their state value functions can be bounded, which induces an additional $O(\log(N_2))$ term. Therefore, we obtain our final sample complexity bounds, scaling as $\frac{O(\log^2(1/\veps))}{\veps}$.

Our \Cref{alg:Twophase} can be directly applied to solving MDP problems without resource constraints and our bounds in \Cref{thm:BoundRe} and \Cref{thm:sample} still hold. Note that in the MDP problems, the parameter $\sigma$ can be lower bounded by $1-\gamma$, which follows from the fact that the matrix $A^*$ can simply be represented by the probability transition matrix. Then, we have the following sample complexity bound for our \Cref{alg:Twophase}.
\begin{proposition}\label{prop:sampleMDP}
For the MDP problems without resource constraints, i.e., $K=0$, assuming $\xi=O(1/(|\mathcal{S}||\mathcal{A}|))$,
with a sample complexity bound of
\begin{equation}\label{eqn:SampleMDP}
O\left( \frac{|\mathcal{S}||\mathcal{A}|}{(1-\gamma)^4\cdot\Delta}\cdot\frac{\log^2(1/\veps)}{\veps}\right),
\end{equation}
with $\Delta=\min\{\delta_1^2, \delta_2^2, \sigma_0^2\}$,
we obtain a policy $\bar{\pi}$ from \Cref{alg:Twophase} (defined in \eqref{eqn:012501}) such that
\[
V_r(\pi^*, \mu_1) - V_r(\bar{\pi}, \mu_1)\leq\veps\text{~~and~~}V_k(\bar{\pi}, \mu_1) - \alpha_k\leq\veps,~\forall k\in[K].
\]
\end{proposition}
Note that one significant term in our sample complexity bound in \Cref{prop:sampleMDP} is $\Delta$. Recall that $\Delta$ represents the change of the objective value of LP \eqref{lp:standard3} if we restrict one non-zero basic variable in an optimal solution to be $0$. In this sense, $\Delta$ can be understood as the gap between the value of the optimal policy and the value of a sub-optimal policy that is restricted from taking the optimal action given a state, corresponding to the instance hardness via corner point as described in \Cref{sec:characterization}. Such a problem-dependent gap shows up broadly in the literature for online decision-making with (logarithmic) problem-dependent guarantees, for example, in the multi-arm-bandit problem \citep{lai1985asymptotically} and the reinforcement learning problem \citep{auer2008near}.

In terms of the dependency of our sample complexity bound on other problem parameters such as $|\mathcal{S}|$, $|\mathcal{A}|$, and $1-\gamma$, we compare to the series of work \citep{sidford2018near, wainwright2019variance, wang2020randomized, agarwal2020model, he2021nearly}, that subsequently achieves a sample complexity bound of $O\left( \frac{|\mathcal{S}|\cdot|\mathcal{A}|}{(1-\gamma)^3\cdot\veps^2} \right)$, where the dependency over $|\mathcal{S}|$, $|\mathcal{A}|$, and $1-\gamma$ is optimal \citep{gheshlaghi2013minimax, lattimore2012pac}. Our sample complexity bound in \eqref{eqn:SampleMDP} has a worse dependency in terms of $1-\gamma$. This is because we construct an empirical LP to estimate the value of LP \eqref{lp:standard3} and resolve the linear equation as in \eqref{eqn:OptQ2}, where the size of the LP and the singular values of the matrix $A^*$ (which is bounded by $1-\gamma$) will play a part. However, our bound \eqref{eqn:SampleMDP} enjoys a better dependency in terms of $\veps$. Also, our bound \eqref{eqn:SampleMDP} holds under the assumption that $\xi=O(1/(|\mathcal{S}||\mathcal{A}|))$. In the extreme case where $\xi$ can be extremely small, our bound will become worse, scaling with $1/\xi$.

\section{Extensions}\label{sec:extensions}
We describe the generalizations of our approach and results in this section. We first describe the generalization to the finite horizon episodic setting, where we maximize the cumulative reward over a finite horizon without discounting. We then describe the generalization to the off-policy and on-policy learning settings, which are beyond the generative model in that we can only get the sample of the state we are currently visiting, instead of the sample of an arbitrary state. We show that our approach can be further developed and obtain a similar $\tilde{O}(1/\eps)$ sample complexity.
\subsection{Finite Horizon Episodic Setting}\label{sec:Finitehorizon}

We now generalize our approach and results to the finite horizon episodic setting, where there are $H$ periods in the horizon. Now there are $H$ transition kernels $\{P_h\}_{h=1}^H$. We let $P_h(s'|s,a)$ denote the probability of transiting to state $s'$ when the state action pair is $(s,a)$ at period $h$. As before, the initial distribution over the state is denoted by $\mu_1$. For each period $h$, there is a stochastic reward function $r_h:\mathcal{S}\times\mathcal{A}\rightarrow[0,1]$ and a stochastic cost function $c_{h,k}:\mathcal{S}\times\mathcal{A}\rightarrow[0,1]$ for each $k\in[K]$. For each state-action pair $(s,a)$ and each period $h$, we also denote by $\hat{r}_h(s,a)=\mathbb{E}[r_h(s,a)]$ and $\hat{c}_{h,k}(s,a)=\mathbb{E}[c_{h,k}(s,a)]$. Then, for any Markovian policy $\pi$, the average reward is defined as
\begin{equation}\label{eqn:Avereward}
V_r(\pi, \mu_1) = \mathbb{E}\left[ \sum_{h=1}^H r_h(s_h, a_h)\mid \mu_1 \right],
\end{equation}
and the constraint is defined as
\begin{equation}\label{eqn:Aveconstraint}
V_k(\pi, \mu_1) = \mathbb{E}\left[ \sum_{h=1}^H c_{h,k}(s_h, a_h)\mid \mu_1 \right] \leq \beta_k,~\forall k\in[K].
\end{equation}
Note that the policy $\pi$ is allowed to be \textit{non-stationary}, which means that the behavior of the policy $\pi$ on two different periods can be different, even given the same current state.
However, the reward functions $\{r_h\}_{h=1}^H$, cost functions $\{c_{h,k}\}_{\forall h\in[H], \forall k\in[K]}$, and the transition kernels $\{P_h\}_{h=1}^H$ are all unknown to the decision maker and needs to be learned from the samples. We still assume the existence of a generative model.
\begin{assumption}\label{assump:2}
For each period $h$, each state and action pair $(s,a)$, we can query the model $\mathcal{M}$ to obtain an observation of $r_h(s,a)$, $c_{h,k}(s,a)$ for each $k\in[K]$, and the new state $s'\in\mathcal{S}$, where the transition from $s$ to $s'$ follows the probability kernel $P_h(s'|s,a)$ independently.
\end{assumption}
We still consider the sample complexity bound of our policy. Denote by $\pi^*$ the optimal Markovian policy. 
For any $\veps$, we aim to find an $\veps$-accurate policy $\pi$ such that
\begin{equation}\label{eqn:complexity2}
V_r(\pi^*, \mu_1) - V_r(\pi, \mu_1)\leq\veps\text{~~and~~}V_k(\pi, \mu_1) - \beta_k\leq\veps,~\forall k\in[K],
\end{equation}
with as few samples as possible.
\subsubsection{LP Reformulation and Estimates}\label{sec:EpiReformulation}
For the finite horizon episodic problem, the occupancy measure is defined as $q_{h, \pi}(s,a)$ for each state-action pair $(s,a)$, each period $h\in[H]$, and the policy $\pi$. Now $q_{h,\pi}(s,a)$ represents the probability of taking action $a$ at period $h$ with state $s$, under policy $\pi$. Following a series of work (e.g. \citet{zimin2013online, bhattacharya2017linear, efroni2020exploration}), we obtain the following linear programming to represent the optimal policy and the optimal occupancy measure.
\begin{subequations} \label{lp:Finite}
\begin{align}
\VF=~\max\ &\sum_{h\in[H]}\sum_{s\in\mathcal{S}}\sum_{a\in\mathcal{A}} \hat{r}_h(s,a)\cdot q_h(s,a)& \label{eqn:InfiObj}
\\ \mathrm{s.t.\ }&\sum_{h\in[H]}\sum_{s\in\mathcal{S}}\sum_{a\in\mathcal{A}} \hat{c}_{h,k}(s,a)\cdot q_h(s,a)\leq \beta_k &\forall k\in[K] \label{eqn:Ficonstraint1}
\\ &\sum_{a\in\mathcal{A}} q_h(s,a)=\sum_{s',a'}P_{h-1}(s|s',a')\cdot q_{h-1}(s',a') &\forall s\in\mathcal{S}, \forall [H]\setminus \{1\} \label{eqn:Ficonstraint2}
\\ & \sum_{a\in\mathcal{A}} q_1(s,a) = \mu_1(s) & \forall s\in\mathcal{S}
\\ & q_h(s,a) \geq 0 &\forall h\in[H], \forall s\in \mathcal{S}, a\in \mathcal{A}, \label{eqn:Ficonstraint3}
\end{align}
\end{subequations}

Again, we can rewrite LP \eqref{lp:Finite} into the standard formulation LP \eqref{lp:standard3}. Now, the matrix $B\in\mathbb{R}^{|\mathcal{S}|H\times|\mathcal{S}||\mathcal{A}|H}$. For $h=1$ and $s\in\mathcal{S}$, the element of $B$ at the $s$-th row and $(s,a,1)$-th column is $1$, for $a\in\mathcal{A}$. For $h\geq2$ and $s\in\mathcal{S}$, the element of $B$ at $(h-1)\cdot|\mathcal{S}|+s$-th row and $(s,a',h)$-th column is $1$, while the element of $B$ at $sh$-th row and $(s',a',h-1)$-th column is $-P_{h-1}(s|s',a')$, for $s'\in\mathcal{S}$ and $a'\in\mathcal{A}$. The element of $B$ at other rows and columns is set to be $0$.  The matrix $C\in\mathbb{R}^{K\times|\mathcal{S}||\mathcal{S}|H}$ and the element at $k$-th row and $(s,a,h)$-th column is $\hat{c}_{h,k}(s,a)$, for $k\in[K]$ and for $s\in\mathcal{S}$, $a\in\mathcal{A}$, $h\in[H]$. The vector $\bmu\in\mathbb{R}^{|\mathcal{S}|H}$ and the first $s$-th element of $\bmu$ is $\mu_1(s)$, for $s\in\mathcal{S}$, and the other elements of $\bmu$ are $0$.
Then, in the same way as described in \Cref{sec:Estimates}, we can construct $\bar{V}_{N_0}$, $\bar{V}_{\mathcal{I}, N_0}$, and $\text{Dual}_{\mathcal{J}, \mathcal{I}, N_0}$ with the newly specified matrix $B$ and $C$.

We now specify the choice of $\lambda_0$ and bound the estimation gap for the finite horizon episodic setting. To this end, we denote by $\mathcal{I}^*$ and $\mathcal{J}^*$ an optimal basis of LP \eqref{lp:Finite}, and we denote by
\begin{equation}\label{def:1APrimematrix}
A^*=\left[\begin{aligned}
&C(\mathcal{J}^*_1, \mathcal{I}^*)\\
&B(\mathcal{J}^*_2, \mathcal{I}^*)
\end{aligned}\right].
\end{equation}
We then define $\sigma^*$ as
\begin{equation}\label{def:1sigmaPrime}
\sigma^*= \sigma_{\min}(A^*)
\end{equation}
where $\sigma_{\min}(a^*)$ refers to the smallest singular value of the matrix $A^*$, where it holds that $\sigma^*>0$. 
We have the following lemma as the bound of the estimation error.
\begin{lemma}\label{lem:FiLUbound}
As long as $\lambda_{N_0}=\Rad(N_0,\eps)\cdot H$,
the following inequality
\begin{equation}\label{eqn:Vgap1}
\begin{aligned}
V \leq\bar{V}_{N_0}+H\cdot\Rad(N_0, \eps)\leq &V+H\cdot\Rad(N_0, \eps)
+\Rad(N_0,\eps)\cdot \frac{2(K+|\mathcal{S}|\cdot H)}{\sigma^*}\\
&+\Rad^2(N_0,\eps)\cdot\frac{K\cdot|\mathcal{S}|+|\mathcal{S}|^2\cdot H}{\sigma^*}.
\end{aligned}
\end{equation}
holds with probability at least $1-H\cdot(K|\mathcal{S}||\mathcal{A}|-|\mathcal{S}|^2|\mathcal{A}|)\cdot\eps$.
\end{lemma}

\subsubsection{Algorithm and Theoretical Bound}

Note that with the LP reformulation $\VF$ in \eqref{lp:Finite}, our \Cref{alg:Idenbasis} and \Cref{alg:Twophase} apply to the episodic setting. The only change we need to make is to replace the matrix $B$ and $C$ with the one described in \Cref{sec:EpiReformulation}. Denote by $\bm{q}^1, \dots, \bm{q}^{N_2}$ the output of \Cref{alg:Twophase}. Then, We define $\bar{\bm{q}}^{N_2}$ such that
$\bar{\bm{q}}^{N_2}_{\hat{\mathcal{I}}^{c}}=0$
and
\[
\bar{\bm{q}}^{N_2}_{\hat{\mathcal{I}}}=\frac{1}{N_2}\cdot\sum_{n=1}^{N_2}\bm{q}^n_{\hat{\mathcal{I}}}.
\]
Here, $\hat{\mathcal{I}}$ refers to the output of \Cref{alg:Idenbasis} during the execution of \Cref{alg:Twophase}. Then, we can construct a policy $\bar{\pi}$ as follows. For each $h\in[H]$ and $(s,a)\in\mathcal{S}\times\mathcal{A}$, it holds
\begin{equation}\label{eqn:051901}
P(\bar{\pi}_h(s)=a) = \left\{
\begin{aligned}
&\frac{\bar{q}^{N_2}_h(s,a)}{\sum_{a'\in\mathcal{A}}\bar{q}^{N_2}_h(s,a')}, &&\text{if~}\sum_{a'\in\mathcal{A}}\bar{q}^{N_2}_h(s,a')>0\\
&1/|\mathcal{A}|, &&\text{if~}\sum_{a'\in\mathcal{A}}\bar{q}^{N_2}_h(s,a')=0,
\end{aligned}
\right.
\end{equation}
where $P(\bar{\pi}_h(s)=a)$ denotes the probability for the policy $\bar{\pi}$ to take the action $a\in\mathcal{A}$ given the state $s\in\mathcal{S}$ at period $h\in[H]$. Similarly, if $\sum_{a'\in\mathcal{A}}\bar{q}^{N_2}_h(s,a')=0$, we can take an arbitrary action. Then, we have the following bound regarding the sample complexity of the policy $\bar{\pi}$, as a corollary of \Cref{thm:sample}.

\begin{corollary}\label{thm:1sample}
With a sample complexity bound of
\[
N=O\left( \frac{H|\mathcal{S}||\mathcal{A}|\cdot\log(1/\veps)}{\min\{\delta_1^2, \delta_2^2, \sigma_0^2\}}+\frac{m^7\cdot\log(1/\xi)}{\xi(\sigma^*)^3(1-\gamma)}\cdot\frac{\log^2(1/\veps)}{\veps} \right),
\]
where $\xi=\min_{h\in[H],(s,a)\in\mathcal{I}^*} \left\{ q_h^*(s,a) \right\}$ and $\bm{q}^*$ denotes the optimal solution to LP \eqref{lp:standard3} corresponding to the optimal basis $\mathcal{I}^*$ and $\mathcal{J}^*$,
we obtain a policy $\bar{\pi}$ from \Cref{alg:Twophase} (defined in \eqref{eqn:051901}) such that
\[
V_r(\pi^*, \mu_1) - V_r(\bar{\pi}, \mu_1)\leq\veps\text{~~and~~}V_k(\bar{\pi}, \mu_1) - \alpha_k\leq\veps,~\forall k\in[K].
\]
\end{corollary}

\subsection{Off-policy Learning Setting.}\label{sec:offpolicy}
Note that in the previous sections, we assume the existence of a generative model where we can obtain the sample of reward, cost, and state transition for an arbitrary state and action pair. In this section, we show that our approaches and results can be directly generalized to more involved sampling environments, i.e., the offline learning environments, and we discuss the extension to the online learning setting in the next \Cref{sec:onpolicy}. We illustrate under the infinite horizon discounted setting described in \Cref{sec:Infinitedis}. However, our results can also be directly applied to the finite horizon episodic setting described in \Cref{sec:Finitehorizon}.

Under the off-policy learning setting, we interact with the true environment to obtain samples by following a behavior policy. Note that the behavior policy is purely for collecting samples and can be different from the target policy.  
To be specific, suppose that at round $n$ the environment is at state $s_n$. Then, in order to obtain the samples, we need to choose an action, denoted by $a_n$, and we only obtain the sample of $r(s_n, a_n)$, $c_k(s_n, a_n)$ for each $k\in[K]$, and the state transition $(s_n, a_n)\rightarrow s_{n+1}$. This will push the environment to state $s_{n+1}$ and we can only obtain information about state $s_{n+1}$ at the next round. Compared to the generative model described in \Cref{assump:1} where we can obtain samples for an arbitrary state and action pair, the off-policy learning setting is more restricted in that we can only obtain information about the state we are currently at. The off-policy learning setting is formally described in the following assumption.
\begin{assumption}\label{assump:OnlineLearning}
Denote by $\mathcal{M}$ the true environment, with the state space
$\mathcal{S}$, action space $\mathcal{A}$, and the transition kernel $P$. Given the current state $s$ and the current action $a$, we obtain a sample of $r(s,a)$, $c_k(s,a)$ for each $k\in[K]$, and the state transition $(s,a)\rightarrow s'$ following the transition kernel. Then, the state of $\mathcal{M}$ will transit to $s'$.
\end{assumption}

The generalization to the off-policy learning setting is direct as we use a behavior policy that is different from our policy to generate samples and construct estimates of the underlying CMDP. We then use the obtained samples to form the sample set $\mathcal{F}$ to carry out \Cref{alg:Idenbasis} and \Cref{alg:Twophase}. Similar to \cite{li2021sample}, we use a stationary policy $\pi_s$ as the behavior policy. Given the stationarity of $\pi_s$, the sample trajectory can be viewed as a sample path of a time-homogeneous Markov chain over the set of state-action pairs $\mathcal{S}\times\mathcal{A}$. Following \cite{paulin2015concentration}, we make the following uniform ergodicity assumption.
\begin{assumption}\label{assump:Ergodicity}
The Markov chain induced by the stationary behavior policy $\pi_s$ is uniformly ergodic.
\end{assumption}
Note that a key property to consider is how often we can obtain a sample of a state-action pair $(s,a)$, which measures the learning efficiency under the stationary behavior policy. Denote by $\mu_s$ the stationary distribution over all state-action pairs under the behavioral policy. We then define
\begin{equation}\label{eqn:Mumin}
\mu_{\min}=\min_{(s,a)\in\mathcal{S}\times\mathcal{A}} \mu_s(s,a).
\end{equation}
Suppose that we obtain $N$ samples under the stationary distribution $\mu_s$. Then, following the standard Hoeffding's inequality, we know that we obtain the number of samples at least $\frac{\mu_{\min}\cdot N_1}{2}$ for each state-action pair $(s,a)$, with a high probability $1-O(1/N_1)$. Therefore, we can just pick the number of samples $\frac{\mu_{\min}\cdot N_1}{2}$ for each state-action pair $(s,a)$ to construct the data set to serve as input to \Cref{alg:Idenbasis} in the execution of \Cref{alg:Twophase}. For the resolving procedure in \Cref{alg:Twophase} and for each round of iteration, we collect samples using the behavior policy until we collect one sample for each $(s,a)$. 
We then have the following sample complexity bound under the off-policy learning setting, which generalizes the analysis in \Cref{thm:sample}.
\begin{corollary}\label{coro:OfflineSample}
For any $\eps>0$, with a sample complexity bound of
\[
N=O\left( \frac{\log(1/\veps)}{\mu_{\min}\cdot\min\{\delta_1^2, \delta_2^2, \sigma_0^2\}}+\frac{m^5\cdot\log(1/\xi)}{\mu_{\min}\cdot\xi(\sigma^*)^3(1-\gamma)}\cdot\frac{\log^2(1/\veps)}{\veps} \right),
\]
where $\mu_{\min}$ is given in \eqref{eqn:Mumin} and $\xi=\min_{(s,a)\in\mathcal{I}^*} \left\{ q^*(s,a) \right\}$ and $\bm{q}^*$ denotes the optimal solution to LP \eqref{lp:standard3} corresponding to the optimal basis $\mathcal{I}^*$ and $\mathcal{J}^*$,
we obtain a policy $\bar{\pi}$ from \Cref{alg:Twophase} such that
\[
V_r(\pi^*, \mu_1) - V_r(\bar{\pi}, \mu_1)\leq\veps\text{~~and~~}V_k(\bar{\pi}, \mu_1) - \alpha_k\leq\veps,~\forall k\in[K].
\]
\end{corollary}


\subsection{On-policy Learning Setting.} \label{sec:onpolicy}
Different from the off-policy learning setting, under the on-policy learning setting, we interact with the true environment to obtain samples by following the target policy. 
To be specific, suppose that at round $n$ the environment is at state $s_n$. Then, denoting by $\bm{q}_t$ the occupancy measure computed by our algorithm, in order to obtain the samples, we need to choose an action following the occupancy measure $\bm{q}_t$, denoted by $a_n$, and we only obtain the sample of $r(s_n, a_n)$, $c_k(s_n, a_n)$ for each $k\in[K]$, and the state transition $(s_n, a_n)\rightarrow s_{n+1}$. This will push the environment to state $s_{n+1}$ and we can only obtain information about state $s_{n+1}$ at the next round. Compared to the generative model described in \Cref{assump:1} where we can obtain samples for an arbitrary state and action pair, the on-policy learning setting is more restricted in that we can only obtain information about the state we are currently at. The on-policy learning setting is even more restrictive than the off-policy learning setting described in \Cref{sec:offpolicy} in that we cannot adopt a behavior policy to collect samples, instead, our algorithm needs to satisfy certain properties such that on one hand, we can approximate the optimal policy with instance-dependent guarantees, and on the other hand, we can collect enough samples over the state-action pairs we are concerned about. 

Our algorithms can be described as follows. There are two phases of our algorithms. In the first phase, we still use \Cref{alg:Idenbasis} to identify the optimal basis $\mathcal{I}^*$ and $\mathcal{J}^*$. Note that the policy will be a stationary policy, and following the same procedure of \Cref{thm:Infibasis2}, we can show that \Cref{alg:Idenbasis} enables us to identify one optimal basis with a high probability within a finite time steps scaling as $O(\log(1/\eps)/\Delta)$. Then, in the second phase, we apply the resolving procedure in \Cref{alg:Twophase} during the remaining time periods. Denoting by $\bm{q}^*$ the optimal occupancy measure obtained from solving the LP \eqref{lp:standard3} corresponding to the optimal basis $\mathcal{I}^*$ and $\mathcal{J}^*$, we can show that during the resolving procedure after a warm start in the first phase, the $\bm{q}^n$ we obtain is always positively supported on $\mathcal{I}^*$. Therefore, if we follow the policy induced by $\bm{q}^n$ at each iteration $n$ for a constant number of times, we can actually show that we are able to collect at least one sample over the costs and the state transition associated with each $(s,a)\in\mathcal{I}^*$, with a high probability, which is already enough to estimate the matrix $A^*=\begin{bmatrix}
C(\mathcal{J}_1^*, \mathcal{I}^*)\\
B(\mathcal{J}_2^*, \mathcal{I}^*)
\end{bmatrix}$ more and more accurately during the resolving procedure. This result shows that our resolving procedure would automatically guarantee exploitation and exploration at the same time, and there is no need to do additional exploration in our algorithmic design. The formal algorithm for the online setting is presented in \Cref{alg:TwophaseOnline}.

\begin{algorithm}[ht!]
\caption{The Adaptive-resolving Algorithm for On-policy Learning}
\label{alg:TwophaseOnline}
\begin{algorithmic}[1]
\State \textbf{Input:} the number of rounds $N_1$ for carrying out \Cref{alg:Idenbasis}, the number of rounds $N_2$ for resolving, the limit of the number of steps $N_3$ for sampling, an estimate $\xi'$ such that $\frac{\xi}{4}\leq\xi'\leq\xi/2$.
\State Follow a stationary policy by $N_1$ periods and obtain the dataset $\mathcal{F}_{N_1}$.
\State Obtain the output $\hat{\mathcal{I}}$ and $\hat{\mathcal{J}}$ from \Cref{alg:Idenbasis} with input $\mathcal{F}_{N_1}$ and $\eps=O(\veps^2/(K\cdot|\mathcal{S}||\mathcal{A}|))$.
\State Let $\hat{\mathcal{J}}_{1}=\hat{\mathcal{J}}\cap[K]$ and $\hat{\mathcal{J}}_{2}=\hat{\mathcal{J}}\cap\mathcal{S}$.
\State Initialize $\mathcal{H}_1=\emptyset$, $\bm{\alpha}^1=N_2\cdot \bm{\alpha}$, $\bmu^1=N_2\cdot\bmu$, and a state $s$ such that $(s,a)\in\hat{\mathcal{I}}$ for an $a$.
\For{$n=1,\dots, N_2$}
\State Construct estimates $\bar{C}^n(\hat{\mathcal{J}}_{1}, \hat{\mathcal{I}})$ and $\bar{B}^n(\hat{\mathcal{J}}_{2}, \hat{\mathcal{I}})$ using the sample set $\mathcal{H}_{n}$.
\State Construct a solution $\tilde{\bm{q}}^n$ such that $\tilde{\bm{q}}^n_{\hat{\mathcal{I}}^c}=0$ and $\tilde{\bm{q}}^*_{\hat{\mathcal{I}}}$ is the solution to
\begin{equation}\label{eqn:11OptQ2}
\begin{bmatrix}
\bar{C}^n(\hat{\mathcal{J}}_{1}, \hat{\mathcal{I}})\\
\bar{B}^n(\hat{\mathcal{J}}_{2}, \hat{\mathcal{I}})
\end{bmatrix}
\cdot\tilde{\bm{q}}^n_{\hat{\mathcal{I}}}=\begin{bmatrix}
\frac{\bm{\alpha}^n_{\hat{\mathcal{J}}_{1}}}{N_2-n+1}\\
\frac{\bmu^n_{\hat{\mathcal{J}}_2}}{N_2-n+1}
\end{bmatrix}.
\end{equation}
\State Project $\tilde{\bm{q}}^n$ to the set $\{\bm{q}: \|\bm{q}\|_1\leq 2, \bm{q}_{\hat{\mathcal{I}}^c}=0, \bm{q}_{\hat{\mathcal{I}}}\geq\xi'\}$ to obtain $\bm{q}^n$.
\State Construct a policy $\bar{\pi}^n$ such that 
\begin{equation}\label{eqn:barpin}
P(\bar{\pi}^n(s)=a) = \left\{
\begin{aligned}
&\frac{q^{n}(s,a)}{\sum_{a'\in\mathcal{A}}q^{n}(s,a')}, &&\text{if~}\sum_{a'\in\mathcal{A}}q^{n}(s,a')>0\\
&1/|\mathcal{A}|, &&\text{if~}\sum_{a'\in\mathcal{A}}q^{n}(s,a')=0,
\end{aligned}
\right.
\end{equation}
\State Follow the policy $\bar{\pi}^n$ until we visit each $(s,a)\in\hat{\mathcal{I}}$ or number of steps reaches the limit $N_3$. 
\State Collect sample of $r^n(s,a)$, $c^n_k(s,a)$ for each $k\in\hat{\mathcal{J}}_1\subset[K]$, and $s^n(s,a)$, for $(s,a)\in\hat{\mathcal{I}}$.
\State Update $\mathcal{H}_{n+1}=\mathcal{H}_n\cup\left\{ r^n(s,a), c^n_k(s,a), s^n(s,a), \forall (s,a)\in\hat{\mathcal{I}}, \forall k\in\hat{\mathcal{J}}_1\right\}$.
\State Denote by $\bm{c}^n(s,a)=(c^n_k(s,a))_{\forall k\in\hat{\mathcal{J}}_{1}}$ and do the update:
\begin{equation}\label{eqn:11UpdateAlpha2}
\bm{\alpha}^{n+1}_{\hat{\mathcal{J}}_{1}}=\bm{\alpha}^{n}_{\hat{\mathcal{J}}_{1}}-\sum_{(s,a)\in\hat{\mathcal{I}}}\bm{c}^n(s,a)\cdot q^n(s,a).
\end{equation}
\State Do the update:
\begin{equation}\label{eqn:11UpdateMu2}
\bmu^{n+1}_{s'}=\bmu^n_{s'}-\sum_{(s,a)\in\hat{\mathcal{I}}}q^n(s,a)\cdot(\delta_{s',s}-\gamma\bI_{\{s'=s^n(s,a)\}}),~~~\forall s'\in\hat{\mathcal{J}}_{2}.
\end{equation}
where $\bI_{\{s'=s^n(s,a)\}}$ is an indicator function of whether the state transition $s^n(s,a)$ equals $s'$.
\EndFor
\State \textbf{Output:} $\bm{q}^1, \dots, \bm{q}^{N_2}$.
\end{algorithmic}
\end{algorithm}

We now conduct the theoretical analysis of our algorithm. The analysis of the \Cref{alg:Idenbasis} in \Cref{alg:TwophaseOnline} is the same as \Cref{thm:Infibasis2}, and we can show that it holds $\hat{\mathcal{I}}=\mathcal{I}^*$ and $\hat{\mathcal{J}}=\mathcal{J}^*$ with a high probability. In order to analyze the resolving procedure, a key point is to characterize the distance between $\bm{q}^n$ and $\bm{q}^*$, and show that following the policy $\bar{\pi}^n$ induced by $\bm{q}^n$ would guarantee automatic exploration over the entire set $\mathcal{I}^*$ for collecting samples over costs and state transition. In fact, denoting by $\xi=\min_{(s,a)\in\mathcal{I}^*} \left\{ q^*(s,a) \right\}$, we can show that $\|\bm{q}^n - \bm{q}^*\|_{\infty}\leq \frac{\xi}{2}$. This would further imply that the induced policy $\pi^n$ guarantees that $P(\pi^n(s)=a)\geq\frac{\xi}{4}$ for every $(s,a)\in\mathcal{I}^*$ and hence each $(s,a)\in\mathcal{I}^*$ can be fully explored. 
We summarize the above arguments in the following lemma.
\begin{lemma}\label{lem:smallxi0}
Suppose that the sets $\hat{\mathcal{I}}$ and $\hat{\mathcal{J}}$ obtained in step 3 of \Cref{alg:TwophaseOnline} satisfy that $\hat{\mathcal{I}}=\mathcal{I}^*$ and $\hat{\mathcal{J}}=\mathcal{J}^*$.
Then, when $N_4\leq n\leq \tau_0$, for a constant $N_4$ and a stopping time $\tau_0$ defined as
\begin{equation}\label{eqn:tau0}
\tau_0=\min_{n\in[N_2]}\{ \left\|\tilde{\bm{\alpha}}_{\mathcal{J}^*_1}(n) - \bm{\alpha}_{\mathcal{J}^*_1}\right\|_{\infty}>\nu\text{~or~}\left\|\tilde{\bm{\mu}}_{\mathcal{J}^*_2}(n) - \bm{\mu}_{\mathcal{J}^*_2}\right\|_{\infty}>\nu \}
\end{equation}
for a constant $\nu$,
with a probability at least $1-\eps$, it holds that 
\[
\|\bm{q}^n - \bm{q}^*\|_{\infty}\leq\frac{\xi}{2}
\]
where $\bm{q}^n$ denotes the solution obtained from the resolving procedure of \Cref{alg:TwophaseOnline} and $\bm{q}^*$ denotes the optimal solution to LP \eqref{lp:standard3} corresponding to the basis $\mathcal{I}^*$ and $\mathcal{J}^*$.
\end{lemma}
In order to further bound the sample complexity, we denote by $\Pi$ the set of policies $\pi$ with the corresponding occupancy measure supported on $\mathcal{I}^*$ and it holds $P(\pi(s)=a)\geq\xi'$ for every $(s,a)\in\mathcal{I}^*$. We also denote by $t_{\mathrm{mix}}(\pi)$ the mixing time of the Markov chain corresponding to the policy $\pi$. We then define
\begin{equation}\label{eqn:mixingtime}
    t_{\mathrm{mix}} = \max_{\pi\in\Pi}\{t_{\mathrm{mix}}(\pi)\}. 
\end{equation}
Denoting by $\bm{q}^\pi$ the occupancy measure corresponding to a policy $\pi\in\Pi$, we define
\begin{equation}\label{eqn:SmallOccupMeasure}
    \xi_{\mathrm{small}} = \min_{\pi\in\Pi}\left\{ \min_{(s,a)\in\mathcal{I}^*}\{q^{\pi}(s,a)\} \right\}
\end{equation}
The largest mixing time $t_{\mathrm{mix}}$ and the smallest occupancy measure $\xi_{\mathrm{small}}$ will be used to bound by how many samples we are guaranteed to obtain at least one sample for each $(s,a)\in\mathcal{I}^*$ if we are following a policy $\bar{\pi}^n$ induced by $\bm{q}^n$ as defined in \eqref{eqn:barpin}. We assume that every Markov chain induced by any $\pi\in\Pi$ is ergodic. We finally use the output $\bm{q}^1, \dots, \bm{q}^{N_2}$ to construct the policy $\bar{\pi}$ by
\begin{equation}\label{eqn:11012501}
P(\bar{\pi}(s)=a) = \left\{
\begin{aligned}
&\frac{\bar{q}^{N_2}(s,a)}{\sum_{a'\in\mathcal{A}}\bar{q}^N(s,a')}, && \text{if~}\sum_{a'\in\mathcal{A}}\bar{q}^{N_2}(s,a')>0\\
&1/|\mathcal{A}|, &&\text{if~}\sum_{a'\in\mathcal{A}}\bar{q}^{N_2}(s,a')=0.
\end{aligned}\right.
\end{equation}
where $\bar{\bm{q}}^{N_2}=\frac{1}{N_2}\cdot\sum_{n=1}^{N_2}\bm{q}^n$.
We have the following sample complexity bound over the on-policy learning setting of our \Cref{alg:TwophaseOnline}.
\begin{theorem}\label{thm:onlinesample}
With a sample complexity bound of
\[
N=O\left( \frac{\log(1/\veps)}{\mu_{\min}\cdot\min\{\delta_1^2, \delta_2^2, \sigma_0^2\}}+\frac{t_{\mathrm{mix}}\cdot m^5\cdot\log(1/\xi)}{\xi_{\mathrm{small}}\cdot(\xi\sigma^*)^3(1-\gamma)}\cdot\frac{\log^2(1/\veps)}{\veps} \right),
\]
where $\mu_{\min}$ is defined in \eqref{eqn:Mumin}, $t_{\mathrm{mix}}$ is defined in \eqref{eqn:mixingtime}, and $\xi_{\mathrm{small}}$ is defined in \eqref{eqn:SmallOccupMeasure}, and
we obtain a policy $\bar{\pi}$ from \Cref{alg:Twophase} (defined in \eqref{eqn:11012501}) such that
\[
V_r(\pi^*, \mu_1) - V_r(\bar{\pi}, \mu_1)\leq\veps\text{~~and~~}V_k(\bar{\pi}, \mu_1) - \alpha_k\leq\veps,~\forall k\in[K].
\]
\end{theorem}

\begin{remark}
Note that in our \Cref{alg:TwophaseOnline}, we need to specify the number of rounds $N_1$ for carrying out \Cref{alg:Idenbasis}, the number of rounds $N_2$ for resolving, and the limit of the number of periods $N_3$ for sampling in each iteration of the resolving procedure. The value of $N_1$, $N_2$, and $N_3$ depends on some problem parameters which could be unknown to the decision maker. However, on one hand, we can estimate these problem parameters up to a constant level of accuracy with a high probability, which would not change the order of our sample complexity bound. On the other hand, we can adopt a standard doubling trick to avoid setting these values explicitly.
For example, in order to avoid setting the value of $N_3$ explicitly, in each iteration of the resolving procedure, we can initialize with some $N_3$ and exponentially increase the value of $N_3$ until we have at least one sample of $(s,a)\in\hat{\mathcal{I}}$. The final value of $N_3$ will be bounded by a constant with a high probability. 
\end{remark}




\section{Concluding Remarks}

In this work, we present a new algorithm that achieves logarithmic regrets or $\tilde{O}(1/\eps)$ sample complexity bound for the CMDP problems, for the first time in the literature. To achieve these improvements, we develop a new algorithmic framework to analyze the CMDP problems, motivated by the online LP literature. We consider the LP reformulation of the CMDP problems and we solve the LP in an online manner. Besides introducing the online LP framework, our method is new even compared to the online LP literature, as we do not require the non-degeneracy assumption to derive the logarithmic regret. In our research, we adopt the basic tabular setting and the infinite-horizon discounted setting to derive our results. We further extend to the finite-horizon episodic setting, off-policy learning setting, and the on-policy learning setting. There are also multiple ways to further extend our approaches. For example, one may incorporate the function approximation to deal with the continuous state-action space setting, or consider the setting where the state and the reward are not directly observed. There could also be other ways to characterize the instance hardness other than the LP-based corner point approach. We leave deriving instance-dependent guarantees for more involved settings in CMDP problems for future research.

\bibliographystyle{abbrvnat}
\bibliography{bibliography}

@book{stewart1998perturbation,
  title={Perturbation theory for the singular value decomposition},
  author={Stewart, Gilbert W},
  year={1998},
  publisher={Digital repository at the University of Maryland}
}

@article{megiddo1991finding,
  title={On finding primal-and dual-optimal bases},
  author={Megiddo, Nimrod},
  journal={ORSA Journal on Computing},
  volume={3},
  number={1},
  pages={63--65},
  year={1991},
  publisher={INFORMS}
}

@article{alvo2023neural,
  title={Neural inventory control in networks via hindsight differentiable policy optimization},
  author={Alvo, Matias and Russo, Daniel and Kanoria, Yash},
  journal={arXiv preprint arXiv:2306.11246},
  year={2023}
}

@inproceedings{bhandari2021linear,
  title={On the linear convergence of policy gradient methods for finite mdps},
  author={Bhandari, Jalaj and Russo, Daniel},
  booktitle={International Conference on Artificial Intelligence and Statistics},
  pages={2386--2394},
  year={2021},
  organization={PMLR}
}

@article{chen2024landscape,
  title={Landscape of Policy Optimization for Finite Horizon MDPs with General State and Action},
  author={Chen, Xin and Hu, Yifan and Zhao, Minda},
  journal={arXiv preprint arXiv:2409.17138},
  year={2024}
}

@inproceedings{wei2023constant,
  title={Constant regret primal-dual policy for multi-way dynamic matching},
  author={Wei, Yehua and Xu, Jiaming and Yu, Sophie H},
  booktitle={Abstract Proceedings of the 2023 ACM SIGMETRICS International Conference on Measurement and Modeling of Computer Systems},
  pages={79--80},
  year={2023}
}

@article{banerjee2024good,
  title={Good prophets know when the end is near},
  author={Banerjee, Siddhartha and Freund, Daniel},
  journal={Management Science},
  year={2024},
  publisher={INFORMS}
}

@article{paulin2015concentration,
  title={Concentration inequalities for Markov chains by Marton couplings and spectral methods},
  author={Paulin, Daniel},
  journal={Electronic Journal of Probability},
  volume={20},
  year={2015}
}

@article{li2021sample,
  title={Sample complexity of asynchronous Q-learning: Sharper analysis and variance reduction},
  author={Li, Gen and Wei, Yuting and Chi, Yuejie and Gu, Yuantao and Chen, Yuxin},
  journal={IEEE Transactions on Information Theory},
  volume={68},
  number={1},
  pages={448--473},
  year={2021},
  publisher={IEEE}
}

@article{chen2021primal,
  title={A primal-dual approach to constrained markov decision processes},
  author={Chen, Yi and Dong, Jing and Wang, Zhaoran},
  journal={arXiv preprint arXiv:2101.10895},
  year={2021}
}

@article{sun2024constrained,
  title={Constrained Reinforcement Learning Under Model Mismatch},
  author={Sun, Zhongchang and He, Sihong and Miao, Fei and Zou, Shaofeng},
  journal={arXiv preprint arXiv:2405.01327},
  year={2024}
}

@article{ao2024online,
  title={Online Local False Discovery Rate Control: A Resource Allocation Approach},
  author={Ao, Ruicheng and Chen, Hongyu and Simchi-Levi, David and Zhu, Feng},
  journal={arXiv preprint arXiv:2402.11425},
  year={2024}
}

@article{duan2024taming,
  title={Taming" data-hungry" reinforcement learning? Stability in continuous state-action spaces},
  author={Duan, Yaqi and Wainwright, Martin J},
  journal={arXiv preprint arXiv:2401.05233},
  year={2024}
}

@article{li2020breaking,
  title={Breaking the sample size barrier in model-based reinforcement learning with a generative model},
  author={Li, Gen and Wei, Yuting and Chi, Yuejie and Gu, Yuantao and Chen, Yuxin},
  journal={Operations Research},
  volume={72},
  number={1},
  pages={203--221},
  year={2024}
}

@book{kakade2003sample,
  title={On the sample complexity of reinforcement learning},
  author={Kakade, Sham Machandranath},
  year={2003},
  publisher={University of London, University College London (United Kingdom)}
}

@article{kearns2002sparse,
  title={A sparse sampling algorithm for near-optimal planning in large Markov decision processes},
  author={Kearns, Michael and Mansour, Yishay and Ng, Andrew Y},
  journal={Machine learning},
  volume={49},
  pages={193--208},
  year={2002},
  publisher={Springer}
}

@article{foster2020instance,
  title={Instance-dependent complexity of contextual bandits and reinforcement learning: A disagreement-based perspective},
  author={Foster, Dylan J and Rakhlin, Alexander and Simchi-Levi, David and Xu, Yunzong},
  journal={arXiv preprint arXiv:2010.03104},
  year={2020}
}

@article{xu2024towards,
  title={Towards optimal problem dependent generalization error bounds in statistical learning theory},
  author={Xu, Yunbei and Zeevi, Assaf},
  journal={Mathematics of Operations Research},
  year={2024},
  publisher={INFORMS}
}

@article{graves1997asymptotically,
  title={Asymptotically efficient adaptive choice of control laws incontrolled markov chains},
  author={Graves, Todd L and Lai, Tze Leung},
  journal={SIAM journal on control and optimization},
  volume={35},
  number={3},
  pages={715--743},
  year={1997},
  publisher={SIAM}
}

@article{li2023revenue,
  title={Revenue Management with Calendar-Aware and Dependent Demands: Asymptotically Tight Fluid Approximations},
  author={Li, Weiyun and Rusmevichientong, Paat and Topaloglu, Huseyin},
  journal={Available at SSRN 4543277},
  year={2023}
}

@article{jiang2023constant,
  title={Constant Approximation for Network Revenue Management with Markovian-Correlated Customer Arrivals},
  author={Jiang, Jiashuo},
  journal={arXiv preprint arXiv:2305.05829},
  year={2023}
}

@inproceedings{jia2023online,
  title={Online resource allocation in Markov Chains},
  author={Jia, Jianhao and Li, Hao and Liu, Kai and Liu, Ziqi and Zhou, Jun and Gravin, Nikolai and Tang, Zhihao Gavin},
  booktitle={Proceedings of the ACM Web Conference 2023},
  pages={3498--3507},
  year={2023}
}

@article{wainwright2019variance,
  title={Variance-reduced $ Q $-learning is minimax optimal},
  author={Wainwright, Martin J},
  journal={arXiv preprint arXiv:1906.04697},
  year={2019}
}

@article{sidford2018near,
  title={Near-optimal time and sample complexities for solving Markov decision processes with a generative model},
  author={Sidford, Aaron and Wang, Mengdi and Wu, Xian and Yang, Lin and Ye, Yinyu},
  journal={Advances in Neural Information Processing Systems},
  volume={31},
  year={2018}
}

@inproceedings{agarwal2020model,
  title={Model-based reinforcement learning with a generative model is minimax optimal},
  author={Agarwal, Alekh and Kakade, Sham and Yang, Lin F},
  booktitle={Conference on Learning Theory},
  pages={67--83},
  year={2020},
  organization={PMLR}
}

@article{he2021nearly,
  title={Nearly minimax optimal reinforcement learning for discounted MDPs},
  author={He, Jiafan and Zhou, Dongruo and Gu, Quanquan},
  journal={Advances in Neural Information Processing Systems},
  volume={34},
  pages={22288--22300},
  year={2021}
}

@inproceedings{lattimore2012pac,
  title={PAC bounds for discounted MDPs},
  author={Lattimore, Tor and Hutter, Marcus},
  booktitle={Algorithmic Learning Theory: 23rd International Conference, ALT 2012, Lyon, France, October 29-31, 2012. Proceedings 23},
  pages={320--334},
  year={2012},
  organization={Springer}
}

@article{gheshlaghi2013minimax,
  title={Minimax PAC bounds on the sample complexity of reinforcement learning with a generative model},
  author={Gheshlaghi Azar, Mohammad and Munos, R{\'e}mi and Kappen, Hilbert J},
  journal={Machine learning},
  volume={91},
  pages={325--349},
  year={2013},
  publisher={Springer}
}

@article{auer2008near,
  title={Near-optimal regret bounds for reinforcement learning},
  author={Auer, Peter and Jaksch, Thomas and Ortner, Ronald},
  journal={Advances in neural information processing systems},
  volume={21},
  year={2008}
}

@article{jiang2022degeneracy,
  title={Degeneracy is OK: Logarithmic Regret for Network Revenue Management with Indiscrete Distributions},
  author={Jiang, Jiashuo and Ma, Will and Zhang, Jiawei},
  journal={arXiv preprint arXiv:2210.07996},
  year={2022}
}

@book{hiriart1996convex,
  title={Convex analysis and minimization algorithms I: Fundamentals},
  author={Hiriart-Urruty, Jean-Baptiste and Lemar{\'e}chal, Claude},
  volume={305},
  year={1996},
  publisher={Springer science \& business media}
}

@article{nedic2009subgradient,
  title={Subgradient methods for saddle-point problems},
  author={Nedi{\'c}, Angelia and Ozdaglar, Asuman},
  journal={Journal of optimization theory and applications},
  volume={142},
  pages={205--228},
  year={2009},
  publisher={Springer}
}

@inproceedings{jin2020efficiently,
  title={Efficiently solving MDPs with stochastic mirror descent},
  author={Jin, Yujia and Sidford, Aaron},
  booktitle={International Conference on Machine Learning},
  pages={4890--4900},
  year={2020},
  organization={PMLR}
}

@article{wang2020randomized,
  title={Randomized linear programming solves the Markov decision problem in nearly linear (sometimes sublinear) time},
  author={Wang, Mengdi},
  journal={Mathematics of Operations Research},
  volume={45},
  number={2},
  pages={517--546},
  year={2020},
  publisher={INFORMS}
}

@inproceedings{wagenmaker2022beyond,
  title={Beyond no regret: Instance-dependent pac reinforcement learning},
  author={Wagenmaker, Andrew J and Simchowitz, Max and Jamieson, Kevin},
  booktitle={Conference on Learning Theory},
  pages={358--418},
  year={2022},
  organization={PMLR}
}

@article{dann2021beyond,
  title={Beyond value-function gaps: Improved instance-dependent regret bounds for episodic reinforcement learning},
  author={Dann, Christoph and Marinov, Teodor Vanislavov and Mohri, Mehryar and Zimmert, Julian},
  journal={Advances in Neural Information Processing Systems},
  volume={34},
  pages={1--12},
  year={2021}
}

@article{simchowitz2019non,
  title={Non-asymptotic gap-dependent regret bounds for tabular mdps},
  author={Simchowitz, Max and Jamieson, Kevin G},
  journal={Advances in Neural Information Processing Systems},
  volume={32},
  year={2019}
}

@inproceedings{zanette2019tighter,
  title={Tighter problem-dependent regret bounds in reinforcement learning without domain knowledge using value function bounds},
  author={Zanette, Andrea and Brunskill, Emma},
  booktitle={International Conference on Machine Learning},
  pages={7304--7312},
  year={2019},
  organization={PMLR}
}

@inproceedings{wagenmaker2023instance,
  title={Instance-optimality in interactive decision making: Toward a non-asymptotic theory},
  author={Wagenmaker, Andrew J and Foster, Dylan J},
  booktitle={The Thirty Sixth Annual Conference on Learning Theory},
  pages={1322--1472},
  year={2023},
  organization={PMLR}
}

@article{zhou2022anchor,
  title={Anchor-changing regularized natural policy gradient for multi-objective reinforcement learning},
  author={Zhou, Ruida and Liu, Tao and Kalathil, Dileep and Kumar, PR and Tian, Chao},
  journal={Advances in Neural Information Processing Systems},
  volume={35},
  pages={13584--13596},
  year={2022}
}

@article{liu2021policy,
  title={Policy Optimization for Constrained MDPs with Provable Fast Global Convergence},
  author={Liu, Tao and Zhou, Ruida and Kalathil, Dileep and Kumar, PR and Tian, Chao},
  journal={arXiv preprint arXiv:2111.00552},
  year={2021}
}

@article{flajolet2015logarithmic,
  title={Logarithmic regret bounds for bandits with knapsacks},
  author={Flajolet, Arthur and Jaillet, Patrick},
  journal={arXiv preprint arXiv:1510.01800},
  year={2015}
}

@article{lai1985asymptotically,
  title={Asymptotically efficient adaptive allocation rules},
  author={Lai, Tze Leung and Robbins, Herbert},
  journal={Advances in applied mathematics},
  volume={6},
  number={1},
  pages={4--22},
  year={1985},
  publisher={Academic Press}
}

@article{chow2018lyapunov,
  title={A lyapunov-based approach to safe reinforcement learning},
  author={Chow, Yinlam and Nachum, Ofir and Duenez-Guzman, Edgar and Ghavamzadeh, Mohammad},
  journal={Advances in neural information processing systems},
  volume={31},
  year={2018}
}

@article{li2021faster,
  title={Faster algorithm and sharper analysis for constrained markov decision process},
  author={Li, Tianjiao and Guan, Ziwei and Zou, Shaofeng and Xu, Tengyu and Liang, Yingbin and Lan, Guanghui},
  journal={arXiv preprint arXiv:2110.10351},
  year={2021}
}

@inproceedings{gladin2023algorithm,
  title={Algorithm for constrained Markov decision process with linear convergence},
  author={Gladin, Egor and Lavrik-Karmazin, Maksim and Zainullina, Karina and Rudenko, Varvara and Gasnikov, Alexander and Takac, Martin},
  booktitle={International Conference on Artificial Intelligence and Statistics},
  pages={11506--11533},
  year={2023},
  organization={PMLR}
}

@inproceedings{ding2021provably,
  title={Provably efficient safe exploration via primal-dual policy optimization},
  author={Ding, Dongsheng and Wei, Xiaohan and Yang, Zhuoran and Wang, Zhaoran and Jovanovic, Mihailo},
  booktitle={International Conference on Artificial Intelligence and Statistics},
  pages={3304--3312},
  year={2021},
  organization={PMLR}
}

@article{liu2021learning,
  title={Learning policies with zero or bounded constraint violation for constrained mdps},
  author={Liu, Tao and Zhou, Ruida and Kalathil, Dileep and Kumar, Panganamala and Tian, Chao},
  journal={Advances in Neural Information Processing Systems},
  volume={34},
  pages={17183--17193},
  year={2021}
}

@article{li2022online,
  title={Online linear programming: Dual convergence, new algorithms, and regret bounds},
  author={Li, Xiaocheng and Ye, Yinyu},
  journal={Operations Research},
  volume={70},
  number={5},
  pages={2948--2966},
  year={2022},
  publisher={INFORMS}
}

@inproceedings{achiam2017constrained,
  title={Constrained policy optimization},
  author={Achiam, Joshua and Held, David and Tamar, Aviv and Abbeel, Pieter},
  booktitle={International conference on machine learning},
  pages={22--31},
  year={2017},
  organization={PMLR}
}

@inproceedings{tamar2012policy,
  title={Policy gradients with variance related risk criteria},
  author={Tamar, Aviv and Di Castro, Dotan and Mannor, Shie},
  booktitle={Proceedings of the twenty-ninth international conference on machine learning},
  pages={387--396},
  year={2012}
}

@article{ono2015chance,
  title={Chance-constrained dynamic programming with application to risk-aware robotic space exploration},
  author={Ono, Masahiro and Pavone, Marco and Kuwata, Yoshiaki and Balaram, J},
  journal={Autonomous Robots},
  volume={39},
  pages={555--571},
  year={2015},
  publisher={Springer}
}

@article{fisac2018general,
  title={A general safety framework for learning-based control in uncertain robotic systems},
  author={Fisac, Jaime F and Akametalu, Anayo K and Zeilinger, Melanie N and Kaynama, Shahab and Gillula, Jeremy and Tomlin, Claire J},
  journal={IEEE Transactions on Automatic Control},
  volume={64},
  number={7},
  pages={2737--2752},
  year={2018},
  publisher={IEEE}
}

@article{kober2013reinforcement,
  title={Reinforcement learning in robotics: A survey},
  author={Kober, Jens and Bagnell, J Andrew and Peters, Jan},
  journal={The International Journal of Robotics Research},
  volume={32},
  number={11},
  pages={1238--1274},
  year={2013},
  publisher={SAGE Publications Sage UK: London, England}
}

@article{shani2005mdp,
  title={An MDP-based recommender system.},
  author={Shani, Guy and Heckerman, David and Brafman, Ronen I and Boutilier, Craig},
  journal={Journal of Machine Learning Research},
  volume={6},
  number={9},
  year={2005}
}

@article{mnih2013playing,
  title={Playing atari with deep reinforcement learning},
  author={Mnih, Volodymyr and Kavukcuoglu, Koray and Silver, David and Graves, Alex and Antonoglou, Ioannis and Wierstra, Daan and Riedmiller, Martin},
  journal={arXiv preprint arXiv:1312.5602},
  year={2013}
}

@book{higham2002accuracy,
  title={Accuracy and stability of numerical algorithms},
  author={Higham, Nicholas J},
  year={2002},
  publisher={SIAM}
}

@inproceedings{li2021symmetry,
  title={The symmetry between arms and knapsacks: A primal-dual approach for bandits with knapsacks},
  author={Li, Xiaocheng and Sun, Chunlin and Ye, Yinyu},
  booktitle={International Conference on Machine Learning},
  pages={6483--6492},
  year={2021},
  organization={PMLR}
}

@article{efroni2020exploration,
  title={Exploration-exploitation in constrained mdps},
  author={Efroni, Yonathan and Mannor, Shie and Pirotta, Matteo},
  journal={arXiv preprint arXiv:2003.02189},
  year={2020}
}

@article{bhattacharya2017linear,
  title={Linear programming formulation for non-stationary, finite-horizon Markov decision process models},
  author={Bhattacharya, Arnab and Kharoufeh, Jeffrey P},
  journal={Operations Research Letters},
  volume={45},
  number={6},
  pages={570--574},
  year={2017},
  publisher={Elsevier}
}

@article{zimin2013online,
  title={Online learning in episodic Markovian decision processes by relative entropy policy search},
  author={Zimin, Alexander and Neu, Gergely},
  journal={Advances in neural information processing systems},
  volume={26},
  year={2013}
}

@book{altman1999constrained,
  title={Constrained Markov decision processes},
  author={Altman, Eitan},
  year={1999},
  publisher={vol7.~CRCPress}
}

@article{ma2022optimal,
  title={Optimal Regularized Online Convex Allocation by Adaptive Re-Solving},
  author={Ma, Wanteng and Cao, Ying and Tsang, Danny HK and Xia, Dong},
  journal={arXiv preprint arXiv:2209.00399},
  year={2022}
}

@article{jasin2012re,
  title={A re-solving heuristic with bounded revenue loss for network revenue management with customer choice},
  author={Jasin, Stefanus and Kumar, Sunil},
  journal={Mathematics of Operations Research},
  volume={37},
  number={2},
  pages={313--345},
  year={2012},
  publisher={INFORMS}
}

@article{bumpensanti2020re,
  title={A re-solving heuristic with uniformly bounded loss for network revenue management},
  author={Bumpensanti, Pornpawee and Wang, He},
  journal={Management Science},
  volume={66},
  number={7},
  pages={2993--3009},
  year={2020},
  publisher={INFORMS}
}

@article{vera2021bayesian,
  title={The bayesian prophet: A low-regret framework for online decision making},
  author={Vera, Alberto and Banerjee, Siddhartha},
  journal={Management Science},
  volume={67},
  number={3},
  pages={1368--1391},
  year={2021},
  publisher={INFORMS}
}

@article{arlotto2019uniformly,
  title={Uniformly bounded regret in the multisecretary problem},
  author={Arlotto, Alessandro and Gurvich, Itai},
  journal={Stochastic Systems},
  volume={9},
  number={3},
  pages={231--260},
  year={2019},
  publisher={INFORMS}
}

@article{vera2021online,
  title={Online allocation and pricing: Constant regret via bellman inequalities},
  author={Vera, Alberto and Banerjee, Siddhartha and Gurvich, Itai},
  journal={Operations Research},
  volume={69},
  number={3},
  pages={821--840},
  year={2021},
  publisher={INFORMS}
}

@article{molinaro2014geometry,
  title={The geometry of online packing linear programs},
  author={Molinaro, Marco and Ravi, Ramamoorthi},
  journal={Mathematics of Operations Research},
  volume={39},
  number={1},
  pages={46--59},
  year={2014},
  publisher={INFORMS}
}

@article{agrawal2014dynamic,
  title={A dynamic near-optimal algorithm for online linear programming},
  author={Agrawal, Shipra and Wang, Zizhuo and Ye, Yinyu},
  journal={Operations Research},
  volume={62},
  number={4},
  pages={876--890},
  year={2014},
  publisher={INFORMS}
}

@inproceedings{gupta2014experts,
  title={How experts can solve LPs online},
  author={Gupta, Anupam and Molinaro, Marco},
  booktitle={European Symposium on Algorithms},
  pages={517--529},
  year={2014},
  organization={Springer}
}

@article{ferguson1989solved,
  title={Who solved the secretary problem?},
  author={Ferguson, Thomas S},
  journal={Statistical science},
  volume={4},
  number={3},
  pages={282--289},
  year={1989},
  publisher={Institute of Mathematical Statistics}
}

@article{arlotto2020logarithmic,
  title={Logarithmic regret in the dynamic and stochastic knapsack problem with equal rewards},
  author={Arlotto, Alessandro and Xie, Xinchang},
  journal={Stochastic Systems},
  volume={10},
  number={2},
  pages={170--191},
  year={2020},
  publisher={INFORMS}
}

@article{buchbinder2009online,
  title={Online primal-dual algorithms for covering and packing},
  author={Buchbinder, Niv and Naor, Joseph},
  journal={Mathematics of Operations Research},
  volume={34},
  number={2},
  pages={270--286},
  year={2009},
  publisher={INFORMS}
}

@article{mehta2007adwords,
  title={Adwords and generalized online matching},
  author={Mehta, Aranyak and Saberi, Amin and Vazirani, Umesh and Vazirani, Vijay},
  journal={Journal of the ACM (JACM)},
  volume={54},
  number={5},
  pages={22--es},
  year={2007},
  publisher={ACM New York, NY, USA}
}

@inproceedings{kesselheim2014primal,
  title={Primal beats dual on online packing LPs in the random-order model},
  author={Kesselheim, Thomas and T{\"o}nnis, Andreas and Radke, Klaus and V{\"o}cking, Berthold},
  booktitle={Proceedings of the forty-sixth annual ACM symposium on Theory of computing},
  pages={303--312},
  year={2014}
}

@article{jasin2014reoptimization,
  title={Reoptimization and self-adjusting price control for network revenue management},
  author={Jasin, Stefanus},
  journal={Operations Research},
  volume={62},
  number={5},
  pages={1168--1178},
  year={2014},
  publisher={INFORMS}
}

\clearpage

\OneAndAHalfSpacedXI

%
%
%

\begin{APPENDICES}
\crefalias{section}{appendix}

\section{Numerical Studies}

We implement our \Cref{alg:Twophase} to study the numerical performance. We consider a CMDP problem with the state space $|\mathcal{S}|=10$ and the action space $|\mathcal{A}|=10$. We set the discount factor $\gamma=0.7$. We then randomly generate the probability transition kernel $P$. To be specific, for each state $s\in\mathcal{S}$, action $a\in\mathcal{A}$, and the future state $s'\in\mathcal{S}$, we uniformly generate a randomly variable $p_{s,a,s'}$. Then, the transition probability is defined as $P(s'|s,a)=\frac{p_{s,a,s'}}{\sum_{s''\in\mathcal{S}}p_{s,a,s''}}$. For each state-action pair $(s,a)\in\mathcal{S}\times\mathcal{A}$, the expected reward $\hat{r}(s,a)$ is uniformly generated from the interval $[1,2]$ (with the reward for the first action set to be $0$). The actual reward $r(s,a)=\hat{r}(s,a)+\eta$,  where $\eta$ is uniformly distributed among $[-0.5, 0.5]$. There are $K=5$ constraints and for each constraint $k\in[K]$ and each state-action pair $(s,a)\in\mathcal{S}\times\mathcal{A}$, we define the expected cost $\hat{c}_k(s,a)$ to be uniformly generated from $[1,2]$. The actual cost $c_k(s,a)=\hat{c}_k(s,a)+\eta'$, where $\eta'$ is uniformly distributed among $[-0.5, 0.5]$.

For each total iterations $N$, We apply \Cref{alg:Twophase} and obtain the output $\bm{q}^1, \dots, \bm{q}^N$. We compare $\bar{\bm{q}}^N$ with the optimal occupancy measure. Since our algorithm is a randomized algorithm, we study the performance of our algorithm in expectation. To be specific, given the problem instance and a fixed $N$, we implement our algorithm repeatedly for $M=500$ rounds. Denote by $\bar{\bm{q}}^N_m$ the output of our \Cref{alg:Twophase} at round $m$, for $m\in[M]$. We define the error term as $\text{Err}(N)=\frac{1}{M}\cdot\sum_{m=1}^M\|\bar{\bm{q}}_m^N-\bm{q}^*\|_{1}/\|\bm{q}^*\|_1$. We study how the error term $\text{Err}(N)$ scales with $N$. The results are displayed in \Cref{fig:02}. As we can see, the error term $\text{Err}(N)$ decreases quickly with respect to $N$. Moreover, since our \Cref{alg:Twophase} requires only solving a set of linear equations in each iteration, the computation cost of our \Cref{alg:Twophase} is also moderate.

\begin{figure}
    \centering
    \includegraphics[width=0.5\linewidth]{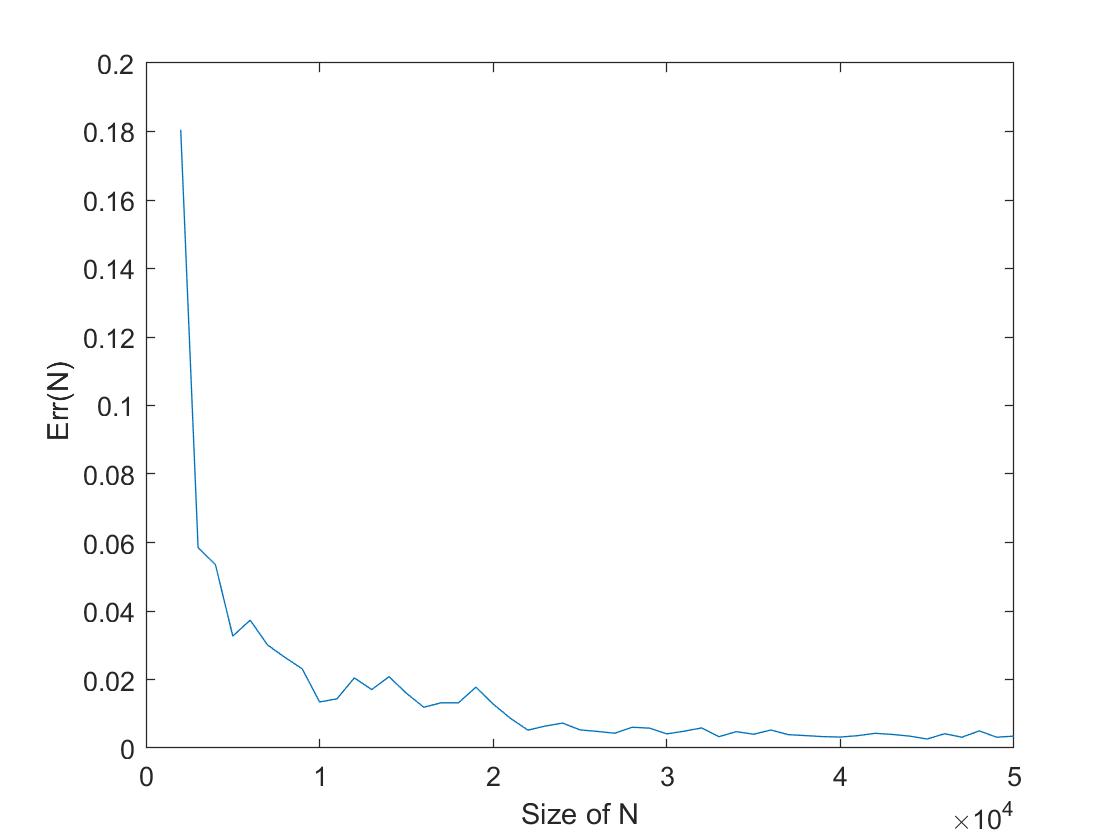}
    \caption{The computational performance of \Cref{alg:Twophase}. The x-label denotes the size of $N$, while the y-label denotes the error term $\text{Err}(N)$.}
    \label{fig:02}
\end{figure}

\section{Missing Proofs in \Cref{sec:reformulation}}

\subsection{Proof of Lemma \ref{lem:InfiBasis}}\label{pf:Lemma1}

We can rewrite the LP \eqref{lp:standard3} into the standard formulation given as follows
\begin{equation}\label{eqn:110301}
V=\max~~\hat{\bm{r}}^\top\bm{q}~~\mbox{s.t.}~C\bm{q}+\bm{p}=\bm{\alpha}, ~B\bm{q}=\bmu, ~\bm{q}\geq0, ~\bm{p}\geq0,
\end{equation}
where we introduce the variable $\bm{p}\in\mathbb{R}^K$ to serve as the slackness variables to the cost constraints. Denote by $\mathcal{I}'$ an optimal basis to the LP \eqref{eqn:110301} in the standard form. We know that the basis $\mathcal{I}'$ can be decomposed into two parts. We denote by $\mathcal{I}'_1=\mathcal{I}'\cap[K]$ the basic variables belonging to the slack variables $\bm{p}$. We then denote by $\mathcal{I}'_2=\mathcal{I}'\cap(\mathcal{S}\times\mathcal{A})$ the basic variables belonging to the decision variables $\bm{q}$. Then, for the optimal basic solution corresponding to the optimal basis $\mathcal{I}'$, which we denote by $(\bm{q}^*, \bm{p}^*)$, we know that the following linear equations must hold
\begin{equation}\label{eqn:110302}
    C\bm{q}^*+\bm{p}^*=\bm{\alpha}, ~B\bm{q}^*=\bmu, ~\bm{p}^*_{(\mathcal{I}'_1)^c}=0, ~\bm{q}^*_{(\mathcal{I}'_2)^c}=0.
\end{equation}
Moreover, we know that $(\bm{q}^*, \bm{p}^*)$ is the unique solution to the linear system described in \eqref{eqn:110302}.

We note that from the definition of the basis set for LP in the standard formulation, it holds $|\mathcal{I}'|=K+|\mathcal{S}|=|\mathcal{I}'_1|+|\mathcal{I}'_2|$. Then, it holds that $|\mathcal{I}'_2|=K-|\mathcal{I}'_1|+|\mathcal{S}|=|(\mathcal{I}'_1)^c|+|\mathcal{S}|$. We know that the linear system
\begin{equation}\label{eqn:110303}
    C((\mathcal{I}'_1)^c, \mathcal{I}'_2)\bm{q}_{\mathcal{I}'_2}=\bm{\alpha}_{(\mathcal{I}'_1)^c},~C(\mathcal{I}'_1, \mathcal{I}'_2)\bm{q}_{\mathcal{I}'_2}+\bm{p}^*_{\mathcal{I}'_1}=\bm{\alpha}_{\mathcal{I}'_1},~B(:,\mathcal{I}'_2)\bm{q}^*_{\mathcal{I}'_2}=\bmu
\end{equation}
also enjoys a unique solution as $(\bm{p}^*_{\mathcal{I}'_1}, \bm{q}^*_{\mathcal{I}'_2})$. Otherwise, if there exists a different solution to the linear system \eqref{eqn:110303}, we can append the solution $\bm{p}^*_{(\mathcal{I}'_1)^c}=0, ~\bm{q}^*_{(\mathcal{I}'_2)^c}=0$ to the other solution to obtain a new solution to the linear system \eqref{eqn:110302}, which would violate the uniqueness of the solution to the linear system \eqref{eqn:110302}. 

The uniqueness of the solution to the linear system \eqref{eqn:110303} implies that all the constraint vectors in the linear system \eqref{eqn:110303} are linearly independent of each other, since we have $|\mathcal{I}'_2|=|(\mathcal{I}'_1)^c|+|\mathcal{S}|$ (the number of constraints equals the number of variables in the linear system \eqref{eqn:110303}). Therefore, we know that the constraint vectors in the following linear system, which is a subset of the constraint vectors in the linear system \eqref{eqn:110303}, are also linearly independent of each other,
\begin{equation}\label{eqn:110304}
    C((\mathcal{I}'_1)^c, \mathcal{I}'_2)\bm{q}_{\mathcal{I}'_2}=\bm{\alpha}_{(\mathcal{I}'_1)^c},~B(:,\mathcal{I}'_2)\bm{q}^*_{\mathcal{I}'_2}=\bmu.
\end{equation}
Also, from $|\mathcal{I}'_2|=|(\mathcal{I}'_1)^c|+|\mathcal{S}|$, we note that the number of constraints equals the number of decision variables in the linear system \eqref{eqn:110304}. In this way, we show that the linear system \eqref{eqn:110304} enjoys a unique solution, which is $\bm{q}^*_{\mathcal{I}'_2}$, and the
square matrix 
\begin{equation}\label{eqn:110305}
    A'=\begin{bmatrix}
        C((\mathcal{I}'_1)^c, \mathcal{I}'_2)\\
        B(:,\mathcal{I}'_2)
    \end{bmatrix}
\end{equation}
is a full rank matrix. We can further denote by $\mathcal{I}''_2$ the support set of $\bm{q}^*$ such that $\mathcal{I}''_2\subset\mathcal{I}'_2$ and it holds $\bm{q}^*_{\mathcal{I}''_2}>0$. Since the columns of the matrix $A'$ in \eqref{eqn:110305} are linearly independent of each other, we know that the matrix
\begin{equation}\label{eqn:110306}
    A''=\begin{bmatrix}
        C((\mathcal{I}'_1)^c, \mathcal{I}''_2)\\
        B(:,\mathcal{I}''_2)
    \end{bmatrix}
\end{equation}
is of full column-rank. Also, we know that $\bm{q}^*_{\mathcal{I}''_2}$ can be expressed as the solution to the linear system
\begin{equation}\label{eqn:110307}
    A''\bm{q}^*_{\mathcal{I}''_2}=\begin{bmatrix}
        C((\mathcal{I}'_1)^c, \mathcal{I}''_2)\\
        B(:,\mathcal{I}''_2)
    \end{bmatrix}\bm{q}^*_{\mathcal{I}''_2} = \begin{bmatrix}
        \bm{\alpha}_{(\mathcal{I}'_1)^c}\\
        \bmu
    \end{bmatrix}.
\end{equation}
Now we focus on the linear system in \eqref{eqn:110307}. It has $|(\mathcal{I}'_1)^c|+|\mathcal{S}|=|\mathcal{I}'_2|$ number of constraints and has $|\mathcal{I}''_2|$ number of decision variables. In other words, the linear system \eqref{eqn:110307} has more constraints than the variables. Also, we know that the matrix $A''$ is of full column-rank. This implies that we can find an index subset of the rows of $A''$, which we denote by $\mathcal{J}^*$, such that $|\mathcal{J}^*|=|\mathcal{I}''_2|$ and the square matrix $A''(\mathcal{J}^*,:)$ is of full rank. Since we also have
\begin{equation}\label{eqn:110308}
    A''(\mathcal{J}^*, :)\bm{q}^*_{\mathcal{I}''_2} = \begin{bmatrix}
        \bm{\alpha}_{(\mathcal{I}'_1)^c}\\
        \bmu
    \end{bmatrix}
\end{equation}
and the matrix $A''(\mathcal{J}^*, :)$ is a full rank square matrix. We know that after specifying $\mathcal{I}^*=\mathcal{I}''_2$, the sets $\mathcal{I}^*$ and $\mathcal{J}^*$ and the solution $\bm{q}^*$ would satisfy the conditions in \Cref{lem:InfiBasis}. Our proof is thus completed.

\section{Missing Proofs in \Cref{sec:Estimates}}

\subsection{Proof of Lemma \ref{lem:LUbound}}\label{pf:Lemma2}
We now condition on the event that $|\bar{r}_{N_0}(s,a)-\hat{r}(s,a)|$, $|\bar{c}_{k, N_0}(s,a)-\hat{c}_k(s,a)|$, and $|\bar{P}_{N_0}(s'|s,a)-P(s'|s,a)|$ for each $k\in[K]$, $(s,a)\in\mathcal{S}\times\mathcal{A}$ and $s'\in\mathcal{S}$ are all bounded by $\Rad(N_0, \eps)$. From the union bound, we know that this event happens with probability at least $1-(K|\mathcal{S}||\mathcal{A}|-|\mathcal{S}|^2|\mathcal{A}|)\cdot\eps$.

Note that for LP \eqref{lp:Infinite}, by summing up the constraint \eqref{eqn:Inficonstraint2} for all $s\in\mathcal{S}$, we obtain that any feasible solution $\bm{q}$ for LP \eqref{lp:Infinite} would satisfy
\begin{equation}\label{eqn:013101}
    \|\bm{q}\|_1=1.
\end{equation}

We first upper bound the gap $V-\bar{V}_{N_0}$. Denote by $\bm{q}^*$ one optimal solution to $V$. Then, from the feasibility of $\bm{q}^*$, we know that
\begin{equation}\label{eqn:013103}
\bar{C}_{N_0}\bm{q}^*=C\bm{q}^*+( \bar{C}_{N_0}-C)\bm{q}^*\leq \bm{\alpha}+\Rad(N_0,\eps)\leq\bm{\alpha}+\lambda_{N_0}
\end{equation}
where the first inequality follows from $\|\bm{q}^*\|_1=1$ \eqref{eqn:013101} and all elements of $\bar{C}_{N_0}-C$ are upper bounded by $\Rad(N_0, \eps)$. Also, we know that
\begin{equation}\label{eqn:013104}
\bar{B}_{N_0}\bm{q}^*=  B\bm{q}^*+(\bar{B}_{N_0}-B)\bm{q}^*\leq (1-\gamma)\cdot\bmu+\gamma\cdot\Rad(N_0,\eps)\leq (1-\gamma)\cdot\bmu+\lambda_{N_0}
\end{equation}
where the first inequality follows from $\|\bm{q}^*\|_1=1$ \eqref{eqn:013101} and all elements of $\bar{B}_{N_0}-B$ are upper bounded by $\gamma\cdot\Rad(N_0, \eps)$. Similarly, we have that
\begin{equation}\label{eqn:0013104}
\bar{B}_{N_0}\bm{q}^*=  B\bm{q}^*+(\bar{B}_{N_0}-B)\bm{q}^*\geq (1-\gamma)\cdot\bmu-\gamma\cdot\Rad(N_0,\eps)\geq  (1-\gamma)\cdot\bmu-\lambda_{N_0}.
\end{equation}
Therefore, as long as
\begin{equation}\label{eqn:0013105}
\lambda_{N_0}\geq\Rad(N_0,\eps),
\end{equation}
we know that $\bm{q}^*$ is a feasible solution to $\bar{V}_{N_0}$. We have that
\begin{equation}\label{eqn:013105}
\bar{\bm{r}}_{N_0}^\top\bm{q}^*\geq \hat{\bm{r}}^\top\bm{q}^*-\Rad(N_0,\eps)
\end{equation}
by noting $\|\bm{q}^*\|_1=1$ \eqref{eqn:013101} and all elements of $\bar{\bm{r}}_{N_0}-\hat{\bm{r}}$ are upper bounded by $\Rad(N_0, \eps)$. Therefore, we can obtain the bound
\begin{equation}\label{eqn:013106}
V-\Rad(N_0,\eps)\leq \bar{\bm{r}}_{N_0}^\top\bm{q}^* \leq \bar{V}_{N_0}.
\end{equation}

We then lower bound the gap $V-\bar{V}_{N_0}$. We first define
\begin{equation}\label{lp:UB2}
\begin{aligned}
\bar{V}_{N_0}(\lambda'_{N_0})=& \max&&(\bar{\bm{r}}_{N_0}-\lambda'_{N_0})^\top \bm{q} \\
&~~ \mbox{s.t.} &&\bar{C}_{N_0}\bm{q}\leq \bm{\alpha}+\lambda_{N_0}\\
& && \bar{B}_{N_0}\bm{q}\leq\bm{\mu}+\lambda_{N_0}\\
& && \bar{B}_{N_0}\bm{q}\geq\bm{\mu}-\lambda_{N_0}\\
& &&\bm{q}\geq0,
\end{aligned}
\end{equation}
for any constant $\lambda'_{N_0}$. Clearly, any optimal solution $\bar{\bm{q}}^*$ to $\bar{V}_{N_0}$ will be a feasible solution to $\bar{V}_{N_0}(\lambda'_{N_0})$. Moreover, by summing up the constraints $\bar{B}_{N_0}\bm{q}\leq\bm{\mu}+\lambda_{N_0}$ for all $s\in\mathcal{S}$, we know that
\begin{equation}\label{eqn:021703}
\|\bar{\bm{q}}^*\|_1\leq 1+\frac{|\mathcal{S}|}{1-\gamma}\cdot\lambda_{N_0}.
\end{equation}
Then, it holds that
\begin{equation}\label{eqn:021704}
\bar{V}_{N_0}\leq (\bar{\bm{r}}_{N_0}-\lambda'_{N_0})^\top \bar{\bm{q}}^* + \lambda'_{N_0}\cdot\|\bar{\bm{q}}^*\|_1\leq \bar{V}_{N_0}(\lambda'_{N_0})+\lambda'_{N_0}+\frac{|\mathcal{S}|}{1-\gamma}\cdot\lambda_{N_0}\lambda'_{N_0}.
\end{equation}
We then compare the value between $\bar{V}_{N_0}(\lambda'_{N_0})$ and $V$. The dual of $\bar{V}_{N_0}(\lambda'_{N_0})$ is given below.
\begin{equation}\label{Dual:021701}
\begin{aligned}
\text{Dual}'_{N_0}(\lambda'_{N_0})=& \min&& \bm{\alpha}^\top\bm{y}+\bm{\mu}^\top(\bm{z}_1-\bm{z}_2 )+\lambda_{N_0}\cdot( \|\bm{y}\|_1+\|\bm{z}_1\|_1+\|\bm{z}_2\|_1 ) \\
&~~ \mbox{s.t.} &&\bar{C}_{N_0}^\top \bm{y}+ \bar{B}_{N_0}^\top (\bm{z}_1-\bm{z}_2)\geq\bar{\bm{r}}_{N_0}-\lambda'_{N_0}\\
& &&\bm{y}\geq0, \bm{z}_1\geq0, \bm{z}_2\geq0.
\end{aligned}
\end{equation}
Denote by $\bm{y}^*$ and $\bm{z}^*$ one optimal solution to the dual of LP \eqref{lp:standard3}, given below.
\begin{equation}\label{Dual:standard}
\begin{aligned}
\text{Dual}=& \min&& \bm{\alpha}^\top\bm{y}+\bm{\mu}^\top\bm{z} \\
&~~ \mbox{s.t.} &&C^\top \bm{y}+ B^\top \bm{z}\geq\hat{\bm{r}}\\
& &&\bm{y}\geq0, \bm{z}\geq-\infty.
\end{aligned}
\end{equation}
We now show that $\bm{y}^*$ and $\bm{z}^*$ is also a feasible solution to $\text{Dual}'_{N_0}$, with $\lambda'_{N_0}=\frac{1}{\min_{k\in[K]}\{\alpha_k\}}\cdot\left(1+\frac{|\mathcal{S}|}{1-\gamma}\right)\cdot\Rad(N_0,\eps)$. We have the following claim regarding the upper bound on $\|\bm{y}^*\|_{\infty}$ and $\|\bm{z}^*\|_{\infty}$.
\begin{claim}\label{claim:dualBound}
There exists an optimal solution $\bm{y}^*$ and $\bm{z}^*$ to the Dual \eqref{Dual:standard} such that
\[
\|\bm{y}^*\|_1\leq \frac{1}{\min_{k\in[K]}\{\alpha_k\}}\text{~~and~~}\|\bm{z}^*\|_{\infty} \leq \frac{1}{1-\gamma}\cdot \frac{1}{\min_{k\in[K]}\{\alpha_k\}}.
\]
\end{claim}
Then, we define $\bar{\bm{y}}^*=\bm{y}^*$, $\bm{z}_1^*=\max\{0, \bm{z}^*\}$ and $\bm{z}_2^*=\max\{0, -\bm{z}^*\}$. We have
\[\begin{aligned}
\bar{C}_{N_0}^\top \bar{\bm{y}}^*+ \bar{B}_{N_0}^\top (\bm{z}_1^*-\bm{z}_2^*)&\geq C^\top \bm{y}^*+B^\top \bm{z}^*-\Rad(N_0,\eps)\cdot\left(\|\bm{y}^*\|_1+\|\bm{z}^*\|_{1} \right)\\
&\geq C^\top \bm{y}^*+B^\top \bm{z}^*-\Rad(N_0,\eps)\cdot \frac{1}{\min_{k\in[K]}\{\alpha_k\}}\cdot\left(1+\frac{|\mathcal{S}|}{1-\gamma}\right)\\
&\geq\hat{\bm{r}}-\frac{1}{\min_{k\in[K]}\{\alpha_k\}}\cdot\left(1+\frac{|\mathcal{S}|}{1-\gamma}\right)\cdot\Rad(N_0,\eps).
\end{aligned}\]
Thus, we know that $\bar{\bm{y}}^*$ and $\bm{z}_1^*, \bm{z}_2^*$ is also a feasible solution to $\text{Dual}_{N_0}(\lambda'_{N_0})$, and we have
\begin{equation}\label{eqn:021705}
\text{Dual}'_{N_0}(\lambda'_{N_0})\leq\text{Dual}+\lambda_{N_0}\cdot(\|\bm{y}^*\|_1+\|\bm{z}^*\|_1)\leq \text{Dual}+\frac{\Rad(N_0,\eps)}{\min_{k\in[K]}\{\alpha_k\}}\cdot\left(1+\frac{|\mathcal{S}|}{1-\gamma} \right).
\end{equation}
Combing \eqref{eqn:021705} with \eqref{eqn:021704} and also noting that $\bar{V}_{N_0}(\lambda'_{N_0})=\text{Dual}'_{N_0}(\lambda'_{N_0})$, we have
\[
\bar{V}_{N_0}\leq V+\frac{2\Rad(N_0,\eps)}{\min_{k\in[K]}\{\alpha_k\}}\cdot\left(1+\frac{|\mathcal{S}|}{1-\gamma} \right)+\frac{\Rad^2(N_0,\eps)}{\min_{k\in[K]}\{\alpha_k\}}\cdot\left(|\mathcal{S}|+\frac{|\mathcal{S}|^2}{1-\gamma} \right)
\]
which completes our proof.

\begin{myproof}[Proof of \Cref{claim:dualBound}]
We first bound $\|\bm{y}^*\|_{\infty}$. We utilize the approach in \cite{hiriart1996convex, nedic2009subgradient}. We define a Lagrangian function, with only $\bm{y}$ as the Lagrangian dual variable.
\begin{equation}\label{eqn:Lagrangian}
    L(\bm{y}, \bm{q}):= \bm{\alpha}^\top\bm{y}+\hat{\bm{r}}^\top\bm{q}-\bm{y}^\top C\bm{q}
\end{equation}
where the feasible set for $\bm{q}$ is $\{ \bm{q}\geq0: B\bm{q}=\bmu \}$ and the feasible set for $\bm{y}$ is $\{\bm{y}\geq0\}$. Following Lemma 3 in \cite{nedic2009subgradient}, it is without loss of generality to restrict the feasible set of $\bm{y}$ to the set $\{\bm{y}\geq0: \|\bm{y}\|_1\leq\rho\}$, where the constant $\rho$ is defined as
\begin{equation}\label{eqn:rho}
    \rho=\frac{\hat{\bm{r}}^\top\bm{q}^*-\hat{\bm{r}}^\top\bar{\bm{q}}}{\min_{k\in[K]}\left\{\alpha_k-C(k,:)\bar{\bm{q}}\right\}}
\end{equation}
where $\bar{\bm{q}}$ is the occupancy measure that satisfies Slater's condition as stated in \Cref{assump:Slater}. Note that we have
\[
\rho\leq\frac{1}{\min_{k\in[K]}\{\alpha_k\}}.
\]
Therefore, we obtain the following bound on $\bm{y}^*$:
\begin{equation}\label{eqn:021603}
    \|\bm{y}^*\|_1\leq \rho.
\end{equation}

We now proceed to bound $\bm{z}^*$. Denote by $\bm{q}^*$ the optimal solution corresponding to the optimal dual solution $(\bm{y}^*, \bm{z}^*)$, with $\bm{y}^*$ bounded as in \eqref{eqn:021603}. We also denote by $\pi^*$ the optimal policy corresponding to $\bm{q}^*$. Then, from the complementary slackness condition, as long as $\bm{q}^*(s,a)>0$ for a $(s,a)$, we must have
\begin{equation}\label{eqn:021601}
C(:,(s,a))^\top \bm{y}^* + B(:, (s,a))^\top\bm{z}^* = \hat{r}(s,a).
\end{equation}
We now multiply both sides of \eqref{eqn:021601} by $q^*(s,a)/\sum_{a'\in\mathcal{A}}q^*(s,a')$, and sum up over $a$, for each state $s$. Then we get
\begin{equation}\label{eqn:021602}
(C^{\pi^*})^\top\bm{y}^*+B^{\pi^*}\bm{z}^*=\hat{\bm{r}}^{\pi^*}.
\end{equation}
Here, $C^{\pi^*}\in\mathbb{R}^{K\times|\mathcal{S}|}$ and the element at $k$-th row and $s$-th column is $\sum_{a\in\mathcal{A}}c_k(s,a)\cdot \frac{q^*(s,a)}{\sum_{a'\in\mathcal{A}}q^*(s,a')}$. $B^{\pi^*}=I-\gamma\cdot P^{\pi^*}\in\mathbb{R}^{|\mathcal{S}|\times\mathcal{S}}$, where $P^{\pi^*}(s,s')=\sum_{a\in\mathcal{A}}\pi^*(a|s)\cdot P(s'|s,a)$ denotes the transition probability matrix under the policy $\pi^*$. Also, $\hat{\bm{r}}^{\pi^*}\in\mathbb{R}^{|\mathcal{S}|}$ with $\hat{r}^{\pi^*}(s)=\sum_{a\in\mathcal{A}}\hat{r}(s,a)\cdot \frac{q^*(s,a)}{\sum_{a'\in\mathcal{A}}q^*(s,a')}$. Then, we have
\begin{equation}\label{eqn:021604}
    \bm{z}^* = (B^{\pi^*})^{-1}\cdot\left( \hat{\bm{r}}^{\pi^*}-(C^{\pi^*})^\top\bm{y}^* \right)
\end{equation}
From \cite{jin2020efficiently}, we know that
\begin{equation}\label{eqn:021605}
\|(B^{\pi^*})^{-1}\|_{\infty}\leq \frac{1}{1-\gamma}.
\end{equation}
Also, from the bound on $\bm{y}^*$ in \eqref{eqn:021603}, we know that
\begin{equation}\label{eqn:021606}
\left\| \hat{\bm{r}}^{\pi^*}-(C^{\pi^*})^\top\bm{y}^* \right\|_{\infty}\leq \rho.
\end{equation}
Therefore, we have that
\begin{equation}\label{eqn:021607}
    \|\bm{z}^*\|_{\infty}\leq\frac{\rho}{1-\gamma},
\end{equation}
which completes our proof.
\end{myproof}

\section{Missing Proofs in \Cref{sec:algorithm}}
\subsection{Proof of \Cref{thm:BasisTrue}}\label{pf:thmBasisTrue}
We now prove \Cref{thm:BasisTrue}. Consider $\mathcal{I}$ as its final value in the iteration of \Cref{alg:IdenbasisTrue}, we know that further dropping each element $i$ from $\mathcal{I}$ will change the LP value of $V_{\mathcal{I}}$, otherwise, the element $i$ would have already been dropped in the execution of \Cref{alg:IdenbasisTrue}. 

We first show that $|\mathcal{I}|\leq|\mathcal{J}_0|$. We prove by contradiction. Suppose that $|\mathcal{I}|>|\mathcal{J}_0|$, then we can apply \Cref{lem:InfiBasis} to the LP $V_{\mathcal{I}}$. As a result, we know that there exists a subset $\mathcal{I}'\subset \mathcal{I}$ and an optimal solution $\bm{q}'$ to $V_{\mathcal{I}}$ such that $\bm{q}'_{(\mathcal{I}')^c} = 0$ and $|\mathcal{I}'|\leq\min\{|\mathcal{I}|, |\mathcal{J}_0|\}<|\mathcal{I}|$. Therefore, we know that $\bm{q}'$ is also an optimal solution to $V_{\mathcal{I}'}$, which implies that $V_{\mathcal{I}'}=V_{\mathcal{I}}=V$ and $|\mathcal{I}'|<|\mathcal{I}|$. However, this conclusion contradicts with our previous argument that we could not further one element from the set $\mathcal{I}$ without changing the LP value $V_{\mathcal{I}}$. As a result, we know that $|\mathcal{I}|\leq|\mathcal{J}_0|$.


Denoting by $\mathcal{J}$ as its final value in the execution of \Cref{alg:IdenbasisTrue}, we are now ready to show that $|\mathcal{I}|=|\mathcal{J}|$. We prove by contradiction. Since we have proved $|\mathcal{I}|\leq|\mathcal{J}_0|$, it can never happen that $|\mathcal{I}|>|\mathcal{J}|$. Now suppose $|\mathcal{I}|<|\mathcal{J}|$ and we show that we can derive contradiction. Since we know that for any optimal solution $\bm{q}_{\mathcal{I}}'$ to $V_{\mathcal{I}}$ it must hold that $\bm{q}'_{\mathcal{I}}$, from complementary slackness condition, we know that the constraints in the dual LP of $V_{\mathcal{I}}$ must be binding for any optimal dual solution. In other words, any optimal solution to the LP $\text{Dual}_{\mathcal{J}_0, \mathcal{I}}$ must be an optimal solution to the LP $\text{Dual}'_{\mathcal{J}_0, \mathcal{I}}$, where the LP $\text{Dual}'_{\mathcal{J}', \mathcal{I}}$ is defined as follows for any set $\mathcal{J}'$.
\begin{subequations}\label{dual:DiscardEqual}
\begin{align}
\text{Dual}'_{\mathcal{J}',\mathcal{I}}= \min~~&\bm{\alpha}^\top \bm{y}+\bmu^\top \bm{z} \\
 \mbox{s.t.}~~ &(C(:,\mathcal{I}))^\top\bm{y}+(B(:,\mathcal{I}))^\top \bm{z}=\hat{\bm{r}}_{\mathcal{I}}\\
 &\bm{y}_{(\mathcal{J}'_1)^c}=0, ~\bm{z}_{(\mathcal{J}'_2)^c}=0, \\
&\bm{y}\geq0, \bm{z}\geq-\infty.
\end{align}
\end{subequations}
Now we focus on the LP $\text{Dual}'_{\mathcal{J}, \mathcal{I}}$, which we know must have an optimal solution since an optimal solution to $\text{Dual}_{\mathcal{J}, \mathcal{I}}$ is an optimal solution to $\text{Dual}_{\mathcal{J}_0, \mathcal{I}}$ and thus an optimal solution to $\text{Dual}'_{\mathcal{J}_0, \mathcal{I}}$ satisfying the elements in the index set $\mathcal{J}^c$ equal $0$, which finally implies that it is an optimal solution to $\text{Dual}'_{\mathcal{J}, \mathcal{I}}$. For the LP $\text{Dual}'_{\mathcal{J}, \mathcal{I}}$ in the standard form, we can consider an optimal basis to $\text{Dual}'_{\mathcal{J}, \mathcal{I}}$, which we denote by $\mathcal{J}''\subset\mathcal{J}$. The basis set $\mathcal{J}''$ satisfies that $|\mathcal{J}''|=|\mathcal{I}|<|\mathcal{J}|$ and the matrix $\begin{bmatrix}
C(\mathcal{J}''_1, \mathcal{I})\\
B(\mathcal{J}''_2, \mathcal{I})
\end{bmatrix}$ is of full rank.  Now since $|\mathcal{J}''|<|\mathcal{J}|$, there must exists an element $j$ such that $j\in\mathcal{J}$ and $j\notin\mathcal{J}''$. We denote the set $\mathcal{J}'=\mathcal{J}\backslash\{j\}$. Then, since 
\[
\text{Dual}'_{\mathcal{J}'', \mathcal{I}}=\text{Dual}'_{\mathcal{J}, \mathcal{I}}=\text{Dual}_{\mathcal{J}, \mathcal{I}}=V
\]
we must have
\[
\text{Dual}_{\mathcal{J}', \mathcal{I}}=\text{Dual}'_{\mathcal{J}', \mathcal{I}}=\text{Dual}'_{\mathcal{J}, \mathcal{I}}=V.
\]
Also, since the matrix $\begin{bmatrix}
C(\mathcal{J}''_1, \mathcal{I})\\
B(\mathcal{J}''_2, \mathcal{I})
\end{bmatrix}$ is a submatrix of $\begin{bmatrix}
C(\mathcal{J}'_1, \mathcal{I})\\
B(\mathcal{J}'_2, \mathcal{I})
\end{bmatrix}$ and the matrix $\begin{bmatrix}
C(\mathcal{J}''_1, \mathcal{I})\\
B(\mathcal{J}''_2, \mathcal{I})
\end{bmatrix}$ is full rank, we know that the matrix $\begin{bmatrix}
C(\mathcal{J}'_1, \mathcal{I})\\
B(\mathcal{J}'_2, \mathcal{I})
\end{bmatrix}$ has a rank at least $|\mathcal{I}|$. Note that the matrix $\begin{bmatrix}
C(\mathcal{J}'_1, \mathcal{I})\\
B(\mathcal{J}'_2, \mathcal{I})
\end{bmatrix}$ has at most $|\mathcal{I}|$ columns, we know that the matrix $\begin{bmatrix}
C(\mathcal{J}'_1, \mathcal{I})\\
B(\mathcal{J}'_2, \mathcal{I})
\end{bmatrix}$ is full rank. Therefore, the element $j$ should have been dropped from the set $\mathcal{J}$ in the execution of \Cref{alg:IdenbasisTrue}. In this way, we prove that $|\mathcal{I}|=|\mathcal{J}|$. 

We finally prove that the linear equations in \eqref{eqn:Lsystem} are satisfied for $\mathcal{I}$, $\mathcal{J}$, and an optimal solution. We focus on the LP $V_{\mathcal{I}}$. Note that by the above arguments, we know that $|\mathcal{I}|=|\mathcal{J}|$ and the matrix $\begin{bmatrix}
C(\mathcal{J}_1, \mathcal{I})\\
B(\mathcal{J}_2, \mathcal{I})
\end{bmatrix}$ is full rank. We also know that $\mathcal{J}$ forms an optimal basis to the LP $\text{Dual}'_{\mathcal{J}, \mathcal{I}}$, which implies that $\mathcal{J}$ forms an optimal basis to the LP $\text{Dual}_{\mathcal{J}_0, \mathcal{I}}$. Therefore, we know that the dual variables to the LP $V_{\mathcal{I}}$ within the index set $\mathcal{J}$ are basic dual variables, which further implies that the corresponding primal constraints in $V_{\mathcal{I}}$ are binding. To be specific, denote by $\bm{q}^*_{\mathcal{I}}$ the optimal primal solution to $V_{\mathcal{I}}$ corresponding to the optimal dual basic solution given by the basis $\mathcal{J}$, then we have that the primal constraints in the index set $\mathcal{J}$ are binding for $\bm{q}^*_{\mathcal{I}}$, i.e., we have
\[
\begin{bmatrix}
C(\mathcal{J}_1, \mathcal{I})\\
B(\mathcal{J}_2, \mathcal{I})
\end{bmatrix}\cdot\bm{q}^*_{\mathcal{I}} = \begin{bmatrix}
\bm{\alpha}_{\mathcal{J}_1}\\
\bm{\mu}_{\mathcal{J}_2}
\end{bmatrix}
\]
The full-rankness of the matrix $\begin{bmatrix}
C(\mathcal{J}_1, \mathcal{I})\\
B(\mathcal{J}_2, \mathcal{I})
\end{bmatrix}$ guarantees that the above linear system is uniquely determined. Further complementing with setting $\bm{q}^*_{\mathcal{I}^c}=0$ shows that the set $\mathcal{I}$, $\mathcal{J}$ and the solution $\bm{q}^*$ satisfy the conditions in \Cref{lem:InfiBasis}. Our proof is thus completed.

\section{Missing Proofs in \Cref{sec:analysis}}
\subsection{Proof of \Cref{thm:Infibasis2}}\label{pf:Thm1}
We now condition on the event that $|\bar{r}_{N_0}(s,a)-\hat{r}(s,a)|$, $|\bar{c}_{k, N_0}(s,a)-\hat{c}_k(s,a)|$, and $|\bar{P}_{N_0}(s'|s,a)-P(s'|s,a)|$ for each $k\in[K]$, $(s,a)\in\mathcal{S}\times\mathcal{A}$ and $s'\in\mathcal{S}$ are all bounded by $\Rad(N_0, \eps)$, which is denoted as event $\mathcal{E}$. We know that this event $\mathcal{E}$ happens with probability at least $1-(K|\mathcal{S}||\mathcal{A}|-|\mathcal{S}|^2|\mathcal{A}|)\cdot\eps$.

In order to complete our proof, we show that the procedures described in \Cref{alg:Idenbasis} are in fact identical to the procedures in \Cref{alg:IdenbasisTrue} conditional on the event described above happens. Then, further combining with \Cref{thm:BasisTrue}, we complete our proof of $\mathcal{I}_{N_0}=\mathcal{I}^*$ and $\mathcal{J}_{N_0}=\mathcal{J}^*$. 

We first show that the operation of checking whether $V=V_{\mathcal{I}'}$ in \Cref{alg:IdenbasisTrue} is in fact identical to the operation of checking whether $|\bar{V}_{N_0}- \bar{V}_{\mathcal{I}', N_0}|\leq 2\Gap_1(N_0,\eps)+2\Gap_2(N_0,\eps)$ in \Cref{alg:Idenbasis}, for any set $\mathcal{I}'$. Now, for a set $\mathcal{I}'$, suppose that it holds $V_{\mathcal{I}'}=V$, then, as a result of Lemma \ref{lem:LUbound} (it is direct to see that the estimation error bound established in \Cref{lem:LUbound} also holds for $V_{\mathcal{I}'}$), we have
\begin{equation}\label{eqn:20101}
|\bar{V}_{N_0}- \bar{V}_{\mathcal{I}', N_0}|\leq |V-V_{\mathcal{I}'}|+2\Gap_1(N_0,\eps)+2\Gap_2(N_0,\eps)=2\Gap_1(N_0,\eps)+2\Gap_2(N_0,\eps).
\end{equation} 
On the other hand, for an arbitrary set $\mathcal{I}'$, suppose that it holds $|\bar{V}_{N_0}- \bar{V}_{\mathcal{I}', N_0}|\leq 2\Gap_1(N_0,\eps)+2\Gap_2(N_0,\eps)$. We prove that this would imply $V=V_{\mathcal{I}'}$ by deriving contradictions. If $V=V_{\mathcal{I}'}$ does not hold, then following the definition of $\delta_1$ in \eqref{eqn:deltaGap}, we must have 
\[
V-V_{\mathcal{I}'}\geq\delta_1.
\]
Then, from \Cref{lem:LUbound}, we have that
\begin{equation}\label{eqn:110401}
|\bar{V}_{N_0}- \bar{V}_{\mathcal{I}', N_0}|\geq |V-V_{\mathcal{I}'}|-|V-\bar{V}_{N_0}|-|\bar{V}_{\mathcal{I}', N_0}-V_{\mathcal{I}'}|\geq\delta_1- 2\Gap_1(N_0,\eps)-2\Gap_2(N_0,\eps).
\end{equation}
Therefore, as long as the sample size $N_0$ is large enough such that
\begin{equation}\label{eqn:110402}
   4\Gap_1(N_0,\eps)+4\Gap_2(N_0,\eps)<\delta_1,
\end{equation}
the inequality in \eqref{eqn:110401} would imply that $|\bar{V}_{N_0}- \bar{V}_{\mathcal{I}', N_0}|>2\Gap_1(N_0,\eps)+2\Gap_2(N_0,\eps)$, which contradicts with the initial condition that $|\bar{V}_{N_0}- \bar{V}_{\mathcal{I}', N_0}|\leq 2\Gap_1(N_0,\eps)+2\Gap_2(N_0,\eps)$. In this way, we show that the operation of checking whether $V=V_{\mathcal{I}'}$ in \Cref{alg:IdenbasisTrue} is in fact identical to the operation of checking whether $|\bar{V}_{N_0}- \bar{V}_{\mathcal{I}', N_0}|\leq 2\Gap_1(N_0,\eps)+2\Gap_2(N_0,\eps)$ in \Cref{alg:Idenbasis}, for any set $\mathcal{I}'$. As a result, \Cref{alg:Idenbasis} and \Cref{alg:IdenbasisTrue} would obtain exactly the same output set $\mathcal{I}$ for the non-zero basic variables.

We then show that the operation of checking $V=\text{Dual}_{\mathcal{J}', \mathcal{I}}$ in \Cref{alg:IdenbasisTrue} is in fact identical to the operation of checking whether $|\bar{V}_{N_0}- \text{Dual}_{\mathcal{J}', \mathcal{I}, N_0}|\leq 2\Gap_1(N_0,\eps)+2\Gap_2(N_0,\eps)$ in \Cref{alg:Idenbasis}, for any sets $\mathcal{J}'$ and $\mathcal{I}$. We now prove the following claim regarding the estimation error over the dual LP $\text{Dual}_{\mathcal{J}, \mathcal{I}}$, which is analogous to \Cref{lem:LUbound}.
\begin{claim}\label{claim:DualEstimate}
Conditional on the event $\mathcal{E}$ happens, for any sets $\mathcal{I}$ and $\mathcal{J}$, it happens that 
\[
|\text{Dual}_{\mathcal{J}, \mathcal{I}}-\text{Dual}_{\mathcal{J}, \mathcal{I}, N_0}|\leq \Gap_1(N_0,\eps)+\Gap_2(N_0,\eps).
\]
\end{claim}

The proof of \Cref{claim:DualEstimate} is relegated to the end of the proof. Now, for sets $\mathcal{I}$ and $\mathcal{J}'$, suppose that it holds $\text{Dual}_{\mathcal{J}', \mathcal{I}}=V$, then, as a result of \Cref{claim:DualEstimate}, we have
\begin{equation}\label{eqn:110403}
|\bar{V}_{N_0}- \text{Dual}_{\mathcal{J}', \mathcal{I}, N_0}|\leq |V-\text{Dual}_{\mathcal{J}', \mathcal{I}}|+2\Gap_1(N_0,\eps)+2\Gap_2(N_0,\eps)=2\Gap_1(N_0,\eps)+2\Gap_2(N_0,\eps).
\end{equation} 
On the other hand, for arbitrary sets $\mathcal{I}$ and $\mathcal{J}'$, suppose that it holds $|\bar{V}_{N_0}- \text{Dual}_{\mathcal{J}', \mathcal{I}, N_0}|\leq 2\Gap_1(N_0,\eps)+2\Gap_2(N_0,\eps)$. We prove that this would imply $V=\text{Dual}_{\mathcal{J}', \mathcal{I}}$ by deriving contradictions. If $V=\text{Dual}_{\mathcal{J}', \mathcal{I}}$ does not hold, then following the definition of $\delta_2$ in \eqref{eqn:deltaDual}, we must have 
\[
\text{Dual}_{\mathcal{J}', \mathcal{I}}-V\geq\delta_2.
\]
Then, from \Cref{claim:DualEstimate}, we have that
\begin{equation}\label{eqn:110404}
|\bar{V}_{N_0}- \text{Dual}_{\mathcal{J}', \mathcal{I}, N_0}|\geq |V-\text{Dual}_{\mathcal{J}', \mathcal{I}}|-|V-\bar{V}_{N_0}|-| \text{Dual}_{\mathcal{J}', \mathcal{I}}-\text{Dual}_{\mathcal{J}', \mathcal{I}, N_0}|\geq\delta_2- 2\Gap_1(N_0,\eps)-2\Gap_2(N_0,\eps).
\end{equation}
Therefore, as long as the sample size $N_0$ is large enough such that
\begin{equation}\label{eqn:110405}
   4\Gap_1(N_0,\eps)+4\Gap_2(N_0,\eps)<\delta_2,
\end{equation}
the inequality in \eqref{eqn:110404} would imply that $|\bar{V}_{N_0}- \text{Dual}_{\mathcal{J}', \mathcal{I}, N_0}|>2\Gap_1(N_0,\eps)+2\Gap_2(N_0,\eps)$, which contradicts with the initial condition that $|\bar{V}_{N_0}- \text{Dual}_{\mathcal{J}', \mathcal{I}, N_0}|\leq 2\Gap_1(N_0,\eps)+2\Gap_2(N_0,\eps)$. In this way, we show that the operation of checking whether $V=\text{Dual}_{\mathcal{J}', \mathcal{I}}$ in \Cref{alg:IdenbasisTrue} is in fact identical to the operation of checking whether $|\bar{V}_{N_0}- \text{Dual}_{\mathcal{J}', \mathcal{I}, N_0}|\leq 2\Gap_1(N_0,\eps)+2\Gap_2(N_0,\eps)$ in \Cref{alg:Idenbasis}, for any sets $\mathcal{I}$ and $\mathcal{J}'$.

We finally show that the operation of checking whether the smallest singular value of $\begin{bmatrix}
\bar{C}_{N_0}(\mathcal{J}'_1, \mathcal{I})\\
\bar{B}_{N_0}(\mathcal{J}'_2, \mathcal{I})
\end{bmatrix}$, denoted as $\bar{\sigma}$, satisfies that $\bar{\sigma}\geq|\mathcal{J}'||\mathcal{I}|\cdot\Rad(N_0, \eps)$ in \Cref{alg:Idenbasis} is equivalent to the operation of checking whether the matrix $\begin{bmatrix}
C(\mathcal{J}'_1, \mathcal{I})\\
B(\mathcal{J}'_2, \mathcal{I})
\end{bmatrix}$ is full rank in \Cref{alg:IdenbasisTrue}, for any sets $\mathcal{I}$ and $\mathcal{J}'$. Note that for any sets $\mathcal{I}$ and $\mathcal{J}'$, we denote by $\sigma'$ the smallest singular value of the matrix $\begin{bmatrix}
C(\mathcal{J}'_1, \mathcal{I})\\
B(\mathcal{J}'_2, \mathcal{I})
\end{bmatrix}$ and we denote by $\bar{\sigma}$ the smallest singular value of the matrix $\begin{bmatrix}
\bar{C}_{N_0}(\mathcal{J}'_1, \mathcal{I})\\
\bar{B}_{N_0}(\mathcal{J}'_2, \mathcal{I})
\end{bmatrix}$. Following Theorem 1 of \cite{stewart1998perturbation}, we know that the difference between the singular values of the matrix $\begin{bmatrix}
C(\mathcal{J}'_1, \mathcal{I})\\
B(\mathcal{J}'_2, \mathcal{I})
\end{bmatrix}$ and the corresponding singular values of the matrix $\begin{bmatrix}
\bar{C}_{N_0}(\mathcal{J}'_1, \mathcal{I})\\
\bar{B}_{N_0}(\mathcal{J}'_2, \mathcal{I})
\end{bmatrix}$ is upper bounded by the $l_2$-norm of their difference, i.e., we can show that 
\begin{equation}\label{eqn:110406}
    |\sigma'-\bar{\sigma}|\leq \left\| \begin{bmatrix}
C(\mathcal{J}'_1, \mathcal{I})\\
B(\mathcal{J}'_2, \mathcal{I})
\end{bmatrix} - \begin{bmatrix}
\bar{C}_{N_0}(\mathcal{J}'_1, \mathcal{I})\\
\bar{B}_{N_0}(\mathcal{J}'_2, \mathcal{I})
\end{bmatrix} \right\|_2 < |\mathcal{J}'||\mathcal{I}|\cdot\Rad(N_0, \eps),
\end{equation}
where the second inequality holds because the absolute value of each element of the matrix $\begin{bmatrix}
C(\mathcal{J}'_1, \mathcal{I})\\
B(\mathcal{J}'_2, \mathcal{I})
\end{bmatrix} - \begin{bmatrix}
\bar{C}_{N_0}(\mathcal{J}'_1, \mathcal{I})\\
\bar{B}_{N_0}(\mathcal{J}'_2, \mathcal{I})
\end{bmatrix}$ is upper bounded by $\Rad(N_0, \eps)$ conditional on the event $\mathcal{E}$ happens. 

Now suppose that condition $\bar{\sigma}\geq|\mathcal{J}'||\mathcal{I}|\cdot\Rad(N_0, \eps)$ holds. Then, following \eqref{eqn:110406}, we must have 
\[
\sigma'\geq \bar{\sigma} - |\sigma'-\bar{\sigma}| > |\mathcal{J}'||\mathcal{I}|\cdot\Rad(N_0, \eps)- |\mathcal{J}'||\mathcal{I}|\cdot\Rad(N_0, \eps) = 0.
\]
Therefore, we know that the matrix $\begin{bmatrix}
C(\mathcal{J}'_1, \mathcal{I})\\
B(\mathcal{J}'_2, \mathcal{I})
\end{bmatrix}$ is full rank. 

On the other hand, suppose that the matrix $\begin{bmatrix}
C(\mathcal{J}'_1, \mathcal{I})\\
B(\mathcal{J}'_2, \mathcal{I})
\end{bmatrix}$ is full rank such that $\sigma'>0$. As long as the sample size $N_0$ is large enough such that 
\begin{equation}\label{eqn:110407}
    2|\mathcal{J}'||\mathcal{I}|\cdot\Rad(N_0, \eps)\leq \sigma_0\leq\sigma', 
\end{equation}
where the second inequality follows from the definition of $\sigma_0$ in \eqref{def:sigma0}, from inequality \eqref{eqn:110406}, we know that
\[
\bar{\sigma}\geq \sigma' - |\sigma'-\bar{\sigma}| \geq 2|\mathcal{J}'||\mathcal{I}|\cdot\Rad(N_0, \eps)-|\mathcal{J}'||\mathcal{I}|\cdot\Rad(N_0, \eps) = |\mathcal{J}'||\mathcal{I}|\cdot\Rad(N_0, \eps).
\]
In this way, we show that the operation of checking whether the smallest singular value of $\begin{bmatrix}
\bar{C}_{N_0}(\mathcal{J}'_1, \mathcal{I})\\
\bar{B}_{N_0}(\mathcal{J}'_2, \mathcal{I})
\end{bmatrix}$, denoted as $\bar{\sigma}$, satisfies that $\bar{\sigma}\geq|\mathcal{J}'||\mathcal{I}|\cdot\Rad(N_0, \eps)$ in \Cref{alg:Idenbasis} is equivalent to the operation of checking whether the matrix $\begin{bmatrix}
C(\mathcal{J}'_1, \mathcal{I})\\
B(\mathcal{J}'_2, \mathcal{I})
\end{bmatrix}$ is full rank in \Cref{alg:IdenbasisTrue}, for any sets $\mathcal{I}$ and $\mathcal{J}'$.

Conditional on the event $\mathcal{E}$ happens, we know that every step in \Cref{alg:Idenbasis} is essentially identical to the step in \Cref{alg:Idenbasis}. Therefore, we know that the output sets $\mathcal{I}_{N_0}$ and $\mathcal{J}_{N_0}$ of \Cref{alg:Idenbasis} satisfy that $\mathcal{I}_{N_0}=\mathcal{I}^*$ and $\mathcal{J}_{N_0}=\mathcal{J}^*$. Further combining with \Cref{thm:BasisTrue}, we know that $\mathcal{I}_{N_0}$ and $\mathcal{J}_{N_0}$ satisfy the conditions described in \Cref{lem:InfiBasis} conditional on the event $\mathcal{E}$ happens, which happens with a probability at least $1-(K|\mathcal{S}||\mathcal{A}|-|\mathcal{S}|^2|\mathcal{A}|)\cdot\eps$. Our proof is thus completed.

\begin{myproof}[Proof of Claim \ref{claim:DualEstimate}]
Consider the dual LP of $\text{Dual}_{\mathcal{J}, \mathcal{I}}$, we have that 
\begin{equation}\label{lp:DualIJ}
\begin{aligned}
\text{Dual}_{\mathcal{J}, \mathcal{I}}=V_{\mathcal{I}, \mathcal{J}}= \max~~&\hat{\bm{r}}^\top \bm{q} && \text{Dual}_{\mathcal{J},\mathcal{I}, N_0}=\bar{V}_{\mathcal{J}, \mathcal{I}, N_0}=  &&\max&&(\bar{\bm{r}}_{N_0})^\top \bm{q} \\
~~ \mbox{s.t.} ~~&C(\mathcal{J}_1, :)\bm{q}\leq\bm{\alpha} && && ~~\mbox{s.t.} &&\bar{C}_{N_0}(\mathcal{J}_1,:)\bm{q}\leq \bm{\alpha}_{\mathcal{J}_1}+\lambda_{N_0}\\
&  B(\mathcal{J}_2,:)\bm{q}=\bm{\mu} && &&  && \|\bar{B}_{N_0}(\mathcal{J}_2, :)\bm{q}-\bm{\mu}\|_{\infty}\leq \lambda_{N_0}\\
& \bm{q}_{\mathcal{I}^c}=0 && && &&\bm{q}_{\mathcal{I}^c}=0\\
& \bm{q}\geq0, && && &&\bm{q}\geq0.
\end{aligned}
\end{equation}
In order to bound $\text{Dual}_{\mathcal{J}, \mathcal{I}}$ and $\text{Dual}_{\mathcal{J}, \mathcal{I}, N_0}$, it is sufficient to bound $V_{\mathcal{I}, \mathcal{J}}$ and $\bar{V}_{\mathcal{I}, \mathcal{J}, N_0}$.

We first upper bound the gap $V_{\mathcal{I}, \mathcal{J}}-\bar{V}_{\mathcal{I}, \mathcal{J}, N_0}$. Denote by $\bm{q}^*$ one optimal solution to $V_{\mathcal{I}, \mathcal{J}}$. Then, from the feasibility of $\bm{q}^*$, we know that
\begin{equation}\label{eqn:0131031}
\bar{C}_{N_0}(\mathcal{J}_1, :)\bm{q}^*=C(\mathcal{J}_1, :)\bm{q}^*+( \bar{C}_{N_0}(\mathcal{J}_1, :)-C(\mathcal{J}_1, :))\bm{q}^*\leq \bm{\alpha}+\Rad(N_0,\eps)\leq\bm{\alpha}+\lambda_{N_0}
\end{equation}
where the first inequality follows from the fact that an occupancy measure $\bm{q}^*$ satisfies $\|\bm{q}^*\|_1=1$ and all elements of $\bar{C}_{N_0}(\mathcal{J}_1, :)-C(\mathcal{J}_1, :)$ are upper bounded by $\Rad(N_0, \eps)$. Also, we know that
\begin{equation}\label{eqn:0131041}
\bar{B}_{N_0}(\mathcal{J}_2, :)\bm{q}^*=  B(\mathcal{J}_2, :)\bm{q}^*+(\bar{B}_{N_0}(\mathcal{J}_2, :)-B(\mathcal{J}_2, :))\bm{q}^*\leq \bmu+\gamma\cdot\Rad(N_0,\eps)\leq\bmu+\lambda_{N_0}
\end{equation}
where the first inequality follows from $\|\bm{q}^*\|_1=1$ and all elements of $\bar{B}_{N_0}(\mathcal{J}_2, :)-B(\mathcal{J}_2, :)$ are upper bounded by $\gamma\cdot\Rad(N_0, \eps)$. Similarly, we have that
\begin{equation}\label{eqn:00131041}
\bar{B}_{N_0}(\mathcal{J}_2, :)\bm{q}^*=  B(\mathcal{J}_2, :)\bm{q}^*+(\bar{B}_{N_0}(\mathcal{J}_2, :)-B(\mathcal{J}_2, :))\bm{q}^*\geq \bmu-\gamma\cdot\Rad(N_0,\eps)\geq  \bmu-\lambda_{N_0}.
\end{equation}
Therefore, as long as
\begin{equation}\label{eqn:00131051}
\lambda_{N_0}\geq\Rad(N_0,\eps),
\end{equation}
we know that $\bm{q}^*$ is a feasible solution to $\bar{V}_{\mathcal{I}, \mathcal{J}, N_0}$. We have that
\begin{equation}\label{eqn:0131051}
\bar{\bm{r}}_{N_0}^\top\bm{q}^*\geq \hat{\bm{r}}^\top\bm{q}^*-\Rad(N_0,\eps)
\end{equation}
by noting $\|\bm{q}^*\|_1=1$ and all elements of $\bar{\bm{r}}_{N_0}-\hat{\bm{r}}$ are upper bounded by $\Rad(N_0, \eps)$. Therefore, we can obtain the bound
\begin{equation}\label{eqn:0131061}
V_{\mathcal{I}, \mathcal{J}}-\Rad(N_0,\eps)\leq \bar{\bm{r}}_{N_0}^\top\bm{q}^* \leq \bar{V}_{\mathcal{I}, \mathcal{J}, N_0}.
\end{equation}

We then lower bound the gap $V_{\mathcal{I}, \mathcal{J}}-\bar{V}_{\mathcal{I}, \mathcal{J}, N_0}$. We first define
\begin{equation}\label{lp:UB21}
\begin{aligned}
\bar{V}_{\mathcal{I}, \mathcal{J}, N_0}(\lambda'_{N_0})=& \max&&(\bar{\bm{r}}_{N_0}-\lambda'_{N_0})^\top \bm{q} \\
&~~ \mbox{s.t.} &&\bar{C}_{N_0}(\mathcal{J}_1, :)\bm{q}\leq \bm{\alpha}+\lambda_{N_0}\\
& && \bar{B}_{N_0}(\mathcal{J}_2, :)\bm{q}\leq\bm{\mu}+\lambda_{N_0}\\
& && \bar{B}_{N_0}(\mathcal{J}_2, :)\bm{q}\geq\bm{\mu}-\lambda_{N_0}\\
& &&\bm{q}_{\mathcal{I}^c}=0,  \bm{q}\geq0,
\end{aligned}
\end{equation}
for any constant $\lambda'_{N_0}$. Clearly, any optimal solution $\bar{\bm{q}}^*$ to $\bar{V}_{\mathcal{I}, \mathcal{J}, N_0}$ will be a feasible solution to $\bar{V}_{\mathcal{I}, \mathcal{J}, N_0}(\lambda'_{N_0})$. Moreover, by summing up the constraints $\bar{B}_{N_0}(\mathcal{J}_2, :)\bm{q}\leq\bm{\mu}+\lambda_{N_0}$ for all $s\in\mathcal{J}_2$, we know that
\begin{equation}\label{eqn:0217031}
\|\bar{\bm{q}}^*\|_1\leq 1+\frac{|\mathcal{S}|}{1-\gamma}\cdot\lambda_{N_0}.
\end{equation}
Then, it holds that
\begin{equation}\label{eqn:0217041}
\bar{V}_{\mathcal{I}, \mathcal{J}, N_0}\leq (\bar{\bm{r}}_{N_0}-\lambda'_{N_0})^\top \bar{\bm{q}}^* + \lambda'_{N_0}\cdot\|\bar{\bm{q}}^*\|_1\leq \bar{V}_{\mathcal{I}, \mathcal{J}, N_0}(\lambda'_{N_0})+\lambda'_{N_0}+\frac{|\mathcal{S}|}{1-\gamma}\cdot\lambda_{N_0}\lambda'_{N_0}.
\end{equation}
We then compare the value between $\bar{V}_{\mathcal{I}, \mathcal{J},N_0}(\lambda'_{N_0})$ and $V_{\mathcal{I}, \mathcal{J}}$. The dual of $\bar{V}_{\mathcal{I}, \mathcal{J},N_0}(\lambda'_{N_0})$ is given below.
\begin{equation}\label{Dual:0217011}
\begin{aligned}
\text{Dual}'_{\mathcal{J}, \mathcal{I}, N_0}(\lambda'_{N_0})=& \min&& \bm{\alpha}^\top\bm{y}+\bm{\mu}^\top(\bm{z}_1-\bm{z}_2 )+\lambda_{N_0}\cdot( \|\bm{y}\|_1+\|\bm{z}_1\|_1+\|\bm{z}_2\|_1 ) \\
&~~ \mbox{s.t.} &&\bar{C}_{N_0}(:, \mathcal{I})^\top \bm{y}+ \bar{B}_{N_0}(:, \mathcal{I})^\top (\bm{z}_1-\bm{z}_2)\geq\bar{\bm{r}}_{\mathcal{I}, N_0}-\lambda'_{N_0}\\
& &&\bm{y}_{\mathcal{J}_1^c} = 0, \bm{z}_{1, \mathcal{J}_2^c} =  \bm{z}_{2, \mathcal{J}_2^c} = 0\\
& &&\bm{y}\geq0, \bm{z}_1\geq0, \bm{z}_2\geq0.
\end{aligned}
\end{equation}
Denote by $\bm{y}^*$ and $\bm{z}^*$ one optimal solution to the dual LP $\text{Dual}_{\mathcal{J}, \mathcal{I}}$.
We now show that $\bm{y}^*$ and $\bm{z}^*$ is also a feasible solution to $\text{Dual}'_{\mathcal{J}, \mathcal{I}, N_0}$, with $\lambda'_{N_0}=\frac{1}{\min_{k\in[K]}\{\alpha_k\}}\cdot\left(1+\frac{|\mathcal{S}|}{1-\gamma}\right)\cdot\Rad(N_0,\eps)$. Following the same procedure as proving \Cref{claim:dualBound}, we can show that
\[
\|\bm{y}^*\|_1\leq \frac{1}{\min_{k\in[K]}\{\alpha_k\}}\text{~~and~~}\|\bm{z}^*\|_{\infty} \leq \frac{1}{1-\gamma}\cdot \frac{1}{\min_{k\in[K]}\{\alpha_k\}}.
\]
Then, we define $\bar{\bm{y}}^*=\bm{y}^*$, $\bm{z}_1^*=\max\{0, \bm{z}^*\}$ and $\bm{z}_2^*=\max\{0, -\bm{z}^*\}$. We have
\[\begin{aligned}
\bar{C}_{N_0}(:, \mathcal{I})^\top \bar{\bm{y}}^*+ \bar{B}_{N_0}(:, \mathcal{I})^\top (\bm{z}_1^*-\bm{z}_2^*)&\geq C(:, \mathcal{I})^\top \bm{y}^*+B(:, \mathcal{I})^\top \bm{z}^*-\Rad(N_0,\eps)\cdot\left(\|\bm{y}^*\|_1+\|\bm{z}^*\|_{1} \right)\\
&\geq C(:, \mathcal{I})^\top \bm{y}^*+B(:, \mathcal{I})^\top \bm{z}^*-\Rad(N_0,\eps)\cdot \frac{1}{\min_{k\in[K]}\{\alpha_k\}}\cdot\left(1+\frac{|\mathcal{S}|}{1-\gamma}\right)\\
&\geq\hat{\bm{r}}-\frac{1}{\min_{k\in[K]}\{\alpha_k\}}\cdot\left(1+\frac{|\mathcal{S}|}{1-\gamma}\right)\cdot\Rad(N_0,\eps).
\end{aligned}\]
Thus, we know that $\bar{\bm{y}}^*$ and $\bm{z}_1^*, \bm{z}_2^*$ is also a feasible solution to $\text{Dual}_{\mathcal{J}, \mathcal{I},N_0}(\lambda'_{N_0})$, and we have
\begin{equation}\label{eqn:0217051}
\text{Dual}'_{\mathcal{J}, \mathcal{I}, N_0}(\lambda'_{N_0})\leq\text{Dual}_{\mathcal{J}, \mathcal{I},}+\lambda_{N_0}\cdot(\|\bm{y}^*\|_1+\|\bm{z}^*\|_1)\leq \text{Dual}_{\mathcal{J}, \mathcal{I},}+\frac{\Rad(N_0,\eps)}{\min_{k\in[K]}\{\alpha_k\}}\cdot\left(1+\frac{|\mathcal{S}|}{1-\gamma} \right).
\end{equation}
Combing \eqref{eqn:0217051} with \eqref{eqn:0217041} and also noting that $\bar{V}_{\mathcal{I}, \mathcal{J}, N_0}(\lambda'_{N_0})=\text{Dual}'_{\mathcal{J}, \mathcal{I}, N_0}(\lambda'_{N_0})$, we have
\[
\bar{V}_{\mathcal{I}, \mathcal{J}, N_0}\leq V_{\mathcal{I}, \mathcal{J}}+\frac{2\Rad(N_0,\eps)}{\min_{k\in[K]}\{\alpha_k\}}\cdot\left(1+\frac{|\mathcal{S}|}{1-\gamma} \right)+\frac{\Rad^2(N_0,\eps)}{\min_{k\in[K]}\{\alpha_k\}}\cdot\left(|\mathcal{S}|+\frac{|\mathcal{S}|^2}{1-\gamma} \right)
\]
which completes our proof.
\end{myproof}

\subsection{Proof of Lemma \ref{lem:RemainRe}}\label{pf:Lemma3}
Note that the distribution of $\bm{q}^n$ is independent of the distribution of $\bm{r}^n$. We know that
\[\begin{aligned}
\mathbb{E}\left[\sum_{n=1}^N (\bm{r}^n)^\top\bm{q}^n \right]=\sum_{n=1}^N (\hat{\bm{r}})^\top\mathbb{E}[\bm{q}^n]= \sum_{n=1}^N (\hat{\bm{r}}_{\mathcal{I}^*})^\top\mathbb{E}\left[\bm{q}^n_{\mathcal{I}^*}\right] 
\end{aligned}\]
Denote by $\bm{q}^*$ the optimal primal basic solution corresponding to the optimal basis $\mathcal{I}^*$ and $\mathcal{J}^*$, and denote by $\bm{y}^*, \bm{z}^*$ the optimal dual solution corresponding to $\bm{q}^*$.
From the complementary slackness condition and noting that $\bm{q}^*_{\mathcal{I}^*}>0$, we know that
\begin{equation}\label{eqn:020204}
(C(\mathcal{J}_1^*, \mathcal{I}^*))^\top\bm{y}_{\mathcal{J}_1^*}^*+(B(\mathcal{J}^*_2,\mathcal{I}^*))^\top\bm{z}_{\mathcal{J}^*_2}^*=\hat{\bm{r}}_{\mathcal{I}^*}.
\end{equation}
Also, we can define a matrix $C^n(\mathcal{J}_1^*, \mathcal{I}^*)$ such that the element of $C^n(\mathcal{J}_1^*, \mathcal{I}^*)$ at the $k\in\mathcal{J}_1^*$ row and $(s,a)\in\mathcal{I}^*$ column is $c^n_k(s,a)$. We can also define a matrix $B^n(\mathcal{J}^*_2, \mathcal{I}^*)$ such that the element of $B^n(\mathcal{J}^*_2, \mathcal{I}^*)$ at the $s'\in\mathcal{J}^*_2$ row and $(s,a)\in\mathcal{I}^*$ column is $\delta_{s',s}-\gamma\cdot\bI_{\{ s'=s^n(s,a) \}}$. It is easy to see that
\[
\mathbb{E}[C^n(\mathcal{J}_1^*, \mathcal{I}^*)]= C(\mathcal{J}_1^*, \mathcal{I}^*) \text{~and~}\mathbb{E}[B^n(\mathcal{J}^*_2, \mathcal{I}^*)]= B(\mathcal{J}^*_2, \mathcal{I}^*).
\]
Then, it holds that
\begin{equation}\label{eqn:011801}
\begin{aligned}
\sum_{n=1}^N (\hat{\bm{r}}_{\mathcal{I}^*})^\top\mathbb{E}\left[\bm{q}^n_{\mathcal{I}^*}\right]&=\sum_{n=1}^N \left( (C(\mathcal{J}_1^*, \mathcal{I}^*))^\top\bm{y}_{\mathcal{J}_1^*}^*+(B(\mathcal{J}^*_2,\mathcal{I}^*))^\top\bm{z}_{\mathcal{J}^*_2}^* \right)^\top \mathbb{E}[\bm{q}^n_{\mathcal{I}^*}]\\
&=\mathbb{E}\left[ \sum_{n=1}^N \left( (C^n(\mathcal{J}_1^*, \mathcal{I}^*))^\top\bm{y}_{\mathcal{J}_1^*}^*+(B^n(\mathcal{J}^*_2,\mathcal{I}^*))^\top\bm{z}_{\mathcal{J}^*_2}^* \right)^\top\bm{q}^n_{\mathcal{I}^*} \right]\\
&=\mathbb{E}\left[ \sum_{n=1}^N \left( (\bm{y}^*)^\top(C^n(\mathcal{J}_1^*, \mathcal{I}^*))\bm{q}^n_{\mathcal{I}^*}+(\bm{z}_{\mathcal{J}^*_2}^*)^\top(B^n(\mathcal{J}^*_2,\mathcal{I}^*))\bm{q}^n_{\mathcal{I}^*}\right) \right]
\end{aligned}
\end{equation}
Note that we have
\begin{equation}\label{eqn:011802}
\sum_{n=1}^N C^n(\mathcal{J}_1^*, \mathcal{I}^*))\bm{q}^n_{\mathcal{I}^*}=\bm{\alpha}^{1}_{\mathcal{J}_1^*}-\bm{\alpha}^N_{\mathcal{J}_1^*}
\end{equation}
and
\begin{equation}\label{eqn:011803}
\sum_{n=1}^N B^n(\mathcal{J}^*_2,\mathcal{I}^*))\bm{q}^n_{\mathcal{I}^*}=\bmu^{1}_{\mathcal{J}_2^*}-\bmu^N_{\mathcal{J}_2^*}.
\end{equation}
Plugging \eqref{eqn:011802} and \eqref{eqn:011803} back into \eqref{eqn:011801}, we get that
\[
\sum_{n=1}^N (\hat{\bm{r}}_{\mathcal{I}^*})^\top\mathbb{E}\left[\bm{q}^n_{\mathcal{I}^*}\right]=(\bm{y}_{\mathcal{J}_1^*}^*)^\top \bm{\alpha}^{1}_{\mathcal{J}_1^*} +(\bm{z}_{\mathcal{J}^*_2}^*)^\top \bmu^{1}_{\mathcal{J}^*_2}  - (\bm{y}_{\mathcal{J}_1^*}^*)^\top \mathbb{E}\left[\bm{\alpha}^N_{\mathcal{J}_1^*} \right]-(\bm{z}_{\mathcal{J}^*_2}^*)\top \mathbb{E}\left[\bmu_{\mathcal{J}^*_2}^N\right].
\]
Note that
\[
V_r(\pi^*, \mu_1)=(\bm{y}_{\mathcal{J}_1^*}^*)^\top \bm{\alpha}^1_{\mathcal{J}_1^*} +(\bm{z}_{\mathcal{J}^*_2}^*)^\top \bmu^1_{\mathcal{J}^*_2}
\]
that holds from the strong duality of $\VI$ \eqref{lp:Infinite}.
Then, we have that
\begin{equation}\label{eqn:020201}
\begin{aligned}
&N\cdot V_r(\pi^*, \mu_1)-\sum_{n=1}^N (\hat{\bm{r}})^\top\mathbb{E}[\bm{q}^n]
\leq (\bm{y}_{\mathcal{J}_1^*}^*)^\top\mathbb{E}\left[\bm{\alpha}^N_{\mathcal{J}_1^*}\right]+(\bm{z}_{\mathcal{J}_2^*}^*)^\top \mathbb{E}\left[\bmu_{\mathcal{J}^*_2}^N \right].
\end{aligned}
\end{equation}
Our proof is thus completed.

\subsection{Proof of \Cref{thm:BoundRe}}\label{pf:Thm2}
We specify the value of $\eps$ as $\eps=O(\veps^2/(K\cdot|\mathcal{S}||\mathcal{A}|))$.
We denote by $\mathcal{E}_0$ the event that \Cref{alg:Idenbasis} has successfully identified the optimal basis $\mathcal{I}^*$ and $\mathcal{J}^*$, in the execution of \Cref{alg:Twophase}. Following \Cref{thm:Infibasis2}, we know that
\[
P(\mathcal{E}_0)\geq 1-(K|\mathcal{S}||\mathcal{A}|-|\mathcal{S}|^2|\mathcal{A}|)\cdot\eps.
\]
For any $n\in[N_2]$, we denote by $\mathcal{E}_n$ the event that $|\bar{r}_{n}(s,a)-\hat{r}(s,a)|$, $|\bar{c}_{k, n}(s,a)-\hat{c}_k(s,a)|$, and $|\bar{P}_{n}(s'|s,a)-P(s'|s,a)|$ for each $k\in[K]$, $(s,a)\in\mathcal{S}\times\mathcal{A}$ and $s'\in\mathcal{S}$ are all bounded by $\Rad(n, \eps)$. From standard Hoeffding's inequality, we know that 
\[
P(\mathcal{E}_n)\geq 1-(K|\mathcal{S}||\mathcal{A}|-|\mathcal{S}|^2|\mathcal{A}|)\cdot\eps.
\]
for any $n\in[N_2]$. We now denote by the event $\mathcal{E}$ that
\[
\mathcal{E}=\mathcal{E}_0\cap\left( \cap_{n=1}^{N_2}\mathcal{E}_n \right).
\]
We know that
\[
P(\mathcal{E})\geq 1-(N_2+1)\cdot (K|\mathcal{S}||\mathcal{A}|-|\mathcal{S}|^2|\mathcal{A}|)\cdot\eps.
\]
In what follows, we condition on that the event $\mathcal{E}$ happens, which implies that $\hat{\mathcal{I}}=\mathcal{I}^*$ and $\hat{\mathcal{J}}=\mathcal{J}^*$. 

We consider the stochastic process $\tilde{\alpha}_k(n)$ and $\tilde{\mu}_s(n)$ defined in \eqref{eqn:Average}. For a fixed $\nu>0$ which we specify later, we define a set
\begin{equation}\label{eqn:defX}
\mathcal{X}=\{ \bm{\alpha}'\in\mathbb{R}^{K}: \alpha'_k\in[\alpha_k-\nu, \alpha_k+\nu], \forall k\in\mathcal{J}_1^* \},
\end{equation}
and
\begin{equation}\label{eqn:defY}
\mathcal{Y}=\{ \bm{\mu}'\in\mathbb{R}^{|\mathcal{S}|}: \mu'_s\in[\mu_s-\nu, \mu_s+\nu], \forall s\in\mathcal{J}^*_2 \}.
\end{equation}
It is easy to see that initially, $\tilde{\bm{\alpha}}(1)\in\mathcal{X}$ and $\tilde{\bmu}(1)\in\mathcal{Y}$. We show that $\tilde{\bm{\alpha}}(n)$ and $\tilde{\bmu}(n)$ behave well as long as they stay in the region $\mathcal{X}$ and $\mathcal{Y}$ for a
sufficiently long time. To this end, we define a stopping time
\begin{equation}\label{eqn:Stoptime}
\tau=\min_{n\in[N_2]}\{ \tilde{\bm{\alpha}}(n) \notin\mathcal{X}\text{~or~}\tilde{\bmu}(n)\notin\mathcal{Y} \}.
\end{equation}
Note that in \Cref{alg:Twophase}, to stop $\bm{q}^n$ from behaving ill when $n$ is small, we project it to a set that guarantees $\|\bm{q}^n\|_1\leq 2$. We now show in the following lemma that when $n$ is large enough but smaller than the stopping time $\tau$, there is no need to do projection.
\begin{lemma}\label{lem:projection}
There exist two constants $N'_3$ and $\nu_0$. When $N'_3\leq n\leq\tau$, and $\nu\leq\nu_0$, it holds that $\|\tilde{\bm{q}}_{\mathcal{I}^*}^n\|_1\leq2$, where $\tilde{\bm{q}}_{\mathcal{I}^*}^n$ denotes the solution to the linear equations \eqref{eqn:OptQ2}. Specifically, $N'_3$ is given as follows
\begin{equation}\label{eqn:N0prime}
N'_3=\frac{16m^4}{(\sigma^*)^2}\cdot\log(1/\eps),
\end{equation}
where $m=|\mathcal{I}^*|=|\mathcal{J}^*|$.
Also, $\nu_0$ is given as follows
\begin{equation}\label{eqn:nu0}
\nu_0:=\frac{\sigma^*\cdot(1-\gamma)}{4m^4}.
\end{equation}
\end{lemma}
We set $\nu$ satisfy the condition $\nu\leq\nu_0$ with $\nu_0$ satisfies the condition in Lemma \ref{lem:projection}.
We bound $\mathbb{E}[N-\tau]$ in the following lemma.
\begin{lemma}\label{lem:Stoptime}
Let the stopping time $\tau$ be defined in \eqref{eqn:Stoptime}. It holds that
\[
\mathbb{E}[N_2-\tau]\leq N'_3+
4m\cdot \exp(-\nu^2/8)
\]
where $N'_3$ is given in \eqref{eqn:N0prime}, as long as
\begin{equation}\label{eqn:022104}
N_2\geq N'_3\text{~~and~~}N_2\geq \frac{8}{\nu^2}\geq \frac{8}{\nu^2_0}=\frac{128m^8}{(\sigma^*)^2\cdot(1-\gamma)^2}.
\end{equation}
Also, for any $N'$ such that $N'_3\leq N'\leq N_2$, it holds that
\begin{equation}\label{eqn:HighProbtau}
P(\tau\leq N')\leq \frac{m\cdot\nu^2}{2}\cdot \exp\left( -\frac{\nu^2\cdot(N_2-N'+1)}{8 } \right).
\end{equation}
\end{lemma}
From the definition of the stopping time $\tau$ in \eqref{eqn:Stoptime}, we know that for each $k\in\mathcal{J}_1^*$, it holds
\[
\alpha_k^{\tau-1}\in[(N_2-\tau+1)\cdot(\alpha_k-\nu), ~~~(N_2-\tau+1)\cdot(\alpha_k+\nu)]
\]
Thus, we have that
\begin{equation}\label{eqn:011906}
|\alpha_k^{N_2}|\leq |\alpha_k^{\tau-1}|+ \sum_{t=\tau}^{N_2}\sum_{(s,a)\in\mathcal{I}^*}c_k^n(s,a)\cdot q^n(s,a)
\end{equation}
and thus
\begin{equation}\label{eqn:011907}
\left|\mathbb{E}[\alpha_k^{N_2}]\right| \leq 4\mathbb{E}[N_2-\tau]\leq 4N'_3+
16m\cdot \exp(-\nu^2/8).
\end{equation}
Following the same procedure, we can show that for each $s\in\mathcal{J}^*_2$, it holds that
\begin{equation}\label{eqn:011908}
\left|\mathbb{E}[\mu^{N_2}(s)]\right|\leq 4\mathbb{E}[N_2-\tau]\leq 4N'_3+
16m\cdot \exp(-\nu^2/8).
\end{equation}
We finally consider the other constraints $k\in(\mathcal{J}_1^{*})^c$ and $s\in(\mathcal{J}^*_2)^c$. 
Note that following the definition of $\bm{\alpha}^n$ and $\bmu^n$, we have that
\begin{equation}\label{eqn:011909}
A^*\cdot\left(\sum_{n=1}^{N_2}\mathbb{E}[\bm{q}^n_{\mathcal{I}^*}]\right)=\left[\bm{\alpha}_{\mathcal{J}_1^*}^1-\mathbb{E}\left[\bm{\alpha}^{N_2}_{\mathcal{J}_1^*} \right]; \bmu^1_{\mathcal{J}^*_2}-\mathbb{E}\left[\bmu_{\mathcal{J}^*_2}^{N_2} \right]\right].
\end{equation}
Also, from the binding of $\bm{q}^*$ regarding the optimal basis $\mathcal{I}^*$ and $\mathcal{J}^*$, we have
\begin{equation}\label{eqn:011910}
    N_2\cdot A^*\cdot\bm{q}^*_{\mathcal{I}^*}=[\bm{\alpha}^1_{\mathcal{J}_1^*}; \bmu_{\mathcal{J}^*_2}^1].
\end{equation}
Therefore, it holds that
\begin{equation}\label{eqn:011911}
\sum_{n=1}^{N_2}\mathbb{E}\left[\bm{q}^n_{\mathcal{I}^*} \right]=N_2\cdot\bm{q}^*_{\mathcal{I}^*}-(A^*)^{-1}\cdot \left[\mathbb{E}\left[\bm{\alpha}_{\mathcal{J}_1^*}^{N_2} \right]; \mathbb{E}\left[\bmu^{N_2}_{\mathcal{J}^*_2} \right]\right],
\end{equation}
and
\begin{equation}\label{eqn:022202}
\left\|\sum_{n=1}^{N_2}\mathbb{E}\left[\bm{q}^n_{(\mathcal{I}^{*})^c} \right]\right\|_1=0.
\end{equation}
Finally, for any $k\in(\mathcal{J}_1^{*})^c$, we have
\begin{equation}\label{eqn:011912}
\begin{aligned}
(\hat{\bm{c}}_k)^\top \left(\sum_{n=1}^{N_2}\mathbb{E}\left[\bm{q}^n \right]\right)&= N_2\cdot(\hat{\bm{c}}_k)^\top\bm{q}^*_{\mathcal{I}^*}-(\hat{\bm{c}}_k)^\top\cdot (A^*)^{-1}\cdot \left[\mathbb{E}\left[\bm{\alpha}_{\mathcal{J}_1^*}^{N_2}\right];\mathbb{E}\left[\bmu_{\mathcal{J}_2}^{N_2}\right]\right]\\
&=N_2\cdot(\hat{\bm{c}}_k)^\top\bm{q}^*-(\hat{\bm{c}}_k)^\top\cdot (A^*)^{-1}\cdot \left[\mathbb{E}\left[\bm{\alpha}_{\mathcal{J}_1^*}^{N_2}\right];\mathbb{E}\left[\bmu_{\mathcal{J}_2^*}^{N_2}\right]\right]
\end{aligned}
\end{equation}
From the feasibility of $\bm{q}^*$, we know that
\[
N_2\cdot\alpha_k\geq N_2\cdot(\hat{\bm{c}}_k)^\top\bm{q}^*.
\]
Therefore, for any $k\in(\mathcal{J}_1^{*})^c$, it holds that
\begin{equation}\label{eqn:20301}
\begin{aligned}
N_2\cdot\alpha_k-\sum_{n=1}^{N_2}\hat{\bm{c}}_k^\top\mathbb{E}\left[\bm{q}^n \right] &\leq (\hat{\bm{c}}_k)^\top\cdot (A^*)^{-1}\cdot \left[\mathbb{E}[\bm{\alpha}_{\mathcal{J}_1^*}^{N_2}]; \mathbb{E}[\bmu_{\mathcal{J}^*_2}^{N_2}]\right]\\
&\leq \frac{m}{\sigma^*}\cdot \left( 4N'_3+
16m\cdot \exp(-\nu^2/8) \right).
\end{aligned}
\end{equation}
In the same way, we can prove that
\begin{equation}\label{eqn:110601}
\left\|N_2\cdot\bmu-B\cdot\sum_{n=1}^{N_2}\mathbb{E}[\bm{q}^n] \right\|_{\infty}\leq \frac{m}{\sigma^*}\cdot \left( 4N'_3+
16m\cdot \exp(-\nu^2/8) \right).
\end{equation}
Moreover, the definition of $\sigma^*$ in \eqref{def:sigma} implies that following upper bound on the norm of the dual variable $\bm{y}^*$ and $\bm{z}^*$.
\begin{equation}\label{eqn:022105}
\|\bm{y}^*\|_2+\|\bm{z}^*\|_2=\|(A^{*\top})^{-1}\cdot\hat{\bm{r}}_{\mathcal{J}^*}  \|_2\leq\frac{m^2}{\sigma^*}.
\end{equation}
Therefore, we know that the regret over the reward and the regret over the constraint violation can all be bounded by using \eqref{eqn:011907}, \eqref{eqn:011908}, \eqref{eqn:20301}, and \eqref{eqn:110601}. We present the bounds as follows.
\begin{equation}\label{eqn:022106}
N_2\cdot V_r(\pi^*, \mu_1)-\sum_{n=1}^N \hat{\bm{r}}^\top\mathbb{E}[\bm{q}^n]\leq \frac{m^3}{\sigma^*}\cdot \left( 4N'_3+
16m\cdot \exp(-\nu^2/8)\right)\leq \frac{16m^4}{\sigma^*}+\frac{64m^7}{(\sigma^*)^3}\cdot\log(1/\eps)
\end{equation}
and 
\[
N_2\cdot\alpha_k-\sum_{n=1}^{N_2}\hat{\bm{c}}_k^\top\mathbb{E}\left[\bm{q}^n \right] \leq \frac{16m^4}{\sigma^*}+\frac{64m^7}{(\sigma^*)^3}\cdot\log(1/\eps)
\]
for each $k\in[K]$ and 
\[
\left\|N_2\cdot\bmu-B\cdot\sum_{n=1}^{N_2}\mathbb{E}[\bm{q}^n] \right\|_{\infty}\leq \frac{16m^4}{\sigma^*}+\frac{64m^7}{(\sigma^*)^3}\cdot\log(1/\eps)
\]
as long as the sample size $N_2$ satisfies that
\[
N_2\geq \frac{128m^8}{(\sigma^*)^2\cdot(1-\gamma)^2}\cdot\log(1/\eps).
\]
Our proof is thus completed.


\subsection{Proof of Lemma \ref{lem:projection}}
Denote by $\bm{q}^*$ the optimal solution corresponding to the optimal basis $\mathcal{I}^*$ and $\mathcal{J}^*$. Then, it holds that
\begin{equation}\label{eqn:OptQ3}
\begin{bmatrix}
C(\mathcal{J}_1^*, \mathcal{I}^*)\\
B(\mathcal{J}_2^*, \mathcal{I}^*)
\end{bmatrix}
\cdot\bm{q}^*_{\mathcal{I}^*}=\begin{bmatrix}
\bm{\alpha}_{\mathcal{J}_1^*}\\
\bmu_{\mathcal{J}_2^*}
\end{bmatrix}.
\end{equation}
We compare $\tilde{\bm{q}}^n_{\mathcal{I}^*}$ with $\bm{q}^*_{\mathcal{I}^*}$ when $n$ large enough. Note that for an $n$, $\tilde{\bm{q}}^n$ is the solution to the following linear equations
\begin{equation}\label{eqn:OptQ4}
\begin{bmatrix}
\bar{C}^n(\mathcal{J}_1^*, \mathcal{I}^*)\\
\bar{B}^n(\mathcal{J}^*_2, \mathcal{I}^*)
\end{bmatrix}
\cdot\tilde{\bm{q}}^n_{\mathcal{I}^*}=\begin{bmatrix}
\frac{\bm{\alpha}^n_{\mathcal{J}_1^*}}{N_2-n+1}\\
\frac{\bmu^n_{\mathcal{J}_2^*}}{N_2-n+1}
\end{bmatrix}.
\end{equation}
When $n\leq\tau$, we know that
\begin{equation}\label{eqn:020301}
\left| \bm{\alpha}_{\mathcal{J}_1^*}-\frac{\bm{\alpha}^n_{\mathcal{J}_1^*}}{N_2-n+1} \right|\leq\nu
\end{equation}
and
\begin{equation}\label{eqn:020302}
\left| \bmu_{\mathcal{J}^*_2} - \frac{\bmu_{\mathcal{J}^*_2}^n}{N_2-n+1}  \right| \leq \nu.
\end{equation}
Moreover, we know that the absolute value of each element of $\bar{C}^n(\mathcal{J}_1^*, \mathcal{I}^*)-C(\mathcal{J}_1^*, \mathcal{I}^*)$, and $\bar{B}^n(\mathcal{J}^*_2, \mathcal{I}^*)-B(\mathcal{J}^*_2, \mathcal{I}^*)$ is upper bounded by $\Rad(n, \eps)$. We now bound the distance between the solutions to the linear equations \eqref{eqn:OptQ3} and \eqref{eqn:OptQ4}. The perturbation of the matrix is denoted as
\[
\Delta A^*=\begin{bmatrix}
C(\mathcal{J}_1^*, \mathcal{I}^*)-\bar{C}^n(\mathcal{J}_1^*, \mathcal{I}^*)\\
B(\mathcal{J}^*_2, \mathcal{I}^*)-\bar{B}(\mathcal{J}^*_2, \mathcal{I}^*)
\end{bmatrix}.
\]
Clearly, it holds that
\begin{equation}\label{eqn:020303}
\|\Delta A^*\|_2 \leq \Rad(n, \eps)\cdot m^2,
\end{equation}
where $m=|\mathcal{I}^*|=|\mathcal{J}^*|$.
Therefore, as long as
\begin{equation}\label{eqn:020304}
\|\Delta A^*\|_2 \leq\Rad(n, \eps)\cdot m^2 \leq \frac{1}{2\|(A^*)^{-1}\|_2}\leq \frac{1}{2\sigma^*},
\end{equation}
following standard perturbation analysis of linear equations \citep{higham2002accuracy}, we have that
\begin{equation}\label{eqn:020305}
\begin{aligned}
    \frac{\|\tilde{\bm{q}}^n_{\mathcal{I}^*}-\bm{q}^*_{\mathcal{I}^*}\|_2}{\|\bm{q}^*_{\mathcal{I}^*}\|_2}
    &\leq \frac{\kappa(A^*)}{1-\kappa(A^*)\cdot\frac{\|\Delta A^*\|_2}{\|A^*\|_2}}\cdot \left( \frac{\|\Delta A^*\|_2}{\|A^*\|_2}+\frac{m^2\cdot\nu}{\|[\bm{\alpha}_{\mathcal{J}_1^*}; \bmu_{\mathcal{J}_2^*}]\|_2} \right)\\
    &\leq 2\cdot\kappa(A^*)\cdot\left( \frac{\|\Delta A^*\|_2}{\|A^*\|_2}+\frac{m^2\cdot\nu}{\|[\bm{\alpha}_{\mathcal{J}_1^*}; \bmu_{\mathcal{J}^*_2}]\|_2} \right)\\
    &\leq 2\cdot\kappa(A^*)\cdot\left( \frac{\|\Delta A^*\|_2}{\|A^*\|_2}+\frac{m^2\cdot\nu}{1-\gamma} \right),
\end{aligned}
\end{equation}
where $\kappa(A^*)=\|A^*\|_2\cdot\|(A^*)^{-1}\|_2$ denotes the conditional number of $A^*$. The last inequality follows from $\|[\bm{\alpha}_{\mathcal{J}_1^*}; \bmu_{\mathcal{J}_2^*}]\|_2\geq1-\gamma$. Further, note that $\|\bm{q}^*_{\mathcal{I}^*}\|_2\leq 1$. Therefore, in order to satisfy the condition $\|\tilde{\bm{q}}^n_{\mathcal{I}^*}\|_2\leq 2$, we only need the right hand side of \eqref{eqn:020305} to be upper bounded by 1. Clearly, as long as $n$ satisfies the condition \eqref{eqn:020304} and the following condition
\begin{equation}\label{eqn:020306}
2\cdot\kappa(A^*)\cdot \frac{\|\Delta A^*\|_2}{\|A\|_2}\leq 2\cdot\frac{\Rad(n,\eps)\cdot m^2}{\sigma^*} \leq \frac{1}{2},
\end{equation}
we only need to select a $\nu$ such that
\begin{equation}\label{eqn:020307}
2\cdot\kappa(A^*)\cdot \frac{m^2\cdot\nu}{1-\gamma} \leq\frac{1}{2}.
\end{equation}
Combining \eqref{eqn:020304} and \eqref{eqn:020306}, we know that $n$ needs to satisfy the following condition
\begin{equation}\label{eqn:022101}
n\geq N_3:= \frac{16m^4}{(\sigma^*)^2}\cdot\log(1/\eps).
\end{equation}
Also, $\nu$ is selected to satisfy the following condition
\begin{equation}\label{eqn:022102}
\nu\leq \nu_0:= \frac{\sigma^*\cdot(1-\gamma)}{4m^4}.
\end{equation}
Our proof is thus completed.

\subsection{Proof of Lemma \ref{lem:Stoptime}}
Now we fix a $k\in\mathcal{J}_1^*$. For any $N'$ such that $N'_3\leq N'\leq N_2$, it holds that
\[
\tilde{\alpha}_k(N')-\tilde{\alpha}_k(N'_3)=\sum_{n=N'_3}^{N'-1}(\tilde{\alpha}_k(n+1)-\tilde{\alpha}_k(n)).
\]
We define $\xi_k(n)=\tilde{\alpha}_k(n+1)-\tilde{\alpha}_k(n)$. Then, we have
\[
\tilde{\alpha}_k(N')-\tilde{\alpha}_k(N'_3)=\sum_{n=N'_3}^{N'-1}(\xi_k(n)-\mathbb{E}[\xi_k(n)|\mathcal{F}_n])+\sum_{n=N'_3}^{N'-1}\mathbb{E}[\xi_k(n)|\mathcal{F}_n].
\]
where $\mathcal{F}_n$ denotes the filtration of information up to step $n$.
Note that due to the update in \eqref{eqn:Aveupalpha}, we have
\[
\xi_k(n)=\frac{\tilde{\alpha}_k(n)-\sum_{(s,a)\in\mathcal{I}^*}c_k^n(s,a)\cdot q^n(s,a)}{N_2-n-1}.
\]
Then, it holds that
\begin{equation}\label{eqn:011901}
|\xi_k(n)-\mathbb{E}[\xi_k(n)|\mathcal{F}_n]|\leq \frac{2}{N_2-n+1}
\end{equation}
where the inequality follows from the fact that the value of $\tilde{\alpha}_k(n)$ is deterministic given the filtration $\mathcal{F}_n$ and $\|\bm{q}^n\|_1\leq 2$ for any $n$.
Note that
\[
\{\xi_k(n)-\mathbb{E}[\xi_k(n)|\mathcal{F}_n]\}_{\forall n=N'_3,\dots,N'}
\]
forms a martingale difference sequence. Following Hoeffding's inequality, for any $N''\leq N'$ and any $b>0$, it holds that
\[\begin{aligned}
P\left( \left| \sum_{n=N'_3}^{N''}(\xi_k(n)-\mathbb{E}[\xi_k(n)|\mathcal{F}_n]) \right|\geq b \right)&\leq 2\exp\left( -\frac{b^2}{2\cdot\sum_{n=N'_3}^{N''}1/(N_2-n+1)^2 } \right)\\
&\leq 2\exp\left( -\frac{b^2\cdot(N_2-N''+1)}{2 } \right).
\end{aligned}\]
Therefore, we have that
\begin{equation}\label{eqn:011902}
\begin{aligned}
&P\left( \left| \sum_{n=N'_3}^{N''}(\xi_k(n)-\mathbb{E}[\xi_k(n)|\mathcal{F}_n]) \right|\geq b \text{~for~some~}N'_3\leq N''\leq N' \right)\\
\leq &\sum_{N''=N'_3}^{N'} 2\exp\left( -\frac{b^2\cdot(N_2-N''+1)}{2 } \right)\leq b^2\cdot \exp\left( -\frac{b^2\cdot(N_2-N'+1)}{2 } \right)
\end{aligned}
\end{equation}
holds for any $b>0$.

We now bound the probability that $\tau>N'$ for one particular $N'$ such that $N_3'\leq N'\leq N$. Suppose that $N'\leq \tau$, then,  from Lemma \ref{lem:projection}, for each $n\leq N'$, we know that $\|\tilde{\bm{q}}^n\|_2\leq 2$ and therefore $\bm{q}^n=\tilde{\bm{q}}^n$ as the solution to \eqref{eqn:OptQ2}. We have
\[
\tilde{\alpha}_k(n)=\sum_{(s,a)\in\mathcal{I}^*}\bar{c}_{k,n}(s,a)\cdot q^n(s,a).
\]
It holds that
\begin{equation}\label{eqn:011903}
\left|\mathbb{E}[\xi_k(n)|\mathcal{F}_n]\right|\leq\frac{1}{N_2-n+1}\cdot \sum_{(s,a)\in\mathcal{I}^*}q^n(s,a)\cdot \left|\mathbb{E}[\bar{c}_{k,n}(s,a)]-\hat{c}_{k}^n(s,a)\right|\leq\frac{2\Rad(n,\eps)}{N_2-n+1}.
\end{equation}
Then, we know that
\begin{equation}\label{eqn:011904}
\begin{aligned}
\sum_{n=N_3'}^{N'-1}\left|\mathbb{E}[\xi_k(n)|\mathcal{F}_n]\right|&\leq\sqrt{\frac{\log(2/\eps)}{2}}\cdot\sum_{n=N_3'}^{N'-1}\frac{1}{\sqrt{n}\cdot(N_2-n)}\leq \sqrt{\frac{\log(2/\eps)}{2}}\cdot \sqrt{N'-1}\cdot \sum_{n=N_3'}^{N'-1}\frac{1}{n\cdot(N_2-n)}\\
&=\sqrt{\frac{\log(2/\eps)}{2}}\cdot \frac{\sqrt{N'-1}}{N_2}\cdot \sum_{n=N_3'}^{N'-1}\left( \frac{1}{n}+\frac{1}{N_2-n} \right)\\
&\leq \sqrt{2\log(2/\eps)}\cdot \frac{\sqrt{N'-1}}{N_2}\cdot \log(N_2)
\leq \frac{\sqrt{2\log(2/\eps)}}{\sqrt{N_2}}\cdot\log(N_2)
\leq\frac{\nu}{2}
\end{aligned}
\end{equation}
for a $N_2$ large enough such that
\begin{equation}\label{eqn:022103}
N_2\geq \frac{8}{\nu^2}\geq \frac{8}{\nu^2_0}=\frac{128m^8}{(\sigma^*)^2\cdot(1-\gamma)^2}
\end{equation}
Combining \eqref{eqn:011904} and \eqref{eqn:011902} with $b=\nu/2$, and apply a union bound over all $k\in\mathcal{J}_1^*$ and $s\in\mathcal{J}^*_2$, we know that
\begin{equation}\label{eqn:011905}
P(\tau\leq N')\leq \frac{m\cdot\nu^2}{2}\cdot \exp\left( -\frac{\nu^2\cdot(N_2-N'+1)}{8 } \right).
\end{equation}
Therefore, we know that
\[
\mathbb{E}[N_2-\tau]=\sum_{N'=1}^{N_2} P(\tau \leq N')\leq N'_3+\sum_{N'=N_3'}^{N_2} P(\tau \leq N')\leq N_3'+
4m\cdot \exp(-\nu^2/8)
\]
which completes our proof.

\subsection{Proof of \Cref{thm:sample}}\label{pf:Thm3}
Note that in the proof of \Cref{thm:BoundRe} (in \eqref{eqn:011906}), we have shown the following bounds.
\begin{equation}\label{eqn:1011907}
\left|\mathbb{E}[\alpha_k^{N_2}]\right| \leq 
16m + \frac{64m^4}{(\sigma^*)^2}\cdot\log(1/\eps)
\end{equation}
holds for the each $k\in\mathcal{J}_1^*$.
Also in \eqref{eqn:011907}, for each $s\in\mathcal{J}^*_2$, we have shown that
\begin{equation}\label{eqn:1011908}
\left|\mathbb{E}[\mu^{N_2}_s]\right|\leq 16m + \frac{64m^4}{(\sigma^*)^2}\cdot\log(1/\eps).
\end{equation}

The caveat of directly transferring the bound of \eqref{eqn:1011907} and \eqref{eqn:1011908} into the regret bounds \eqref{eqn:regret} of the policy $\bar{\pi}$ is that, the vector $\bar{\bm{q}}^{N_2}$ does not directly characterize an occupancy measure. This point can be seen by noting that there is a gap between $B\bar{\bm{q}}^{N_2}$ and $\bmu$, though bounded by $O(\log(N_2)/N_2)$ by setting $\eps=O(1/N_2^2)$. However, we can
show that the gap between $\bar{\bm{q}}^{N_2}$ and $\bm{q}^*$ is upper bounded by $O(\log(N_2)/N_2)$, which implies a bound over the gap between the policy $\bar{\pi}^{N_2}$ and the optimal policy $\pi^*$ that corresponds to the occupancy measure $\bm{q}^*$. This bound over the gap between the policy distributions can be then transferred into the bound over the gap between the state-value functions under the policy $\bar{\pi}$ and $\pi^*$. The regret bounds can be obtained then.

We first bound the gap between $\bar{\bm{q}}^{N_2}$ and $\bm{q}^*$. Note that as long as $N_1$ satisfies the condition \eqref{eqn:N0condition}, from \Cref{thm:Infibasis2}, we know that we have $\hat{\mathcal{I}}=\mathcal{I}^*$ and $\hat{\mathcal{J}}=\mathcal{J}^*$ happens with a high probability. We now condition on this event happens. Then, we know that
\begin{equation}\label{eqn:020401}
\bar{\bm{q}}^{N_2}_{(\mathcal{I}^{*})^c}=\bm{q}^*_{(\mathcal{I}^{*})^c}.
\end{equation}
Also, note that following the definition of $\bm{\alpha}^n$ and $\bmu^n$, we have that
\begin{equation}\label{eqn:1011909}
A^*\cdot\left(\sum_{n=1}^{N_2}\mathbb{E}[\bm{q}^n_{\mathcal{I}^*}]\right)=\begin{bmatrix}
\bm{\alpha}_{\mathcal{J}_1^*}^1-\mathbb{E}\left[
\bm{\alpha}^{N_2}_{\mathcal{J}_1^*} \right]\\
\bmu^1_{\mathcal{J}^*_2}-\mathbb{E}\left[\bmu_{\mathcal{J}^*_2}^{N_2} \right]\end{bmatrix}.
\end{equation}
Also, from the binding of $\bm{q}^*$ regarding the optimal basis $\mathcal{I}^*$ and $\mathcal{J}^*$, we have
\begin{equation}\label{eqn:1011910}
    N_2\cdot A^*\cdot\bm{q}^*_{\mathcal{I}^*}=[\bm{\alpha}^1_{\mathcal{J}_1^*}; \bmu_{\mathcal{J}^*_2}^1].
\end{equation}
Then, from \eqref{eqn:1011909} and \eqref{eqn:1011910}, we know that
\begin{equation}
\begin{aligned}
\left\| \mathbb{E}\left[\bar{\bm{q}}^{N_2}_{\mathcal{I}^*}\right]-\bm{q}^*_{\mathcal{I}^*} \right\|_{\infty} &= \left\|\bm{q}^*_{\mathcal{I}^*}-\frac{1}{N_2}\cdot\sum_{n=1}^{N_2}\mathbb{E}\left[\bm{q}^n_{\mathcal{I}^*}\right]\right\|_{\infty}= \frac{\left\|(A^*)^{-1}\cdot \left[\mathbb{E}\left[\bm{\alpha}^{N_2}_{\mathcal{J}^*} \right]; \mathbb{E}\left[\bmu^{N_2}\right]\right] \right\|_{\infty}}{N_2}\leq \frac{\left\|\left[\mathbb{E}\left[\bm{\alpha}^{N_2}_{\mathcal{J}^*}\right]; \mathbb{E}\left[\bmu^{N_2}\right]\right]\right\|_{\infty}}{\sigma^*\cdot N_2}\\
&\leq\frac{1}{\sigma^*\cdot N_2}\cdot \left( 16m + \frac{64m^4}{(\sigma^*)^2}\cdot\log(1/\eps) \right).
\end{aligned}
\end{equation}
From Markov's inequality, for each $i\in\mathcal{I}^*$ and any $g>0$, we know that
\begin{equation}\label{eqn:022107}
P\left( |\bar{\bm{q}}^{N_2}_{i}-\bm{q}^*_i|>g \right) \leq\frac{1}{g\cdot\sigma^*\cdot N_2}\cdot \left( 16m + \frac{64m^4}{(\sigma^*)^2}\cdot\log(1/\eps) \right).
\end{equation}
We denote by
\begin{equation}\label{eqn:Sigma3}
    \xi=\min_{(s,a)\in\mathcal{I}^*} \left\{ q^*(s,a) \right\}>0.
\end{equation}
The policy $\bar{\pi}$ is essentially random by noting that $\bm{q}^{N_2}$ is a random variable, where the randomness comes from the randomness of the filtration $\mathcal{H}_{N_2}$. For each $s\in\mathcal{S}$ and $a\in\mathcal{A}$, we denote by $\bar{\pi}(a|s)$ the (ex-ante) probability that the random policy $\bar{\pi}$ takes the action $a$ given the state $s$. Then, for any $0<g\leq\xi/2$, we note that
\begin{equation}\label{eqn:022108}
|\bm{q}^{N_2}_i-\bm{q}^*_i|\leq g \text{~for~each~}i=(s,a)\in\mathcal{I}^*\text{~implies~that~}\left|\frac{\bm{q}^{N_2}(s,a)}{\sum_{a'\in\mathcal{A}}\bm{q}^{N_2}(s,a')}-\frac{\bm{q}^*(s,a)}{\sum_{a'\in\mathcal{A}}\bm{q}^*(s,a')} \right|\leq\frac{2g}{\xi},
\end{equation}
for each $i=(s,a)\in\mathcal{I}^*$. For any $0<g\leq\xi/2$, note that
\begin{equation}\label{eqn:022109}
P\left( |\bm{q}^{N_2}_i-\bm{q}^*_i|\leq g \text{~for~each~}i=(s,a)\in\mathcal{I}^* \right)\geq 1-\frac{m}{g\cdot\sigma^*\cdot N_2}\cdot \left( 16m + \frac{64m^4}{(\sigma^*)^2}\cdot\log(1/\eps) \right),
\end{equation}
where the inequality follows from the bound \eqref{eqn:022107} and the union bound over $i\in\mathcal{I}^*$. Therefore, for any $0<g\leq\xi/2$ and any $(s,a)$, we know that
\begin{equation}\label{eqn:022110}
P\left(\left|\bar{\pi}(a|s)-\pi^*(a|s)\right|\leq \frac{2g}{\xi}\right)\geq 1-\frac{m}{g\cdot\sigma^*\cdot N_2}\cdot \left( 16m + \frac{64m^4}{(\sigma^*)^2}\cdot\log(1/\eps) \right).
\end{equation}
From the above inequality, for any $(s,a)$, we have that
\begin{equation}\label{eqn:022111}
\begin{aligned}
\left| \mathbb{E}\left[\bar{\pi}(a|s)\right]-\pi^*(a|s) \right|&\leq \mathbb{E}\left[ \left|\bar{\pi}(a|s)-\pi^*(a|s)\right| \right]\leq\frac{2}{N_2\xi}+\frac{2}{\xi}\cdot\int_{g=\frac{1}{N_2}}^{\xi/2}P\left(\left|\bar{\pi}(a|s)-\pi^*(a|s)\right|\geq \frac{2g}{\xi}\right)dg\\
&\leq \frac{2}{N_2\xi}+\frac{2m}{\xi\cdot\sigma^*\cdot N_2}\cdot \left( 16m + \frac{64m^4}{(\sigma^*)^2}\cdot\log(1/\eps) \right)\cdot \int_{g=\frac{1}{N_2}}^{\xi/2}\frac{dg}{g}\\
&=\frac{2}{N_2\xi}+\frac{2m}{\xi\cdot\sigma^*\cdot N_2}\cdot \left( 16m + \frac{64m^4}{(\sigma^*)^2}\cdot\log(1/\eps) \right)\cdot\left( \log(N_2)-\log(\xi/2) \right).
\end{aligned}
\end{equation}
We finally transfer the bound \eqref{eqn:022111} into the regret bounds of policy $\bar{\pi}$. We use the state-value functions $V_r(\pi, s)$, defined for any initial state $s$ and any policy $\pi$ as follows
\begin{equation}\label{def:stateValue}
V_r(\pi, s)=\mathbb{E}\left[ \sum_{t=0}^{\infty}\gamma^t\cdot r(s_t, a_t)\mid s \right],
\end{equation}
where $(s_t, a_t)$ is generated according to the policy $\pi$ and the transition kernel $P$ with the initial state $s$. Note that the value of $V_r(\pi, s)$ for any $s\in\mathcal{S}$ can be obtained from solving Bellman's equation under policy $\pi$
\begin{equation}\label{eqn:Bellman}
V_r(\pi, s)=\mathbb{E}_{a\sim \pi(\cdot|s)}\left[ \hat{r}(s,a)+\gamma\cdot \mathbb{E}_{s'\sim P(\cdot|s,a)}[V_r(\pi, s')] \right].
\end{equation}
We define a matrix $B^{\pi}\in\mathbb{R}^{|\mathcal{S}|\times|\mathcal{S}|}$ such that the $s$-th row $s'$-th column element is
\begin{equation}\label{def:Api}
B^{\pi}(s,s')=\delta_{s,s'}-\gamma\cdot \sum_{a\in\mathcal{A}}\pi(a|s)\cdot P(s'|s,a).
\end{equation}
Then, the matrix $B^{\pi}$ represents the state transition probability matrix under the policy $\pi$.
Denote by
\[
\bm{V}_r(\pi)=(V_r(\pi, s))_{\forall s\in\mathcal{S}}
\]
and
\[
\hat{\bm{r}}(\pi)=(\sum_{a\in\mathcal{A}}\pi(a|s)\cdot\hat{r}(s,a) )_{\forall s\in\mathcal{S}}.
\]
We have that the state values $\bm{V}_r(\pi)$ is the solution to the linear equation
\begin{equation}\label{eqn:020405}
    B^{\pi}\bm{V}_r(\pi)=\hat{\bm{r}}(\pi)
\end{equation}
To bound the regret, we bound the solution to the linear equation \eqref{eqn:020405} with $\pi$ being $\bar{\pi}$ and $\pi^*$ separately. The perturbation of the right hand of the equation \eqref{eqn:020405} is
\[
\Delta \hat{\bm{r}}=\hat{\bm{r}}(\bar{\pi})-\hat{\bm{r}}(\pi^*).
\]
Clearly, we have that
\begin{equation}\label{eqn:020408}
\|\Delta \hat{\bm{r}} \|_{\infty} \leq \frac{2}{N_2\xi}+\frac{2m}{\xi\cdot\sigma^*\cdot N_2}\cdot \left( 16m + \frac{64m^4}{(\sigma^*)^2}\cdot\log(1/\eps) \right)\cdot\left( \log(N_2)-\log(\xi/2) \right).
\end{equation}
The perturbation of the matrix is denoted as
\[
\Delta B=B^{\bar{\pi}}-B^{\pi^*}.
\]
Clearly, it holds that
\begin{equation}\label{eqn:020406}
\|\Delta B\|_{\infty} \leq\frac{2\gamma}{N_2\xi}+\frac{2\gamma m}{\xi\cdot\sigma^*\cdot N_2}\cdot \left( 16m + \frac{64m^4}{(\sigma^*)^2}\cdot\log(1/\eps) \right)\cdot\left( \log(N_2)-\log(\xi/2) \right).
\end{equation}
Therefore, as long as
\begin{equation}\label{eqn:020407}
 \frac{2}{N_2\xi}+\frac{2m}{\xi\cdot\sigma^*\cdot N_2}\cdot \left( 16m + \frac{64m^4}{(\sigma^*)^2}\cdot\log(1/\eps) \right)\cdot\left( \log(N_2)-\log(\xi/2) \right) \leq 1/\|(B^{\pi^*})^{-1}\|_{\infty}=1/\sigma',
\end{equation}
following standard perturbation analysis of linear equations \citep{higham2002accuracy}, we have that
\begin{equation}\label{eqn:020409}
    \frac{\|\bm{V}_r(\bar{\pi})-\bm{V}_r(\pi^*)\|_{\infty}}{\|\bm{V}_r(\pi^*)\|_{\infty}} \leq \kappa(B^{\pi^*})\cdot\left( \frac{\|\Delta B\|_{\infty}}{\|B^{\pi^*}\|_{\infty}}+\frac{\|\Delta \hat{\bm{r}}\|_{\infty}}{\|\hat{\bm{r}}(\pi^*)\|_{\infty}} \right),
\end{equation}
where $\kappa(B^{\pi^*})=\|B^{\pi^*}\|_{\infty}\cdot\|(B^{\pi^*})^{-1}\|_{\infty}$ denotes the conditional number of $B^{\pi^*}$. Note that we have the regret
\begin{equation}\label{eqn:020410}
\begin{aligned}
\text{Regret}_r(\bar{\pi}, N_2)&= \bmu^\top (\bm{V}_r(\bar{\pi})-\bm{V}_r(\pi^*) )\leq (1-\gamma)\|\bm{V}_r(\bar{\pi})-\bm{V}_r(\pi^*)\|_{\infty} \\
&\leq (1-\gamma)\cdot\kappa(B^{\pi^*})\cdot \|\bm{V}_r(\pi^*)\|_{\infty}\cdot\left( \frac{\|\Delta B\|_{\infty}}{\|B^{\pi^*}\|_{\infty}}+\frac{\|\Delta \hat{\bm{r}}\|_{\infty}}{\|\hat{\bm{r}}(\pi^*)\|_{\infty}} \right).
\end{aligned}
\end{equation}
It is clear to see that
\begin{equation}\label{eqn:022115}
\|\bm{V}_r(\pi^*)\|_{\infty}\leq\frac{1}{1-\gamma}
\end{equation}
and
\begin{equation}\label{eqn:022116}
\|\bm{V}_r(\pi^*)\|_{\infty}\leq\frac{\|\hat{\bm{r}}(\pi^*)\|_{\infty}}{1-\gamma}.
\end{equation}
Following \cite{jin2020efficiently}, we have the following bound.
\begin{equation}\label{eqn:022003}
\sigma'= \|(B^{\pi^*})^{-1}\|_{\infty}\leq\frac{1}{1-\gamma}.
\end{equation}
Also, from the definition, we have that
\begin{equation}\label{eqn:022004}
\|B^{\pi^*}\|_{\infty} = 1-\gamma.
\end{equation}
Plugging the bound \eqref{eqn:020408}, \eqref{eqn:020406}, \eqref{eqn:022115}, and \eqref{eqn:022116}, into the inequality \eqref{eqn:020410}, we have that 
\begin{equation}\label{eqn:020411}
\text{Regret}_r(\bar{\pi}, N_2) \leq \frac{2}{N_2\xi(1-\gamma)}+\frac{2m}{\xi\sigma^* N_2(1-\gamma)}\cdot \left( 16m + \frac{64m^4}{(\sigma^*)^2}\cdot\log(1/\eps) \right)\cdot\left( \log(N_2)-\log(\xi/2) \right)
\end{equation}
In a same way, for any $k\in[K]$, we obtain that
\begin{equation}\label{eqn:020412}
\text{Regret}_k(\bar{\pi}, N_2) \leq \frac{2}{N_2\xi(1-\gamma)}+\frac{2m}{\xi\sigma^* N_2(1-\gamma)}\cdot \left( 16m + \frac{64m^4}{(\sigma^*)^2}\cdot\log(1/\eps) \right)\cdot\left( \log(N_2)-\log(\xi/2) \right)
\end{equation}
To show the sample complexity bound, we let the accuracy level $\veps$ satisfies the condition that
\[
\veps= \frac{2}{N_2\xi(1-\gamma)}+\frac{2m}{\xi\sigma^* N_2(1-\gamma)}\cdot \left( 16m + \frac{64m^4}{(\sigma^*)^2}\cdot\log(1/\eps) \right)\cdot\left( \log(N_2)-\log(\xi/2) \right)
\]
with $\eps=O(1/N_2^2)$. Then, we have that
\[
N_2= O\left( \frac{m^5\cdot\log(1/\xi)}{\xi(\sigma^*)^3(1-\gamma)}\cdot\frac{\log^2(1/\veps)}{\veps}\right)
\]
Moreover, in order for $N_1$ to satisfy the condition in \eqref{eqn:N0condition}, we have that
\[
N_1=O\left( \frac{\log(1/\veps)}{\min\{\delta_1^2, \delta_2^2, \sigma_0^2\}}  \right).
\]
We finally note that in each of $N_2$ rounds of the resolving procedure in \Cref{alg:Twophase}, we consume a number of $m^2$ samples. Also, to form the sample set $\mathcal{F}_{N_1}$ to serve as input to \Cref{alg:Idenbasis} in step 2-3 of \Cref{alg:Twophase}, we need a number of $N_1\cdot|\mathcal{S}||\mathcal{A}|$ samples. Therefore, the total number of samples can be bounded as
\[
N=O\left( \frac{|\mathcal{S}||\mathcal{A}|\cdot\log(1/\veps)}{\min\{\delta_1^2, \delta_2^2, \sigma_0^2\}}+\frac{m^7\cdot\log(1/\xi)}{\xi(\sigma^*)^3(1-\gamma)}\cdot\frac{\log^2(1/\veps)}{\veps} \right).
\]
Our proof is thus completed.

\section{Missing Proofs in \Cref{sec:extensions}}
\subsection{Proof of Lemma \ref{lem:FiLUbound}}\label{pf:FiLemma2}
We now condition on the event that $|\bar{r}_{h,N_0}(s,a)-\hat{r}(s,a)|$, $|\bar{c}_{h,k, N_0}(s,a)-\hat{c}_k(s,a)|$, and $|\bar{P}_{N_0}(s'|s,a)-P(s'|s,a)|$ for each $h\in[H]$, each $k\in[K]$, $(s,a)\in\mathcal{S}\times\mathcal{A}$ and $s'\in\mathcal{S}$ are all bounded by $\Rad(N_0, \eps)$. From the union bound, we know that this event happens with probability at least $1-H\cdot(K|\mathcal{S}||\mathcal{A}|-|\mathcal{S}|^2|\mathcal{A}|)\cdot\eps$.

Note that for LP \eqref{lp:Finite}, by summing up the constraint \eqref{eqn:Ficonstraint2} for all $s\in\mathcal{S}$, we obtain that any feasible solution $\bm{q}$ for LP \eqref{lp:Finite} would satisfy
\begin{equation}\label{eqn:1013101}
    \|\bm{q}_h\|_1=1.
\end{equation}
for each $h\in[H]$.

We first upper bound the gap $V-\bar{V}_{N_0}$ under the finite horizon episodic setting. Denote by $\bm{q}^*$ one optimal solution to $V$ \eqref{lp:Finite}. Then, from the feasibility of $\bm{q}^*$, we know that
\begin{equation}\label{eqn:1013103}
\bar{C}_{N_0}\bm{q}^*=C\bm{q}^*+( \bar{C}_{N_0}-C)\bm{q}^*\leq \bm{\alpha}+H\cdot\Rad(N_0,\eps)\leq\bm{\alpha}+\lambda_{N_0}
\end{equation}
where the first inequality follows from $\|\bm{q}_h^*\|_1=1$ in \eqref{eqn:1013101} and all elements of $\bar{C}_{N_0}-C$ are upper bounded by $\Rad(N_0, \eps)$. Also, we know that
\begin{equation}\label{eqn:013104}
\bar{B}_{N_0}\bm{q}^*=  B\bm{q}^*+(\bar{B}_{N_0}-B)\bm{q}^*\leq \bmu+H\cdot\Rad(N_0,\eps)\leq \bmu+\lambda_{N_0}
\end{equation}
where the first inequality follows from $\|\bm{q}_h^*\|_1=1$ in \eqref{eqn:1013101} and all elements of $\bar{B}_{N_0}-B$ are upper bounded by $\Rad(N_0, \eps)$. Similarly, we have that
\begin{equation}\label{eqn:10013104}
\bar{B}_{N_0}\bm{q}^*=  B\bm{q}^*+(\bar{B}_{N_0}-B)\bm{q}^*\geq \bmu-H\cdot\Rad(N_0,\eps)\geq  \bmu-\lambda_{N_0}.
\end{equation}
Therefore, as long as
\begin{equation}\label{eqn:0013105}
\lambda_{N_0}\geq H\cdot\Rad(N_0,\eps),
\end{equation}
we know that $\bm{q}^*$ is a feasible solution to $\bar{V}_{N_0}$. We have that
\begin{equation}\label{eqn:013105}
\bar{\bm{r}}_{N_0}^\top\bm{q}^*\geq \hat{\bm{r}}^\top\bm{q}^*-H\cdot\Rad(N_0,\eps)
\end{equation}
by noting $\|\bm{q}_h^*\|_1=1$ \eqref{eqn:1013101} and all elements of $\bar{\bm{r}}_{N_0}-\hat{\bm{r}}$ are upper bounded by $\Rad(N_0, \eps)$. Therefore, we can obtain the bound
\begin{equation}\label{eqn:1013106}
V-H\cdot\Rad(N_0,\eps)\leq \bar{\bm{r}}_{N_0}^\top\bm{q}^* \leq \bar{V}_{N_0}.
\end{equation}

We then lower bound the gap $V-\bar{V}_{N_0}$. We first define
\begin{equation}\label{lp:1UB2}
\begin{aligned}
\bar{V}_{N_0}(\lambda'_{N_0})=& \max&&(\bar{\bm{r}}_{N_0}-\lambda'_{N_0})^\top \bm{q} \\
&~~ \mbox{s.t.} &&\bar{C}_{N_0}\bm{q}\leq \bm{\alpha}+\lambda_{N_0}\\
& && \bar{B}_{N_0}\bm{q}\leq\bm{\mu}+\lambda_{N_0}\\
& && \bar{B}_{N_0}\bm{q}\geq\bm{\mu}-\lambda_{N_0}\\
& &&\bm{q}\geq0,
\end{aligned}
\end{equation}
for any constant $\lambda'_{N_0}$. Clearly, any optimal solution $\bar{\bm{q}}^*$ to $\bar{V}_{N_0}$ will be a feasible solution to $\bar{V}_{N_0}(\lambda'_{N_0})$. Moreover, by summing up the constraints $\bar{B}_{N_0}\bm{q}\leq\bm{\mu}+\lambda_{N_0}$ for all $s\in\mathcal{S}$, we know that
\begin{equation}\label{eqn:1021703}
\|\bar{\bm{q}}_h^*\|_1\leq 1+|\mathcal{S}|\cdot\lambda_{N_0}.
\end{equation}
Then, it holds that
\begin{equation}\label{eqn:1021704}
\bar{V}_{N_0}\leq (\bar{\bm{r}}_{N_0}-\lambda'_{N_0})^\top \bar{\bm{q}}^* + \lambda'_{N_0}\cdot\|\bar{\bm{q}}^*\|_1\leq \bar{V}_{N_0}(\lambda'_{N_0})+\lambda'_{N_0}+|\mathcal{S}|\cdot\lambda_{N_0}\lambda'_{N_0}.
\end{equation}
We then compare the value between $\bar{V}_{N_0}(\lambda'_{N_0})$ and $V$. The dual of $\bar{V}_{N_0}(\lambda'_{N_0})$ is given below.
\begin{equation}\label{Dual:1021701}
\begin{aligned}
\text{Dual}'_{N_0}(\lambda'_{N_0})=& \min&& \bm{\alpha}^\top\bm{y}+\bm{\mu}^\top(\bm{z}_1-\bm{z}_2 )+\lambda_{N_0}\cdot( \|\bm{y}\|_1+\|\bm{z}_1\|_1+\|\bm{z}_2\|_1 ) \\
&~~ \mbox{s.t.} &&\bar{C}_{N_0}^\top \bm{y}+ \bar{B}_{N_0}^\top (\bm{z}_1-\bm{z}_2)\geq\bar{\bm{r}}_{N_0}-\lambda'_{N_0}\\
& &&\bm{y}\geq0, \bm{z}_1\geq0, \bm{z}_2\geq0.
\end{aligned}
\end{equation}
Denote by $\bm{y}^*$ and $\bm{z}^*$ one optimal solution to the dual of LP \eqref{lp:standard3}, given below.
\begin{equation}\label{Dual:1standard}
\begin{aligned}
\text{Dual}=& \min&& \bm{\alpha}^\top\bm{y}+\bm{\mu}^\top\bm{z} \\
&~~ \mbox{s.t.} &&C^\top \bm{y}+ B^\top \bm{z}\geq\hat{\bm{r}}\\
& &&\bm{y}\geq0, \bm{z}\geq-\infty.
\end{aligned}
\end{equation}
We now show that $\bm{y}^*$ and $\bm{z}^*$ is also a feasible solution to $\text{Dual}'_{N_0}$, with $\lambda'_{N_0}=\Rad(N_0,\eps)\cdot \frac{K+|\mathcal{S}|\cdot H}{\sigma^*}$. We have the following claim regarding the upper bound on $\|\bm{y}^*\|_{\infty}$ and $\|\bm{z}^*\|_{\infty}$.
\begin{claim}\label{claim:1dualBound}
There exists an optimal solution $\bm{y}^*$ and $\bm{z}^*$ to the Dual \eqref{Dual:standard} such that
\[
\|(\bm{y}^*, \bm{z}^*)\|_1 \leq \frac{|\mathcal{S}|\cdot H+K}{\sigma^*}
\]
\end{claim}
Then, we define $\bar{\bm{y}}^*=\bm{y}^*$, $\bm{z}_1^*=\max\{0, \bm{z}^*\}$ and $\bm{z}_2^*=\max\{0, -\bm{z}^*\}$. We have
\[\begin{aligned}
\bar{C}_{N_0}^\top \bar{\bm{y}}^*+ \bar{B}_{N_0}^\top (\bm{z}_1^*-\bm{z}_2^*)&\geq C^\top \bm{y}^*+B^\top \bm{z}^*-\Rad(N_0,\eps)\cdot\left(\|\bm{y}^*\|_1+\|\bm{z}^*\|_{1} \right)\\
&\geq C^\top \bm{y}^*+B^\top \bm{z}^*-\Rad(N_0,\eps)\cdot \frac{K+|\mathcal{S}|\cdot H}{\sigma^*}\\
&\geq\hat{\bm{r}}-\Rad(N_0,\eps)\cdot \frac{K+|\mathcal{S}|\cdot H}{\sigma^*}.
\end{aligned}\]
Thus, we know that $\bar{\bm{y}}^*$ and $\bm{z}_1^*, \bm{z}_2^*$ is also a feasible solution to $\text{Dual}_{N_0}(\lambda'_{N_0})$, and we have
\begin{equation}\label{eqn:1021705}
\text{Dual}'_{N_0}(\lambda'_{N_0})\leq\text{Dual}+\lambda_{N_0}\cdot(\|\bm{y}^*\|_1+\|\bm{z}^*\|_1)\leq \text{Dual}+\Rad(N_0,\eps)\cdot \frac{K+|\mathcal{S}|\cdot H}{\sigma^*}.
\end{equation}
Combining \eqref{eqn:1021705} with \eqref{eqn:1021704} and also noting that $\bar{V}_{N_0}(\lambda'_{N_0})=\text{Dual}'_{N_0}(\lambda'_{N_0})$, we have
\[
\bar{V}_{N_0}\leq V+\Rad(N_0,\eps)\cdot \frac{2(K+|\mathcal{S}|\cdot H)}{\sigma^*}+\Rad^2(N_0,\eps)\cdot\frac{K\cdot|\mathcal{S}|+|\mathcal{S}|^2\cdot H}{\sigma^*}
\]
which completes our proof.

\begin{myproof}[Proof of \Cref{claim:1dualBound}]
For the LP \eqref{lp:standard3}, we denote by $\mathcal{I}^*$ and $\mathcal{J}^*$ an optimal basis, and denote by $\bm{q}^*$ and $(\bm{y}^*, \bm{z}^*)$ the corresponding primal-dual optimal solution. Then, we have that
\begin{equation}\label{eqn:010518}
    C^\top(\mathcal{J}^*_1, \mathcal{I}^*)\bm{y}^*+B^\top(\mathcal{J}^*_2, \mathcal{I}^*)\bm{z}^*=\hat{\bm{r}}_{\mathcal{I}^*}.
\end{equation}
On the other hand, since the primal constraints in the set $(\mathcal{J}^*)^c$ are non-binding, we have $\bm{y}^*_{\mathcal{I}^*}=0$. Therefore, denote by
\begin{equation}\label{def:APrimematrix}
A^*=\left[\begin{aligned}
&C(\mathcal{J}^*_1, \mathcal{I}^*)\\
&B(\mathcal{J}^*_2, \mathcal{I}^*)
\end{aligned}\right].
\end{equation}
We know that
\[
\|(\bm{y}^*, \bm{z}^*)\|_1 = \| (A^*)^{-1}\cdot \hat{\bm{r}}_{\mathcal{I}^*} \|_1.
\]
We then define $\sigma^*$ as
\begin{equation}\label{def:sigmaPrime}
\sigma^*= \sigma_{\min}(A^*).
\end{equation}
Therefore, we have that
\[
\|(\bm{y}^*, \bm{z}^*)\|_1 \leq \frac{|\mathcal{S}|\cdot H+K}{\sigma^*}
\]
which completes our proof.
\end{myproof}

\subsection{Proof of \Cref{thm:1sample}}
The proof follows the proof of \Cref{thm:sample}. The only differences are that now the matrix size is bounded by $H\cdot|\mathcal{S}|+K$. To be specific, considering the LP in \eqref{lp:standard3}, denote by $K\times M$ the size of the matrix $C$ and denote by $R\times M$ the size of the matrix $B$.  

For the infinite horizon discounted problem considered in \Cref{sec:Infinitedis}, we have $M=|\mathcal{S}||\mathcal{A}|$ and $R=|\mathcal{S}|$. The sample complexity bound established in \Cref{thm:sample} essentially equals 
\[
N=O\left( \frac{M\cdot\log(1/\veps)}{\min\{\delta_1^2, \delta_2^2, \sigma_0^2\}}+\frac{m^7\cdot\log(1/\xi)}{\xi(\sigma^*)^3(1-\gamma)}\cdot\frac{\log^2(1/\veps)}{\veps} \right),
\]
where $\xi=\min_{(s,a)\in\mathcal{I}^*} \left\{ q^*(s,a) \right\}>0$ and $\bm{q}^*$ denotes the optimal solution to LP \eqref{lp:standard3} corresponding to the optimal basis $\mathcal{I}^*$ and $\mathcal{J}^*$, $\delta_1$ defined in \eqref{eqn:deltaGap}, $\delta_2$ defined in \eqref{eqn:deltaDual}, $\sigma_0$ defined in \eqref{def:sigma0}, $\sigma^*$ defined in \eqref{def:sigma}, and $m=|\mathcal{I}^*|=|\mathcal{J}^*|$. 

The bound established in \Cref{thm:sample} can be established exactly in the same way for the finite horizon episodic problem, except that the size of the matrix $C$ has now been changed to $K\times H|\mathcal{S}||\mathcal{A}|$ and the size of the matrix $B$ has now been changed to $H|\mathcal{S}|\times H|\mathcal{S}||\mathcal{A}|$. Then, the sample complexity bound can be established as 
\[
N=O\left( \frac{H|\mathcal{S}||\mathcal{A}|\cdot\log(1/\veps)}{\min\{\delta_1^2, \delta_2^2, \sigma_0^2\}}+\frac{m^7\cdot\log(1/\xi)}{\xi(\sigma^*)^3(1-\gamma)}\cdot\frac{\log^2(1/\veps)}{\veps} \right),
\]
where the parameters $\delta_1, \delta_2, \sigma_0, m, \xi$ and $\sigma^*$ are defined for the new LP \eqref{lp:standard3} corresponding to the finite horizon episodic problem.
Our proof is thus completed.

\subsection{Proof of \Cref{coro:OfflineSample}}
Note that in the proof of \Cref{thm:sample}, we have established that
\[
N_1=O\left( \frac{\log(1/\veps)}{\min\{\delta_1^2, \delta_2^2, \sigma_0^2\}}  \right),
\]
and
\[
N_2= O\left( \frac{m^5\cdot\log(1/\xi)}{\xi(\sigma^*)^3(1-\gamma)}\cdot\frac{\log^2(1/\veps)}{\veps}\right).
\]
We now derive the final sample complexity. 

First, note that in order to carry out \Cref{alg:Idenbasis} where we have a number of at least $N_1$ samples for each $(s,a)\in\mathcal{S}\times\mathcal{A}$, we need to obtain a number of $O(N_1/\mu_{\min})$ samples following the behavior policy, which can be directly established from the definition of $\mu_{\min}$ in \eqref{eqn:Mumin} and the Hoeffding's inequality. Therefore, the number of samples used for \Cref{alg:Idenbasis} can be bounded as
\[
N_1/\mu_{\min} = O\left( \frac{\log(1/\veps)}{\mu_{\min}\cdot\min\{\delta_1^2, \delta_2^2, \sigma_0^2\}}  \right).
\]

We then consider the resolving procedure. Note that in each iteration of the resolving procedure in \Cref{alg:Twophase}, we need to obtain one sample for each $(s,a)\in\mathcal{I}^*\times\mathcal{J}^*$, which can be relaxed to obtaining one sample for each $(s,a)\in\mathcal{S}\times\mathcal{A}$. Following the definition of $\mu_{\min}$ in \eqref{eqn:Mumin}, suppose we are obtaining the sample following the behavior policy, then by obtaining a number of samples bounded by $O(1/\mu_{\min})$ (in expectation), we know that we obtain at least one sample for each $(s,a)\in\mathcal{S}\times\mathcal{A}$. This implies that in expectation, the number of samples used for the resolving procedure in \Cref{alg:Twophase} can be bounded as
\[
N_2/\mu_{\min}= O\left( \frac{m^5\cdot\log(1/\xi)}{\mu_{\min}\cdot\xi(\sigma^*)^3(1-\gamma)}\cdot\frac{\log^2(1/\veps)}{\veps}\right).
\]
Our proof is thus completed.

\subsection{Proof of \Cref{lem:smallxi0}}
The proof is modified based on the proof of \Cref{lem:projection}. Denote by $\bm{q}^*$ the optimal solution corresponding to the optimal basis $\mathcal{I}^*$ and $\mathcal{J}^*$. Then, it holds that
\begin{equation}\label{eqn:11OptQ3}
\begin{bmatrix}
C(\mathcal{J}_1^*, \mathcal{I}^*)\\
B(\mathcal{J}_2^*, \mathcal{I}^*)
\end{bmatrix}
\cdot\bm{q}^*_{\mathcal{I}^*}=\begin{bmatrix}
\bm{\alpha}_{\mathcal{J}_1^*}\\
\bmu_{\mathcal{J}_2^*}
\end{bmatrix}.
\end{equation}
We compare $\tilde{\bm{q}}^n_{\mathcal{I}^*}$ with $\bm{q}^*_{\mathcal{I}^*}$ when $n$ is large enough. Note that for an $n$, $\tilde{\bm{q}}^n$ is the solution to the following linear equations
\begin{equation}\label{eqn:11OptQ4}
\begin{bmatrix}
\bar{C}^n(\mathcal{J}_1^*, \mathcal{I}^*)\\
\bar{B}^n(\mathcal{J}^*_2, \mathcal{I}^*)
\end{bmatrix}
\cdot\tilde{\bm{q}}^n_{\mathcal{I}^*}=\begin{bmatrix}
\frac{\bm{\alpha}^n_{\mathcal{J}_1^*}}{N_2-n+1}\\
\frac{\bmu^n_{\mathcal{J}_2^*}}{N_2-n+1}
\end{bmatrix}.
\end{equation}
When $n\leq\tau_0$, we know that
\begin{equation}\label{eqn:11020301}
\left| \bm{\alpha}_{\mathcal{J}_1^*}-\frac{\bm{\alpha}^n_{\mathcal{J}_1^*}}{N_2-n+1} \right|\leq\nu
\end{equation}
and
\begin{equation}\label{eqn:11020302}
\left| \bmu_{\mathcal{J}^*_2} - \frac{\bmu_{\mathcal{J}^*_2}^n}{N_2-n+1}  \right| \leq \nu.
\end{equation}
Moreover, denote by $\mathcal{E}^n$ the event that the absolute value of each element of $\bar{C}^n(\mathcal{J}_1^*, \mathcal{I}^*)-C(\mathcal{J}_1^*, \mathcal{I}^*)$, and $\bar{B}^n(\mathcal{J}^*_2, \mathcal{I}^*)-B(\mathcal{J}^*_2, \mathcal{I}^*)$ is upper bounded by $\Rad(n, \eps)$. Then, from the standard Hoeffding's inequality, we know that 
\[
P(\mathcal{E}^n)\geq 1-\eps.
\]
Conditional on the event $\mathcal{E}^n$ happens, 
we know that the absolute value of each element of $\bar{C}^n(\mathcal{J}_1^*, \mathcal{I}^*)-C(\mathcal{J}_1^*, \mathcal{I}^*)$, and $\bar{B}^n(\mathcal{J}^*_2, \mathcal{I}^*)-B(\mathcal{J}^*_2, \mathcal{I}^*)$ is upper bounded by $\Rad(n, \eps)$. We now bound the distance between the solutions to the linear equations \eqref{eqn:11OptQ3} and \eqref{eqn:11OptQ4}. The perturbation of the matrix is denoted as
\[
\Delta A^*=\begin{bmatrix}
C(\mathcal{J}_1^*, \mathcal{I}^*)-\bar{C}^n(\mathcal{J}_1^*, \mathcal{I}^*)\\
B(\mathcal{J}^*_2, \mathcal{I}^*)-\bar{B}(\mathcal{J}^*_2, \mathcal{I}^*)
\end{bmatrix}.
\]
Clearly, it holds that
\begin{equation}\label{eqn:11020303}
\|\Delta A^*\|_2 \leq \Rad(n, \eps)\cdot m^2,
\end{equation}
where $m=|\mathcal{I}^*|=|\mathcal{J}^*|$.
Therefore, as long as
\begin{equation}\label{eqn:11020304}
\|\Delta A^*\|_2 \leq\Rad(n, \eps)\cdot m^2 \leq \frac{1}{2\|(A^*)^{-1}\|_2}\leq \frac{1}{2\sigma^*},
\end{equation}
following standard perturbation analysis of linear equations \citep{higham2002accuracy}, we have that
\begin{equation}\label{eqn:11020305}
\begin{aligned}
    \frac{\|\tilde{\bm{q}}^n_{\mathcal{I}^*}-\bm{q}^*_{\mathcal{I}^*}\|_2}{\|\bm{q}^*_{\mathcal{I}^*}\|_2}
    &\leq \frac{\kappa(A^*)}{1-\kappa(A^*)\cdot\frac{\|\Delta A^*\|_2}{\|A^*\|_2}}\cdot \left( \frac{\|\Delta A^*\|_2}{\|A^*\|_2}+\frac{m^2\cdot\nu}{\|[\bm{\alpha}_{\mathcal{J}_1^*}; \bmu_{\mathcal{J}_2^*}]\|_2} \right)\\
    &\leq 2\cdot\kappa(A^*)\cdot\left( \frac{\|\Delta A^*\|_2}{\|A^*\|_2}+\frac{m^2\cdot\nu}{\|[\bm{\alpha}_{\mathcal{J}_1^*}; \bmu_{\mathcal{J}^*_2}]\|_2} \right)\\
    &\leq 2\cdot\kappa(A^*)\cdot\left( \frac{\|\Delta A^*\|_2}{\|A^*\|_2}+\frac{m^2\cdot\nu}{1-\gamma} \right),
\end{aligned}
\end{equation}
where $\kappa(A^*)=\|A^*\|_2\cdot\|(A^*)^{-1}\|_2$ denotes the conditional number of $A^*$. The last inequality follows from $\|[\bm{\alpha}_{\mathcal{J}_1^*}; \bmu_{\mathcal{J}_2^*}]\|_2\geq1-\gamma$. Further, note that $\|\bm{q}^*_{\mathcal{I}^*}\|_2\leq 1$. Therefore, in order to satisfy the condition $\|\tilde{\bm{q}}^n_{\mathcal{I}^*}-\|_2\leq 2$, we only need the right hand side of \eqref{eqn:11020305} to be upper bounded by 1. Clearly, as long as $n$ satisfies the condition \eqref{eqn:11020304} and the following condition
\begin{equation}\label{eqn:11020306}
2\cdot\kappa(A^*)\cdot \frac{\|\Delta A^*\|_2}{\|A\|_2}\leq 2\cdot\frac{\Rad(n,\eps)\cdot m^2}{\sigma^*} \leq \frac{1}{2},
\end{equation}
we only need to select a $\nu$ such that
\begin{equation}\label{eqn:11020307}
2\cdot\kappa(A^*)\cdot \frac{m^2\cdot\nu}{1-\gamma} \leq\frac{1}{2}.
\end{equation}
Note that this implies that $\bm{q}^n = \tilde{\bm{q}}^n$ for $n$ such that $N_4\leq n\leq\tau_0$ where $N_4$ satisfies the conditions in \eqref{eqn:11020304} and \eqref{eqn:11020306}. 
Moreover, in order to guarantee the condition that
\[
\|\bm{q}^n-\bm{q}^*\|_{\infty}=\|\tilde{\bm{q}}^n-\bm{q}^*\|_{\infty}\leq \|\tilde{\bm{q}}^n-\bm{q}^*\|_{\infty}\leq\frac{\xi}{4},
\]
we need the right hand side of \eqref{eqn:11020305} to be upper bounded by $\frac{\xi}{4}$. Clearly, as long as the following condition 
\begin{equation}\label{eqn:111501}
2\cdot\kappa(A^*)\cdot \frac{\|\Delta A^*\|_2}{\|A\|_2}\leq 2\cdot\frac{\Rad(n,\eps)\cdot m^2}{\sigma^*} \leq \frac{\xi}{8},
\end{equation}
we only need to select a $\nu$ such that
\begin{equation}\label{eqn:111502}
2\cdot\kappa(A^*)\cdot \frac{m^2\cdot\nu}{1-\gamma} \leq\frac{\xi}{8}.
\end{equation}
Combining \eqref{eqn:11020304}, \eqref{eqn:11020306}, and \eqref{eqn:111501}, we know that $n$ needs to satisfy the following condition
\begin{equation}\label{eqn:111503}
n\geq N_4:= \frac{256m^4}{(\xi\sigma^*)^2}\cdot\log(1/\eps).
\end{equation}
Also, $\nu$ is selected to satisfy the following condition
\begin{equation}\label{eqn:111504}
\nu\leq \nu_1:= \frac{\xi\sigma^*\cdot(1-\gamma)}{32m^4}.
\end{equation}
Our proof is thus completed.

\subsection{Proof of \Cref{thm:onlinesample}}
We specify the value of $\eps$ as $\eps=O(\veps^2/(K\cdot|\mathcal{S}||\mathcal{A}|))$.
We denote by $\mathcal{E}_0$ the event that \Cref{alg:Idenbasis} has successfully identified the optimal basis $\mathcal{I}^*$ and $\mathcal{J}^*$, in the execution of \Cref{alg:TwophaseOnline}. Note that in step 2 of \Cref{alg:TwophaseOnline}, we follow a stationary policy to obtain samples over each $(s,a)\in\mathcal{S}\times\mathcal{A}$. Then, following the definition of $\mu_{\min}$ in \eqref{eqn:Mumin}, we obtain $O(N_1\mu_{\min}\cdot\log(1/\eps))$ a number of samples for each $(s,a)\in\mathcal{S}\times\mathcal{A}$, with a probability at least $1-|\mathcal{S}||\mathcal{A}|\cdot\eps$. 
Therefore, from \Cref{thm:Infibasis2}, we know that for $N_1$ satisfying
\[
N_1=O\left( \frac{\log(1/\veps)}{\mu_{\min}\cdot\min\{\delta_1^2, \delta_2^2, \sigma_0^2\}} \right)
\]
it holds that
\[
P(\mathcal{E}_0)\geq 1-((K+1)\cdot|\mathcal{S}||\mathcal{A}|-|\mathcal{S}|^2|\mathcal{A}|)\cdot\eps.
\]
Conditional on the event $\mathcal{E}_0$ happens, we now assume that for each round $n\in[N_2]$, we are able to collect at least one sample for each $(s,a)\in\mathcal{I}^*$. We later specify the limit $N_3$ to characterize the number of samples needed for each iteration in the resolving procedure in \Cref{alg:TwophaseOnline}.

For any $n\in[N_2]$, we denote by $\mathcal{E}_n$ the event that $|\bar{r}_{n}(s,a)-\hat{r}(s,a)|$, $|\bar{c}_{k, n}(s,a)-\hat{c}_k(s,a)|$, and $|\bar{P}_{n}(s'|s,a)-P(s'|s,a)|$ for each $k\in[K]$, $(s,a)\in\mathcal{S}\times\mathcal{A}$ and $s'\in\mathcal{S}$ are all bounded by $\Rad(n, \eps)$. From standard Hoeffding's inequality, we know that 
\[
P(\mathcal{E}_n)\geq 1-(K|\mathcal{S}||\mathcal{A}|-|\mathcal{S}|^2|\mathcal{A}|)\cdot\eps.
\]
for any $n\in[N_2]$. We now denote by the event $\mathcal{E}$ that
\[
\mathcal{E}=\mathcal{E}_0\cap\left( \cap_{n=1}^{N_2}\mathcal{E}_n \right).
\]
We know that
\[
P(\mathcal{E})\geq 1-(N_2+1)\cdot (K|\mathcal{S}||\mathcal{A}|-|\mathcal{S}|^2|\mathcal{A}|)\cdot\eps.
\]
In what follows, we condition on that the event $\mathcal{E}$ happens, which implies that $\hat{\mathcal{I}}=\mathcal{I}^*$ and $\hat{\mathcal{J}}=\mathcal{J}^*$. 

We consider the stochastic process $\tilde{\alpha}_k(n)$ and $\tilde{\mu}_s(n)$ as follows 
\begin{equation}\label{eqn:AMuaverage}
\tilde{\alpha}_k(n)=\frac{\alpha^n_k}{N_2-n},~\forall k\in\mathcal{J}_1^*\text{~and~} \tilde{\mu}_s(n)=\frac{\mu^n_s}{N_2-n}, ~\forall s\in\mathcal{J}^*_2, \forall n\in[N_2].
\end{equation}
For a fixed $\nu>0$ that corresponds to the value of $\nu$ in \Cref{lem:smallxi0}, we define a set
\begin{equation}\label{eqn:11defX}
\mathcal{X}=\{ \bm{\alpha}'\in\mathbb{R}^{K}: \alpha'_k\in[\alpha_k-\nu, \alpha_k+\nu], \forall k\in\mathcal{J}_1^* \},
\end{equation}
and
\begin{equation}\label{eqn:11defY}
\mathcal{Y}=\{ \bm{\mu}'\in\mathbb{R}^{|\mathcal{S}|}: \mu'_s\in[\mu_s-\nu, \mu_s+\nu], \forall s\in\mathcal{J}^*_2 \}.
\end{equation}
It is easy to see that initially, $\tilde{\bm{\alpha}}(1)\in\mathcal{X}$ and $\tilde{\bmu}(1)\in\mathcal{Y}$. We show that $\tilde{\bm{\alpha}}(n)$ and $\tilde{\bmu}(n)$ behave well as long as they stay in the region $\mathcal{X}$ and $\mathcal{Y}$ for a
sufficiently long time. To this end, we define a stopping time
\begin{equation}\label{eqn:11Stoptime}
\tau_0=\min_{n\in[N_2]}\{ \tilde{\bm{\alpha}}(n) \notin\mathcal{X}\text{~or~}\tilde{\bmu}(n)\notin\mathcal{Y} \}
\end{equation}
which is exactly the same as the stopping time $\tau_0$ defined in \Cref{lem:smallxi0}. Note that following the proof of \Cref{lem:smallxi0}, we in fact prove the following result.
\begin{lemma}\label{lem:onlineprojection}
There exist two constants $N_4$ and $\nu_1$. When $N_4\leq n\leq\tau_0$, and $\nu\leq\nu_1$, it holds that
\[
\|\tilde{\bm{q}}_{\mathcal{I}^*}^n\|_1\leq2,~~\tilde{\bm{q}}^n_{\mathcal{I}^*}\geq\frac{\xi}{2},\text{~~and~~}\|\bm{q}^n - \bm{q}^*\|_{\infty}\leq\frac{\xi}{2}
\] 
where $\tilde{\bm{q}}_{\mathcal{I}^*}^n$ and $\bm{q}^n$ are the solutions obtained in \Cref{alg:TwophaseOnline}. Specifically, $N_4$ is given as follows
\begin{equation}\label{eqn:N0primeonline}
N_4\geq\frac{256m^4}{(\xi\sigma^*)^2}\cdot\log(1/\eps),
\end{equation}
where $m=|\mathcal{I}^*|=|\mathcal{J}^*|$.
Also, $\nu_1$ is given as follows
\begin{equation}\label{eqn:nu1}
\nu_1= \frac{\xi\sigma^*\cdot(1-\gamma)}{32m^4}.
\end{equation}
\end{lemma}
We set $\nu$ satisfy the condition $\nu\leq\nu_1$ with $\nu_1$ satisfies the condition in \Cref{lem:onlineprojection}.
We bound $\mathbb{E}[N-\tau_0]$ in the following lemma.
\begin{lemma}\label{lem:onlineStoptime}
Let the stopping time $\tau_1$ be defined in \eqref{eqn:11Stoptime}. It holds that
\[
\mathbb{E}[N_2-\tau_0]\leq N_4+
4m\cdot \exp(-\nu^2/8)
\]
where $N_4$ is given in \eqref{eqn:N0primeonline}, as long as
\begin{equation}\label{eqn:11022104}
N_2\geq N_4\text{~~and~~}N_2\geq \frac{8}{\nu^2}\geq \frac{8}{\nu^2_1}=\frac{8192m^8}{(\xi\sigma^*)^2\cdot(1-\gamma)^2}.
\end{equation}
Also, for any $N'$ such that $N_4\leq N'\leq N_2$, it holds that
\begin{equation}\label{eqn:HighProbtau}
P(\tau\leq N')\leq \frac{m\cdot\nu^2}{2}\cdot \exp\left( -\frac{\nu^2\cdot(N_2-N'+1)}{8 } \right).
\end{equation}
\end{lemma}
From the definition of the stopping time $\tau_0$ in \eqref{eqn:11Stoptime}, we know that for each $k\in\mathcal{J}_1^*$, it holds
\[
\alpha_k^{\tau-1}\in[(N_2-\tau_0+1)\cdot(\alpha_k-\nu), ~~~(N_2-\tau_0+1)\cdot(\alpha_k+\nu)]
\]
Thus, we have that
\begin{equation}\label{eqn:11011906}
|\alpha_k^{N_2}|\leq |\alpha_k^{\tau_0-1}|+ \sum_{t=\tau_0}^{N_2}\sum_{(s,a)\in\mathcal{I}^*}c_k^n(s,a)\cdot q^n(s,a)
\end{equation}
and thus
\begin{equation}\label{eqn:11011907}
\left|\mathbb{E}[\alpha_k^{N_2}]\right| \leq 4\mathbb{E}[N_2-\tau_0]\leq 4N_4+
16m\cdot \exp(-\nu^2/8).
\end{equation}
Following the same procedure, we can show that for each $s\in\mathcal{J}^*_2$, it holds that
\begin{equation}\label{eqn:11011908}
\left|\mathbb{E}[\mu^{N_2}(s)]\right|\leq 4\mathbb{E}[N_2-\tau_0]\leq 4N_4+
16m\cdot \exp(-\nu^2/8).
\end{equation}
We finally consider the other constraints $k\in(\mathcal{J}_1^{*})^c$ and $s\in(\mathcal{J}^*_2)^c$. 
Note that following the definition of $\bm{\alpha}^n$ and $\bmu^n$, we have that
\begin{equation}\label{eqn:011909}
A^*\cdot\left(\sum_{n=1}^{N_2}\mathbb{E}[\bm{q}^n_{\mathcal{I}^*}]\right)=\left[\bm{\alpha}_{\mathcal{J}_1^*}^1-\mathbb{E}\left[\bm{\alpha}^{N_2}_{\mathcal{J}_1^*} \right]; \bmu^1_{\mathcal{J}^*_2}-\mathbb{E}\left[\bmu_{\mathcal{J}^*_2}^{N_2} \right]\right].
\end{equation}
Also, from the binding of $\bm{q}^*$ regarding the optimal basis $\mathcal{I}^*$ and $\mathcal{J}^*$, we have
\begin{equation}\label{eqn:011910}
    N_2\cdot A^*\cdot\bm{q}^*_{\mathcal{I}^*}=[\bm{\alpha}^1_{\mathcal{J}_1^*}; \bmu_{\mathcal{J}^*_2}^1].
\end{equation}
Therefore, it holds that
\begin{equation}\label{eqn:011911}
\sum_{n=1}^{N_2}\mathbb{E}\left[\bm{q}^n_{\mathcal{I}^*} \right]=N_2\cdot\bm{q}^*_{\mathcal{I}^*}-(A^*)^{-1}\cdot \left[\mathbb{E}\left[\bm{\alpha}_{\mathcal{J}_1^*}^{N_2} \right]; \mathbb{E}\left[\bmu^{N_2}_{\mathcal{J}^*_2} \right]\right],
\end{equation}
and
\begin{equation}\label{eqn:022202}
\left\|\sum_{n=1}^{N_2}\mathbb{E}\left[\bm{q}^n_{(\mathcal{I}^{*})^c} \right]\right\|_1=0.
\end{equation}
Finally, for any $k\in(\mathcal{J}_1^{*})^c$, we have
\begin{equation}\label{eqn:011912}
\begin{aligned}
(\hat{\bm{c}}_k)^\top \left(\sum_{n=1}^{N_2}\mathbb{E}\left[\bm{q}^n \right]\right)&= N_2\cdot(\hat{\bm{c}}_k)^\top\bm{q}^*_{\mathcal{I}^*}-(\hat{\bm{c}}_k)^\top\cdot (A^*)^{-1}\cdot \left[\mathbb{E}\left[\bm{\alpha}_{\mathcal{J}_1^*}^{N_2}\right];\mathbb{E}\left[\bmu_{\mathcal{J}_2}^{N_2}\right]\right]\\
&=N_2\cdot(\hat{\bm{c}}_k)^\top\bm{q}^*-(\hat{\bm{c}}_k)^\top\cdot (A^*)^{-1}\cdot \left[\mathbb{E}\left[\bm{\alpha}_{\mathcal{J}_1^*}^{N_2}\right];\mathbb{E}\left[\bmu_{\mathcal{J}_2^*}^{N_2}\right]\right]
\end{aligned}
\end{equation}
From the feasibility of $\bm{q}^*$, we know that
\[
N_2\cdot\alpha_k\geq N_2\cdot(\hat{\bm{c}}_k)^\top\bm{q}^*.
\]
Therefore, for any $k\in(\mathcal{J}_1^{*})^c$, it holds that
\begin{equation}\label{eqn:1120301}
\begin{aligned}
N_2\cdot\alpha_k-\sum_{n=1}^{N_2}\hat{\bm{c}}_k^\top\mathbb{E}\left[\bm{q}^n \right] &\leq (\hat{\bm{c}}_k)^\top\cdot (A^*)^{-1}\cdot \left[\mathbb{E}[\bm{\alpha}_{\mathcal{J}_1^*}^{N_2}]; \mathbb{E}[\bmu_{\mathcal{J}^*_2}^{N_2}]\right]\\
&\leq \frac{m}{\sigma^*}\cdot \left( 4N_4+
16m\cdot \exp(-\nu^2/8) \right).
\end{aligned}
\end{equation}
In the same way, we can prove that
\begin{equation}\label{eqn:11110601}
\left\|N_2\cdot\bmu-B\cdot\sum_{n=1}^{N_2}\mathbb{E}[\bm{q}^n] \right\|_{\infty}\leq \frac{m}{\sigma^*}\cdot \left( 4N_4+
16m\cdot \exp(-\nu^2/8) \right).
\end{equation}
Moreover, the definition of $\sigma^*$ in \eqref{def:sigma} implies that following upper bound on the norm of the dual variable $\bm{y}^*$ and $\bm{z}^*$.
\begin{equation}\label{eqn:022105}
\|\bm{y}^*\|_2+\|\bm{z}^*\|_2=\|(A^{*\top})^{-1}\cdot\hat{\bm{r}}_{\mathcal{J}^*}  \|_2\leq\frac{m^2}{\sigma^*}.
\end{equation}
Therefore, we know that the regret over the reward and the regret over the constraint violation can all be bounded by using \eqref{eqn:11011907}, \eqref{eqn:11011908}, \eqref{eqn:1120301}, and \eqref{eqn:11110601}. We present the bounds as follows.
\begin{equation}\label{eqn:111601}
N_2\cdot V_r(\pi^*, \mu_1)-\sum_{n=1}^N \hat{\bm{r}}^\top\mathbb{E}[\bm{q}^n]\leq \frac{m^3}{\sigma^*}\cdot \left( 4N_4+
16m\cdot \exp(-\nu^2/8)\right)\leq \frac{16m^4}{\sigma^*}+\frac{1024m^7}{\xi^2(\sigma^*)^3}\cdot\log(1/\eps)
\end{equation}
and 
\begin{equation}\label{eqn:111602}
N_2\cdot\alpha_k-\sum_{n=1}^{N_2}\hat{\bm{c}}_k^\top\mathbb{E}\left[\bm{q}^n \right] \leq \frac{16m^4}{\sigma^*}+\frac{1024m^7}{\xi^2(\sigma^*)^3}\cdot\log(1/\eps)
\end{equation}
for each $k\in[K]$ and 
\begin{equation}\label{eqn:111603}
\left\|N_2\cdot\bmu-B\cdot\sum_{n=1}^{N_2}\mathbb{E}[\bm{q}^n] \right\|_{\infty}\leq \frac{16m^4}{\sigma^*}+\frac{1024m^7}{\xi^2(\sigma^*)^3}\cdot\log(1/\eps)
\end{equation}
as long as the number of rounds $N_2$ satisfies that
\begin{equation}\label{eqn:111604}
N_2\geq \frac{8192m^8}{(\xi\sigma^*)^2\cdot(1-\gamma)^2}\cdot\log(1/\eps).
\end{equation}
We now transfer the bound in \eqref{eqn:111601}, \eqref{eqn:111602}, and \eqref{eqn:111603} into the bound over the policy $\bar{\pi}$ defined in \eqref{eqn:11012501}. We have the following result, which follows from the procedure of \Cref{thm:sample} and we provide the formal proof at the end of this proof for completeness.
\begin{lemma}\label{lem:onlineSample}
Suppose that the bounds in \eqref{eqn:111601}, \eqref{eqn:111602}, and \eqref{eqn:111603} hold. Then, for the policy $\bar{\pi}$ defined in \eqref{eqn:11012501}, and for a number of rounds $N_2$ satisfies that
\[
N_2=O\left( \frac{m^5\cdot\log(1/\xi)}{(\xi\sigma^*)^3(1-\gamma)}\cdot\frac{\log^2(1/\veps)}{\veps}\right)
\]
we have that
\[
V_r(\pi^*, \mu_1) - V_r(\bar{\pi}, \mu_1)\leq\veps\text{~~and~~}V_k(\bar{\pi}, \mu_1) - \alpha_k\leq\veps,~\forall k\in[K].
\]
\end{lemma}

We now bound the final sample complexity of \Cref{alg:TwophaseOnline}. It only remains to bound the number of samples needed in each iteration of the resolving procedure to obtain at least one sample of $(s,a)\in\mathcal{I}^*$, which also gives the value of $N_3$ to be set as the limit of sampling steps. Denote by $\mathcal{I}^*_S$ a collection of the states in that
\[
\mathcal{I}^*_{S} = \{ s\in\mathcal{S}: \text{~there~exists~}a\in\mathcal{A}\text{~such~that~}(s,a)\in\mathcal{I}^* \}.
\]
We first show that initialize from a state $s\in\mathcal{I}^*_S$, we will never reach a state $s\notin\mathcal{I}^*_S$ during the resolving procedure. Suppose that following the resolving steps in \Cref{alg:TwophaseOnline}, we arrive at a state $s\in\mathcal{I}^*_S$, and transit to a state $s'\notin\mathcal{I}^*_S$ by taking an action $a$ such that $(s,a)\in\mathcal{I}^*$. Then, if we follow the optimal policy given by the occupancy measure $\bm{q}^*$, there will also be a positive probability that we transit to $s'$ since we will arrive at the state $s$ and take the action $a$ with a positive probability, which contradicts with the condition that $\bm{q}^*_{(\mathcal{I}^*)^c}=0$. Therefore, we know that we will only visit states in the set $\mathcal{I}^*_S$ during the resolving procedure of \Cref{alg:TwophaseOnline}. 
We further note that for any $n\in[N_2]$, since it holds that $\bm{q}^n_{\mathcal{I}^*}\geq\xi'\geq\frac{\xi}{4}$, we must have
\[
P(\bar{\pi}^n(s)=a)\geq\xi'\geq\frac{\xi}{4},~~~\forall (s,a)\in\mathcal{I}^*.
\]
Therefore, we conclude that for any $n\in[N_2]$, we have that
\[
\bar{\pi}^n\in\Pi.
\]
Following the definition of the smallest mixing time of $t_{\mathrm{mix}}$ in \eqref{eqn:mixingtime}, we know that for each iteration $n\in[N_2]$, after following the policy $\bar{\pi}^n$ by $O(t_{\mathrm{mix}})$ number of steps, the resulting state distribution will be close to the stationary distribution of the policy $\bar{\pi}^n$. Further from the definition of the smallest occupancy measure $\xi_{\mathrm{small}}$ in \eqref{eqn:SmallOccupMeasure}, we know that for each iteration $n\in[N_2]$, after following the policy $\bar{\pi}^n$ by $O(t_{\mathrm{mix}})$ number of steps, we have a positive probability lower bounded by $O(\xi_{\mathrm{small}})$ to visit each $(s,a)\in\mathcal{I}^*$. This further implies that by following the policy $\bar{\pi}^n$ by a $O(t_{\mathrm{mix}}/\xi_{\mathrm{small}})$ number of steps, we obtain at least one sample of each $(s,a)\in\mathcal{I}^*$, with a high probability. Therefore, we can set the value of $N_3$ as $N_3=O(t_{\mathrm{mix}}/\xi_{\mathrm{small}})$. 

Note that the number of samples needed in step 2 of \Cref{alg:TwophaseOnline} for calling \Cref{alg:Idenbasis} can be bounded as
\[
N_1=O\left( \frac{\log(1/\veps)}{\mu_{\min}\cdot\min\{\delta_1^2, \delta_2^2, \sigma_0^2\}} \right),
\]
where $\mu_{\min}$ is given in \eqref{eqn:Mumin}. Given the bound on $N_2$ in \Cref{lem:onlineSample}, we know that the total number of samples (in expectation) needed in the resolving procedure of \Cref{alg:TwophaseOnline} can be bounded as 
\[
O\left( \frac{t_{\mathrm{mix}}\cdot m^5\cdot\log(1/\xi)}{\xi_{\mathrm{small}}\cdot(\xi\sigma^*)^3(1-\gamma)}\cdot\frac{\log^2(1/\veps)}{\veps}\right).
\]
Therefore, we conclude that our \Cref{alg:TwophaseOnline} enjoys a sample complexity of 
\[
N=O\left( \frac{\log(1/\veps)}{\mu_{\min}\cdot\min\{\delta_1^2, \delta_2^2, \sigma_0^2\}}+\frac{t_{\mathrm{mix}}\cdot m^5\cdot\log(1/\xi)}{\xi_{\mathrm{small}}\cdot(\xi\sigma^*)^3(1-\gamma)}\cdot\frac{\log^2(1/\veps)}{\veps} \right),
\]
which completes our proof.

\subsection{Proof of \Cref{lem:onlineStoptime}}
Now we fix a $k\in\mathcal{J}_1^*$. For any $N'$ such that $N_4\leq N'\leq N_2$, it holds that
\[
\tilde{\alpha}_k(N')-\tilde{\alpha}_k(N_3)=\sum_{n=N_4}^{N'-1}(\tilde{\alpha}_k(n+1)-\tilde{\alpha}_k(n)).
\]
We define $\xi_k(n)=\tilde{\alpha}_k(n+1)-\tilde{\alpha}_k(n)$. Then, we have
\[
\tilde{\alpha}_k(N')-\tilde{\alpha}_k(N_4)=\sum_{n=N_4}^{N'-1}(\xi_k(n)-\mathbb{E}[\xi_k(n)|\mathcal{F}_n])+\sum_{n=N_4}^{N'-1}\mathbb{E}[\xi_k(n)|\mathcal{F}_n].
\]
where $\mathcal{F}_n$ denotes the filtration of information up to step $n$.
Note that due to the update of $\tilde{\alpha}_k(n)$ given as follows
\[
\tilde{\alpha}_k(n+1)=\tilde{\alpha}_k(n)-\frac{\sum_{(s,a)\in\mathcal{I}^*}\bm{c}^n(s,a)\cdot q^n(s,a)-\tilde{\alpha}_k(n)}{N_2-n-1},~\forall k\in\mathcal{J}_1^*,
\]
we have
\[
\xi_k(n)=\frac{\tilde{\alpha}_k(n)-\sum_{(s,a)\in\mathcal{I}^*}c_k^n(s,a)\cdot q^n(s,a)}{N_2-n-1}.
\]
Then, it holds that
\begin{equation}\label{eqn:11011901}
|\xi_k(n)-\mathbb{E}[\xi_k(n)|\mathcal{F}_n]|\leq \frac{2}{N_2-n+1}
\end{equation}
where the inequality follows from the fact that the value of $\tilde{\alpha}_k(n)$ is deterministic given the filtration $\mathcal{F}_n$ and $\|\bm{q}^n\|_1\leq 2$ for any $n$.
Note that
\[
\{\xi_k(n)-\mathbb{E}[\xi_k(n)|\mathcal{F}_n]\}_{\forall n=N_4,\dots,N'}
\]
forms a martingale difference sequence. Following Hoeffding's inequality, for any $N''\leq N'$ and any $b>0$, it holds that
\[\begin{aligned}
P\left( \left| \sum_{n=N_4}^{N''}(\xi_k(n)-\mathbb{E}[\xi_k(n)|\mathcal{F}_n]) \right|\geq b \right)&\leq 2\exp\left( -\frac{b^2}{2\cdot\sum_{n=N_4}^{N''}1/(N_2-n+1)^2 } \right)\\
&\leq 2\exp\left( -\frac{b^2\cdot(N_2-N''+1)}{2 } \right).
\end{aligned}\]
Therefore, we have that
\begin{equation}\label{eqn:11011902}
\begin{aligned}
&P\left( \left| \sum_{n=N_4}^{N''}(\xi_k(n)-\mathbb{E}[\xi_k(n)|\mathcal{F}_n]) \right|\geq b \text{~for~some~}N_4\leq N''\leq N' \right)\\
\leq &\sum_{N''=N_4}^{N'} 2\exp\left( -\frac{b^2\cdot(N_2-N''+1)}{2 } \right)\leq b^2\cdot \exp\left( -\frac{b^2\cdot(N_2-N'+1)}{2 } \right)
\end{aligned}
\end{equation}
holds for any $b>0$.

We now bound the probability that $\tau>N'$ for one particular $N'$ such that $N_4\leq N'\leq N$. Suppose that $N'\leq \tau$, then,  from Lemma \ref{lem:projection}, for each $n\leq N'$, we know that $\|\tilde{\bm{q}}^n\|_2\leq 2$ and therefore $\bm{q}^n=\tilde{\bm{q}}^n$ as the solution to \eqref{eqn:11OptQ2}. We have
\[
\tilde{\alpha}_k(n)=\sum_{(s,a)\in\mathcal{I}^*}\bar{c}_{k,n}(s,a)\cdot q^n(s,a).
\]
It holds that
\begin{equation}\label{eqn:11011903}
\left|\mathbb{E}[\xi_k(n)|\mathcal{F}_n]\right|\leq\frac{1}{N_2-n+1}\cdot \sum_{(s,a)\in\mathcal{I}^*}q^n(s,a)\cdot \left|\mathbb{E}[\bar{c}_{k,n}(s,a)]-\hat{c}_{k}^n(s,a)\right|\leq\frac{2\Rad(n,\eps)}{N_2-n+1}.
\end{equation}
Then, we know that
\begin{equation}\label{eqn:11011904}
\begin{aligned}
\sum_{n=N_4}^{N'-1}\left|\mathbb{E}[\xi_k(n)|\mathcal{F}_n]\right|&\leq\sqrt{\frac{\log(2/\eps)}{2}}\cdot\sum_{n=N_4}^{N'-1}\frac{1}{\sqrt{n}\cdot(N_2-n)}\leq \sqrt{\frac{\log(2/\eps)}{2}}\cdot \sqrt{N'-1}\cdot \sum_{n=N_4}^{N'-1}\frac{1}{n\cdot(N_2-n)}\\
&=\sqrt{\frac{\log(2/\eps)}{2}}\cdot \frac{\sqrt{N'-1}}{N_2}\cdot \sum_{n=N_4}^{N'-1}\left( \frac{1}{n}+\frac{1}{N_2-n} \right)\\
&\leq \sqrt{2\log(2/\eps)}\cdot \frac{\sqrt{N'-1}}{N_2}\cdot \log(N_2)
\leq \frac{\sqrt{2\log(2/\eps)}}{\sqrt{N_2}}\cdot\log(N_2)
\leq\frac{\nu}{2}
\end{aligned}
\end{equation}
for a $N_2$ large enough such that
\begin{equation}\label{eqn:11022103}
N_2\geq \frac{8}{\nu^2}\geq \frac{8}{\nu^2_0}=\frac{8192m^8}{(\xi\sigma^*)^2\cdot(1-\gamma)^2}
\end{equation}
Combining \eqref{eqn:11011904} and \eqref{eqn:11011902} with $b=\nu/2$, and apply a union bound over all $k\in\mathcal{J}_1^*$ and $s\in\mathcal{J}^*_2$, we know that
\begin{equation}\label{eqn:11011905}
P(\tau\leq N')\leq \frac{m\cdot\nu^2}{2}\cdot \exp\left( -\frac{\nu^2\cdot(N_2-N'+1)}{8 } \right).
\end{equation}
Therefore, we know that
\[
\mathbb{E}[N_2-\tau]=\sum_{N'=1}^{N_2} P(\tau \leq N')\leq N_4+\sum_{N'=N_4}^{N_2} P(\tau \leq N')\leq N_4+
4m\cdot \exp(-\nu^2/8)
\]
which completes our proof.

\subsection{Proof of \Cref{lem:onlineSample}}
The caveat of directly transferring the bounds of \eqref{eqn:111601}, \eqref{eqn:111602}, and \eqref{eqn:111603} into the regret bounds \eqref{eqn:regret} of the policy $\bar{\pi}$ is that, the vector $\bar{\bm{q}}^{N_2}$ does not directly characterize an occupancy measure. This point can be seen by noting that there is a gap between $B\bar{\bm{q}}^{N_2}$ and $\bmu$, though bounded by $O(\log(N_2)/N_2)$ by setting $\eps=O(1/N_2^2)$. However, we can
show that the gap between $\bar{\bm{q}}^{N_2}$ and $\bm{q}^*$ is upper bounded by $O(\log(N_2)/N_2)$, which implies a bound over the gap between the policy $\bar{\pi}^{N_2}$ and the optimal policy $\pi^*$ that corresponds to the occupancy measure $\bm{q}^*$. This bound over the gap between the policy distributions can be then transferred into the bound over the gap between the state-value functions under the policy $\bar{\pi}$ and $\pi^*$. The regret bounds can be obtained then.

We first bound the gap between $\bar{\bm{q}}^{N_2}$ and $\bm{q}^*$. We have that
\begin{equation}\label{eqn:11020401}
\bar{\bm{q}}^{N_2}_{(\mathcal{I}^{*})^c}=\bm{q}^*_{(\mathcal{I}^{*})^c}=0.
\end{equation}
Also, note that following the definition of $\bm{\alpha}^n$ and $\bmu^n$, we have that
\begin{equation}\label{eqn:111011909}
A^*\cdot\left(\sum_{n=1}^{N_2}\mathbb{E}[\bm{q}^n_{\mathcal{I}^*}]\right)=\begin{bmatrix}
\bm{\alpha}_{\mathcal{J}_1^*}^1-\mathbb{E}\left[
\bm{\alpha}^{N_2}_{\mathcal{J}_1^*} \right]\\
\bmu^1_{\mathcal{J}^*_2}-\mathbb{E}\left[\bmu_{\mathcal{J}^*_2}^{N_2} \right]\end{bmatrix}.
\end{equation}
Also, from the binding of $\bm{q}^*$ regarding the optimal basis $\mathcal{I}^*$ and $\mathcal{J}^*$, we have
\begin{equation}\label{eqn:111011910}
    N_2\cdot A^*\cdot\bm{q}^*_{\mathcal{I}^*}=[\bm{\alpha}^1_{\mathcal{J}_1^*}; \bmu_{\mathcal{J}^*_2}^1].
\end{equation}
Then, from \eqref{eqn:111011909} and \eqref{eqn:111011910}, we know that
\begin{equation}
\begin{aligned}
\left\| \mathbb{E}\left[\bar{\bm{q}}^{N_2}_{\mathcal{I}^*}\right]-\bm{q}^*_{\mathcal{I}^*} \right\|_{\infty} &= \left\|\bm{q}^*_{\mathcal{I}^*}-\frac{1}{N_2}\cdot\sum_{n=1}^{N_2}\mathbb{E}\left[\bm{q}^n_{\mathcal{I}^*}\right]\right\|_{\infty}= \frac{\left\|(A^*)^{-1}\cdot \left[\mathbb{E}\left[\bm{\alpha}^{N_2}_{\mathcal{J}^*} \right]; \mathbb{E}\left[\bmu^{N_2}\right]\right] \right\|_{\infty}}{N_2}\\
&\leq \frac{\left\|\left[\mathbb{E}\left[\bm{\alpha}^{N_2}_{\mathcal{J}^*}\right]; \mathbb{E}\left[\bmu^{N_2}\right]\right]\right\|_{\infty}}{\sigma^*\cdot N_2}\leq\frac{1}{\sigma^*\cdot N_2}\cdot \left( 16m+\frac{1024m^4}{(\xi\sigma^*)^2}\cdot\log(1/\eps) \right).
\end{aligned}
\end{equation}
From Markov's inequality, for each $i\in\mathcal{I}^*$ and any $g>0$, we know that
\begin{equation}\label{eqn:11022107}
P\left( |\bar{\bm{q}}^{N_2}_{i}-\bm{q}^*_i|>g \right) \leq\frac{1}{g\cdot\sigma^*\cdot N_2}\cdot \left( 16m+\frac{1024m^4}{(\xi\sigma^*)^2}\cdot\log(1/\eps) \right).
\end{equation}
The policy $\bar{\pi}$ is essentially random by noting that $\bm{q}^{N_2}$ is a random variable, where the randomness comes from the randomness of the filtration $\mathcal{H}_{N_2}$. For each $s\in\mathcal{S}$ and $a\in\mathcal{A}$, we denote by $\bar{\pi}(a|s)$ the (ex-ante) probability that the random policy $\bar{\pi}$ takes the action $a$ given the state $s$. Then, for any $0<g\leq\xi/2$, we note that
\begin{equation}\label{eqn:11022108}
|\bm{q}^{N_2}_i-\bm{q}^*_i|\leq g \text{~for~each~}i=(s,a)\in\mathcal{I}^*\text{~implies~that~}\left|\frac{\bm{q}^{N_2}(s,a)}{\sum_{a'\in\mathcal{A}}\bm{q}^{N_2}(s,a')}-\frac{\bm{q}^*(s,a)}{\sum_{a'\in\mathcal{A}}\bm{q}^*(s,a')} \right|\leq\frac{2g}{\xi},
\end{equation}
for each $i=(s,a)\in\mathcal{I}^*$. For any $0<g\leq\xi/2$, note that
\begin{equation}\label{eqn:11022109}
P\left( |\bm{q}^{N_2}_i-\bm{q}^*_i|\leq g \text{~for~each~}i=(s,a)\in\mathcal{I}^* \right)\geq 1-\frac{m}{g\cdot\sigma^*\cdot N_2}\cdot \left( 16m+\frac{1024m^4}{(\xi\sigma^*)^2}\cdot\log(1/\eps) \right),
\end{equation}
where the inequality follows from the bound \eqref{eqn:11022107} and the union bound over $i\in\mathcal{I}^*$. Therefore, for any $0<g\leq\xi/2$ and any $(s,a)$, we know that
\begin{equation}\label{eqn:11022110}
P\left(\left|\bar{\pi}(a|s)-\pi^*(a|s)\right|\leq \frac{2g}{\xi}\right)\geq 1-\frac{m}{g\cdot\sigma^*\cdot N_2}\cdot \left( 16m+\frac{1024m^4}{(\xi\sigma^*)^2}\cdot\log(1/\eps) \right).
\end{equation}
From the above inequality, for any $(s,a)$, we have that
\begin{equation}\label{eqn:11022111}
\begin{aligned}
\left| \mathbb{E}\left[\bar{\pi}(a|s)\right]-\pi^*(a|s) \right|&\leq \mathbb{E}\left[ \left|\bar{\pi}(a|s)-\pi^*(a|s)\right| \right]\leq\frac{2}{N_2\xi}+\frac{2}{\xi}\cdot\int_{g=\frac{1}{N_2}}^{\xi/2}P\left(\left|\bar{\pi}(a|s)-\pi^*(a|s)\right|\geq \frac{2g}{\xi}\right)dg\\
&\leq \frac{2}{N_2\xi}+\frac{2m}{\xi\cdot\sigma^*\cdot N_2}\cdot \left( 16m+\frac{1024m^4}{(\xi\sigma^*)^2}\cdot\log(1/\eps) \right)\cdot \int_{g=\frac{1}{N_2}}^{\xi/2}\frac{dg}{g}\\
&=\frac{2}{N_2\xi}+\frac{2m}{\xi\cdot\sigma^*\cdot N_2}\cdot \left( 16m+\frac{1024m^4}{(\xi\sigma^*)^2}\cdot\log(1/\eps) \right)\cdot\left( \log(N_2)-\log(\xi/2) \right).
\end{aligned}
\end{equation}
We finally transfer the bound \eqref{eqn:11022111} into the regret bounds of policy $\bar{\pi}$. We use the state-value functions $V_r(\pi, s)$, defined for any initial state $s$ and any policy $\pi$ as follows
\begin{equation}\label{def:11stateValue}
V_r(\pi, s)=\mathbb{E}\left[ \sum_{t=0}^{\infty}\gamma^t\cdot r(s_t, a_t)\mid s \right],
\end{equation}
where $(s_t, a_t)$ is generated according to the policy $\pi$ and the transition kernel $P$ with the initial state $s$. Note that the value of $V_r(\pi, s)$ for any $s\in\mathcal{S}$ can be obtained from solving Bellman's equation under policy $\pi$
\begin{equation}\label{eqn:11Bellman}
V_r(\pi, s)=\mathbb{E}_{a\sim \pi(\cdot|s)}\left[ \hat{r}(s,a)+\gamma\cdot \mathbb{E}_{s'\sim P(\cdot|s,a)}[V_r(\pi, s')] \right].
\end{equation}
We define a matrix $B^{\pi}\in\mathbb{R}^{|\mathcal{S}|\times|\mathcal{S}|}$ such that the $s$-th row $s'$-th column element is
\begin{equation}\label{def:11Api}
B^{\pi}(s,s')=\delta_{s,s'}-\gamma\cdot \sum_{a\in\mathcal{A}}\pi(a|s)\cdot P(s'|s,a).
\end{equation}
Then, the matrix $B^{\pi}$ represents the state transition probability matrix under the policy $\pi$.
Denote by
\[
\bm{V}_r(\pi)=(V_r(\pi, s))_{\forall s\in\mathcal{S}}
\]
and
\[
\hat{\bm{r}}(\pi)=(\sum_{a\in\mathcal{A}}\pi(a|s)\cdot\hat{r}(s,a) )_{\forall s\in\mathcal{S}}.
\]
We have that the state values $\bm{V}_r(\pi)$ is the solution to the linear equation
\begin{equation}\label{eqn:11020405}
    B^{\pi}\bm{V}_r(\pi)=\hat{\bm{r}}(\pi)
\end{equation}
To bound the regret, we bound the solution to the linear equation \eqref{eqn:11020405} with $\pi$ being $\bar{\pi}$ and $\pi^*$ separately. The perturbation of the right hand of the equation \eqref{eqn:11020405} is
\[
\Delta \hat{\bm{r}}=\hat{\bm{r}}(\bar{\pi})-\hat{\bm{r}}(\pi^*).
\]
Clearly, we have that
\begin{equation}\label{eqn:11020408}
\|\Delta \hat{\bm{r}} \|_{\infty} \leq \frac{2}{N_2\xi}+\frac{2m}{\xi\cdot\sigma^*\cdot N_2}\cdot \left( 16m+\frac{1024m^4}{(\xi\sigma^*)^2}\cdot\log(1/\eps) \right)\cdot\left( \log(N_2)-\log(\xi/2) \right).
\end{equation}
The perturbation of the matrix is denoted as
\[
\Delta B=B^{\bar{\pi}}-B^{\pi^*}.
\]
Clearly, it holds that
\begin{equation}\label{eqn:11020406}
\|\Delta B\|_{\infty} \leq\frac{2\gamma}{N_2\xi}+\frac{2\gamma m}{\xi\cdot\sigma^*\cdot N_2}\cdot \left( 16m+\frac{1024m^4}{(\xi\sigma^*)^2}\cdot\log(1/\eps) \right)\cdot\left( \log(N_2)-\log(\xi/2) \right).
\end{equation}
Therefore, as long as
\begin{equation}\label{eqn:11020407}
\begin{aligned}
 &\frac{2}{N_2\xi}+\frac{2m}{\xi\cdot\sigma^*\cdot N_2}\cdot \left( 16m+\frac{1024m^4}{(\xi\sigma^*)^2}\cdot\log(1/\eps) \right)\cdot\left( \log(N_2)-\log(\xi/2) \right) \\
 \leq &1/\|(B^{\pi^*})^{-1}\|_{\infty}=1/\sigma',
\end{aligned}
\end{equation}
following standard perturbation analysis of linear equations \citep{higham2002accuracy}, we have that
\begin{equation}\label{eqn:11020409}
    \frac{\|\bm{V}_r(\bar{\pi})-\bm{V}_r(\pi^*)\|_{\infty}}{\|\bm{V}_r(\pi^*)\|_{\infty}} \leq \kappa(B^{\pi^*})\cdot\left( \frac{\|\Delta B\|_{\infty}}{\|B^{\pi^*}\|_{\infty}}+\frac{\|\Delta \hat{\bm{r}}\|_{\infty}}{\|\hat{\bm{r}}(\pi^*)\|_{\infty}} \right),
\end{equation}
where $\kappa(B^{\pi^*})=\|B^{\pi^*}\|_{\infty}\cdot\|(B^{\pi^*})^{-1}\|_{\infty}$ denotes the conditional number of $B^{\pi^*}$. Note that we have the regret
\begin{equation}\label{eqn:11020410}
\begin{aligned}
\text{Regret}_r(\bar{\pi}, N_2)&= \bmu^\top (\bm{V}_r(\bar{\pi})-\bm{V}_r(\pi^*) )\leq (1-\gamma)\|\bm{V}_r(\bar{\pi})-\bm{V}_r(\pi^*)\|_{\infty} \\
&\leq (1-\gamma)\cdot\kappa(B^{\pi^*})\cdot \|\bm{V}_r(\pi^*)\|_{\infty}\cdot\left( \frac{\|\Delta B\|_{\infty}}{\|B^{\pi^*}\|_{\infty}}+\frac{\|\Delta \hat{\bm{r}}\|_{\infty}}{\|\hat{\bm{r}}(\pi^*)\|_{\infty}} \right).
\end{aligned}
\end{equation}
It is clear to see that
\begin{equation}\label{eqn:11022115}
\|\bm{V}_r(\pi^*)\|_{\infty}\leq\frac{1}{1-\gamma}
\end{equation}
and
\begin{equation}\label{eqn:11022116}
\|\bm{V}_r(\pi^*)\|_{\infty}\leq\frac{\|\hat{\bm{r}}(\pi^*)\|_{\infty}}{1-\gamma}.
\end{equation}
Following \cite{jin2020efficiently}, we have the following bound.
\begin{equation}\label{eqn:11022003}
\sigma'= \|(B^{\pi^*})^{-1}\|_{\infty}\leq\frac{1}{1-\gamma}.
\end{equation}
Also, from the definition, we have that
\begin{equation}\label{eqn:11022004}
\|B^{\pi^*}\|_{\infty} = 1-\gamma.
\end{equation}
Plugging the bound \eqref{eqn:11020408}, \eqref{eqn:11020406}, \eqref{eqn:11022115}, and \eqref{eqn:11022116}, into the inequality \eqref{eqn:11020410}, we have that
\begin{equation}\label{eqn:11020411}
\text{Regret}_r(\bar{\pi}, N_2) \leq \frac{2}{N_2\xi(1-\gamma)}+\frac{2m}{\xi\sigma^* N_2(1-\gamma)}\cdot \left( 16m+\frac{1024m^4}{(\xi\sigma^*)^2}\cdot\log(1/\eps) \right)\cdot\left( \log(N_2)-\log(\xi/2) \right)
\end{equation}
In a same way, for any $k\in[K]$, we obtain that
\begin{equation}\label{eqn:11020412}
\text{Regret}_k(\bar{\pi}, N_2) \leq \frac{2}{N_2\xi(1-\gamma)}+\frac{2m}{\xi\sigma^* N_2(1-\gamma)}\cdot \left( 16m+\frac{1024m^4}{(\xi\sigma^*)^2}\cdot\log(1/\eps) \right)\cdot\left( \log(N_2)-\log(\xi/2) \right)
\end{equation}
In order to translate the above bounds in \eqref{eqn:11020411} and \eqref{eqn:11020412} into the bound over $N_2$, we let the accuracy level $\veps$ satisfies the condition that
\[
\veps= \frac{2}{N_2\xi(1-\gamma)}+\frac{2m}{\xi\sigma^* N_2(1-\gamma)}\cdot \left( 16m+\frac{1024m^4}{(\xi\sigma^*)^2}\cdot\log(1/\eps) \right)\cdot\left( \log(N_2)-\log(\xi/2) \right)
\]
with $\eps=O(1/N_2^2)$. Then, we have that
\[
N_2= O\left( \frac{m^5\cdot\log(1/\xi)}{(\xi\sigma^*)^3(1-\gamma)}\cdot\frac{\log^2(1/\veps)}{\veps}\right)
\]
Our proof is thus completed.

\end{APPENDICES}

\end{document}